\documentclass[reqno]{amsart}

\usepackage{amsmath}
\usepackage{amsthm}
\usepackage{amsfonts}
\usepackage{amssymb}
\usepackage{enumerate}
\usepackage{graphicx}
\usepackage{stmaryrd}
\usepackage[nocompress]{cite}
\usepackage{comment}

\usepackage{hyperref}
\hypersetup{
    colorlinks,%
    citecolor=black,%
    filecolor=black,%
    linkcolor=black,%
    urlcolor=black
}

\DeclareMathOperator{\tr}{tr}

\DeclareMathOperator{\diag}{diag}
\DeclareMathOperator{\dist}{dist}

\DeclareMathOperator{\Span}{Span}

\newcommand{\Prob}{\mathbb{P}}
\newcommand{\E}{\mathbb{E}}

\renewcommand\Re{\operatorname{Re}}
\renewcommand\Im{\operatorname{Im}}
\newcommand{\eps}{\varepsilon}
\newcommand{\T}{\mathrm{T}}

\theoremstyle{plain}
  \newtheorem{theorem}{Theorem}

  \newtheorem{proposition}[theorem]{Proposition}
  
  \newtheorem{lemma}[theorem]{Lemma}

\theoremstyle{definition}
  \newtheorem{definition}[theorem]{Definition}
  
  \newtheorem{remark}[theorem]{Remark}

\begin{document}
\title[Gaussian noise: optimal estimates for singular subspace perturbation]{Matrices with Gaussian noise: optimal estimates for singular subspace perturbation} 

\author{Sean O'Rourke}
\thanks{S. O'Rourke has been supported in part by NSF grant DMS-1810500.}
\address{Department of Mathematics, University of Colorado at Boulder, Boulder, CO 80309 }
\email{sean.d.orourke@colorado.edu}

\author{Van Vu}
\thanks{V. Vu is partially supported by NSF grant DMS-1902825.}
\address{Department of Mathematics, Yale University, PO Box 208283, New Haven, CT 06520-8283, USA}
\email{van.vu@yale.edu}

\author{Ke Wang}
\thanks{K. Wang is partially supported by Hong Kong RGC grant GRF 16308219, GRF 16304222 and ECS 26304920.}
\address{Department of Mathematics, Hong Kong University of Science and Technology, Hong Kong}
\email{kewang@ust.hk}

\begin{abstract}
The Davis--Kahan--Wedin $\sin \Theta$ theorem describes how the singular subspaces of a matrix change when subjected to a small perturbation.  This classic result is sharp in the worst case scenario.  In this paper, we prove a stochastic version of the Davis--Kahan--Wedin $\sin \Theta$ theorem when the perturbation is a Gaussian random matrix.  Under certain structural assumptions, we obtain an optimal bound that significantly improves upon the classic Davis--Kahan--Wedin $\sin \Theta$ theorem.  One of our key tools is a new perturbation bound for the singular values, which may be of independent interest.   \end{abstract} 

\maketitle

\section{Introduction}

Consider an $N \times n$ (data) matrix $A$.  In practice,  it is common that we only have access to a corrupted (noisy) version $\widetilde{A}$ given by
\begin{equation} \label{def:tildeA}
	\widetilde{A} := A + E, 
\end{equation} 
where $E$ represents the noise matrix.   As a result, one must use $\widetilde A$ as input for all calculations and algorithms intended for $A$. 
A question of fundamental interest  is to estimate the impact of the noise $E$ on the output; see for instance \cite{MR2558901,MR1314843,MR2409803,MR3689312,MR3371006,MR841268,MR3091653,MR3827095,MR2893856,MR3346694,MR3640191} and references therein.

In modern studies, noise is often assumed to be random (e.g., Gaussian) and  the data matrix $A$ possesses certain structural properties.  For example, in a vast number of studies, researchers assume that $A$  has low rank \cite{MR2565240,MR2236170,Tomasi9795,Cands2010MatrixCW}, and our main results focus on this case.

Assume that the $N \times n$ data matrix $A$ has rank $r \geq 1$. We will often think of $r$ as a constant (or a parameter very small compared to the dimensions $N$ and $n$ such as $r \leq \log n$ or $r \leq n^{\epsilon}$).  The singular value decomposition (SVD) of $A$ takes the form $A=U \Sigma V^T,$ where $\Sigma= \diag(\sigma_1,\ldots, \sigma_r)$ is a diagonal matrix containing the non-zero singular values $\sigma_1\ge \sigma_2 \ge \cdots \ge \sigma_r>0$ of $A$; the columns of the matrices $U=(u_1,\ldots,u_r)$ and $V=(v_1,\ldots,v_r)$ are the orthonormal left and right singular vectors of $A$, respectively.  In other words, $u_i$ and $v_i$ are the left and right singular vectors corresponding to $\sigma_i$.  It follows that $U^T U = V^T V = I_r$, where $I_r$ is the $r \times r$ identity matrix.  For convenience, we will take $\sigma_{r+i} = 0$ for all $i \geq 1$.  

Recall that $\widetilde{A}$ is given in \eqref{def:tildeA}. Denote the SVD of $\widetilde A$ similarly by $\widetilde A = {\widetilde U} {\widetilde \Sigma} {\widetilde V}^\T$, where the diagonal entries of ${\widetilde \Sigma}$ are the singular values $\widetilde \sigma_1 \ge \widetilde \sigma_2 \ge \cdots \ge \widetilde \sigma_{\min\{N,n\}} \geq 0$, and the columns of $\widetilde U$ and $\widetilde V$ are the orthonormal left and right singular vectors, denoted by $\widetilde u_i$ and $\widetilde v_i$, respectively. 

Let $\Pi_s$ denotes the orthogonal projection onto the subspace spanned by the $s$ leading singular vectors of $A$  (either left or right). The matrix $\Pi_s A$ is the best rank $s$ approximation of $A$ \cite[Section 2.4]{GVLbook} and 
plays an important role in applications in almost every fields of science involving large data sets. Given the noise issue, it is thus of fundamental interest to bound the difference between $\Pi_s$ and its ``noisy" counterpart 
$\widetilde \Pi_s $ (the projection onto the subspace formed by the leading $s$ singular vectors of $\widetilde A$).  The goal of this paper is to bound this difference.  

As our main result is bit technical, let us first consider a toy case. Assume we want to compute the first (left) singular vector $u_1$ of the matrix $A$. 
If we only have access to the noisy matrix $\widetilde A$, we can only compute $\widetilde u_1$.  The famous Davis--Kahan--Wedin $\sin \Theta$ theorem \cite{DK,Wedin} provides a bound on  the difference between $\widetilde u_1$ and $u_1$.  
Two  parameters appear in  this bound: the gap  (or separation) $\delta_1$ between the largest singular values of $A$ given by 
\[ \delta_1 := \sigma_1 - \sigma_2 \] 
and the spectral norm of $E$ defined by
\[ \|E\| := \max_{\|u\| = 1} \|E u\|, \]
where $\|u\|$ denotes the Euclidean norm of the vector $u$.  

\begin{theorem}[Davis--Kahan--Wedin $\sin \Theta$ theorem] \label{thm:wedin}
One has
\[ \sin \angle (u_1, \widetilde u_1) \leq 2 \frac{ \|E \|}{ \delta_1 }, \] 
where $\angle (u_1, \widetilde u_1)$ is the acute angle between $u_1$  and $\widetilde u_1$, taken in $[0, \pi/2]$.  The same bound holds for $\sin \angle (v_1, \widetilde{v}_1)$.  
\end{theorem}

Theorem \ref{thm:wedin} follows as a simple corollary of the Davis--Kahan--Wedin $\sin \Theta$ theorem; see Theorem 4 from \cite{OVW2}, which also contains an example explaining the necessity of the 
appearance of the gap $\delta_1$.

 In \cite{OVW2}, the current authors considered random noise  and improved 
Theorem \ref{thm:wedin} by showing that a stronger bound 
\begin{equation} \label{eq:OVW}
	\sin \angle (u_1, \widetilde u_1) \lesssim  \frac{ C(r)} {\delta_1} + \frac{ \|E \|}{\sigma_1} + \frac{ \|E\|^2}{\sigma_1 \delta_1} 
\end{equation} 
holds with high probability (the probability space is generated by the randomness of the noise matrix $E$; see  \cite{OVW2} for details). Here, $C(r)$ is a parameter depending polynomially on $r$. 
To see how this improves upon the Davis--Kahan--Wedin $\sin \Theta$ theorem, let us mention that in most settings, the norm of the random matrix $E$ is polynomial in $N+n$. Thus, in the setting where $r$ is significantly smaller than the dimensions $N,n$, the first term 
$C(r)/\delta_1$ improves upon the term $\| E \| /\delta _1$ as it replaces a polynomial in $N+n$ by a polynomial in $r$.  The second term $\| E\| / \sigma_1$ represents the signal-to-noise ratio; notice that the denominator is the singular value $\sigma_1$ which is usually much larger than the gap $\delta_1$ between $\sigma_1$ and $\sigma_2$. Finally, the third term $\frac{ \|E\|^2}{\sigma_1 \delta_1}$ improves upon the term $\| E \| / \delta _1$ by the a factor involving the noise-to-signal ratio $\| E \| / \sigma_1$. This theorem generalized  Theorem 8 of  \cite{MR2846302}, where the second author considered Bernoulli random matrices.

 The first term on the right-hand side of \eqref{eq:OVW} was already conjectured in \cite{MR2846302}.  
 As noted above, the second term represents the signal-to-noise ratio.  Both terms are necessary (see below for further details). Indeed, if the gap $\delta_1$ is too small, there is a chance 
 that the two leading singular values (and thus the corresponding singular vectors) get ``swapped.'' Moreover, if the signal-to-noise ratio is low, then the data matrix is overwhelmed by noise, and the singular vectors behave more like random vectors.  In random matrix theory, this phenomenon is known as the BBP phase transition, named after Baik, Ben Arous, and P\'{e}ch\'{e} \cite{MR2165575}; see \cite{MR2165575,MR4367951} and references therein for further details.  
 
 It is conjectured that the third term on the right-hand side of \eqref{eq:OVW} is not necessary, under some mild assumptions. 
In this paper, we introduce a new method of analyzing random perturbations and confirm this conjecture (up to lower-order corrections). 
As the precise result is a bit technical, let us first state a simpler, but easier to read, version of the our main result. The asymptotic notation is used under the assumption that the dimensions $n$ and $N$ tend to infinity. 

\begin{theorem}[Perturbation with Gaussian noise; simplified asymptotic version]\label{thm:gaussian-friendly} 
Let $A$ and $E$ be $N \times n$ real matrices, where $A$ is deterministic and the entries of $E$ are jointly independent standard Gaussian random variables.  If $A$ has rank $r \geq 1$, then, with probability $1 -o(1)$, 
\begin{equation} \label{eq:gaussian-friendly}
	\sin \angle(u_1,\widetilde u_1) \lesssim \frac{C(r , \log (N+n) )}{ \delta_1}  + \frac{ \|E \|}{\sigma_1}, 
\end{equation} 
where $ C(s, t)$ is a positive parameter which grows at most polynomially in $s$ and $t$.
\end{theorem}

Theorem \ref{thm:gaussian-friendly} shows that the third term on the right-hand side of \eqref{eq:OVW} is not necessary and verifies the conjecture discussed above.  In Section \ref{sec:main}, we state the more detailed version of Theorem \ref{thm:gaussian-friendly} and its generalizations.  In this paper, we focus on the case when the entries of $E$ are jointly independent Gaussian random variables.  This assumption simplifies parts of the (already technical) proofs, but the method does extend to other distributions.  For instance, the results can be extended to the case when $E$ has independent and identically distributed sub-gaussian entries.  To relax the Gaussian assumption, one needs to establish a variant of the isotropic local law presented in Lemma \ref{lemma:assumpGOE} below.  Similar results have been established for more general entry distributions (see the discussion regarding the isotropic local laws in Section \ref{subsec:resolvent} for references).  We plan to discuss the non-Gaussian case in a separate paper.  

One can also consider the case when the entries of $E$ are not necessarily identically distributed (but are still jointly independent).  For example, consider the case when the entries of $E = (E_{ij})$ are jointly independent normal random variables, where $E_{ij}$ has mean zero and variance $\sigma^2_{ij}$.  Depending on the values of the variances $\sigma^2_{ij}$, the resolvent of $E$ may no longer satisfy an isotropic law and instead may satisfy an anisotropic law \cite{KY17}.  It is unclear if our methods can be adapted to deal with the anisotropic case.  Similarly, while we expect our results to also hold when there is a small dependence between the entries of $E$, our methods heavily rely on the joint independence of the entries. 

It is worth noting that when $\delta_1=\sigma_1$ (i.e., the rank-$1$ case) or when the spectral gap is of similar magnitude to the signal strength, $\delta_1 \asymp \sigma_1$, the right-hand side of \eqref{eq:gaussian-friendly} is proportional to $\|E\|/\sigma_1$.  Theorem \ref{thm:lowerbd} below implies that, up to lower order terms and constants, this bound is optimal.  Nevertheless, based on our bounds, it is apparent that $\sin \angle(u_1,\widetilde u_1)$ can be reasonably bounded in most cases, even when $\delta_1\ll \sigma_1$. Such relaxation offers improved outcomes in numerous applications, which will be elaborated upon in a forthcoming paper.

We remark that the bound in \eqref{eq:gaussian-friendly} does not depend on the condition number $\kappa = \sigma_1 / \sigma_r$ of $A$.  Instead, we have modeled our main results after the  Davis--Kahan--Wedin $\sin \Theta$ theorem (Theorem \ref{thm:wedin}), which instead relies on the gap $\delta_1$ to quantify how sensitive the matrix is to perturbations.   

\begin{figure}[!ht]
 \begin{center}
   \includegraphics[width=8cm]{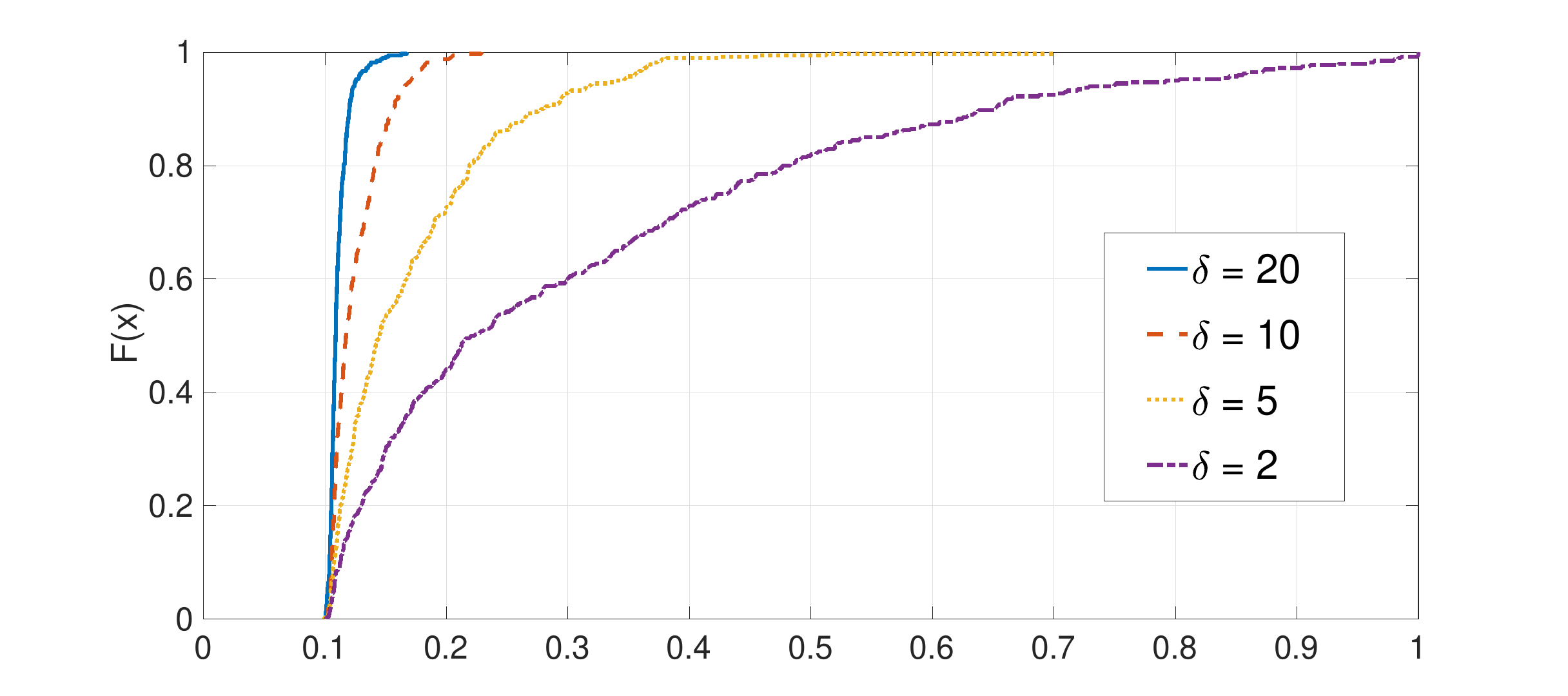}
   \caption{A plot of the cumulative distribution function $F$ of $\sin\angle(u_1,\widetilde u_1)$, where $F(x) = \Prob(\sin\angle(u_1,\widetilde u_1) \leq x)$ for $0 \leq x \leq 1$. We take $N=n=1000$ and $A=\diag(300,300-\delta,0,\cdots,0)$ with rank $r=2$ where the spectral gap $\delta=\delta_1$ is chosen to be $20$, $10$, $5$, and $2$. $E$ is a Gaussian matrix as in Theorem \ref{thm:gaussian-friendly}. Each curve is generated from $400$ samples. }
   \label{fig:per}   
 \end{center}
\end{figure}

Before concluding this section, we discuss the optimality of Theorem \ref{thm:gaussian-friendly}.  Numerical simulations show that, up to the particular form that $C(s,t)$ takes, the first term on the right-hand side of \eqref{eq:gaussian-friendly} is necessary; see Figure \ref{fig:per}. The second term on the right-hand side of \eqref{eq:gaussian-friendly} represents the signal-to-noise ratio; this term is also necessary as can be seen from the following lower bound.  

\begin{theorem}[Lower bound]\label{thm:lowerbd} Let $A$ and $E$ be $N \times n$ real matrices, where $A$ is deterministic with rank $r$ satisfying $1 \leq r \leq \frac{1}{2}\max(N,n)$ and the entries of $E$ are independent and identically distributed sub-gaussian\footnote{A random variable $\xi$ is sub-gaussian if there is a positive constant $c$ such that $\Prob(|\xi|\ge t) \le 2 \exp(-ct^2)$ for all $t\ge 0$; the largest constant $c > 0$ for which this property holds is called the sub-gaussian norm of $\xi$.  Standard normal random variables are sub-gaussian.} random variables with mean zero and unit variance. Then 
\begin{equation}\label{eq:lbdmax}
	\max\{\sin\angle (u_1, \widetilde{u}_1), \sin\angle (v_1, \widetilde{v}_1)\} \ge \frac{1}{8\sqrt{2}}\frac{ \frac{\|E\|}{\sigma_1} }{1+(1 + \sqrt{2})\frac{\|E\|}{\sigma_1}}
\end{equation}
with probability at least $1-C \exp\left(-c \max(N,n) \right),$ where $C, c>0$ are constants depending only on the sub-gaussian norm of the entries of $E$.
\end{theorem}

Notice that if $ \frac{\|E\|}{\sigma_1} \le 1$, then the right-hand side of \eqref{eq:lbdmax} simplifies to $c \frac{\|E\|}{\sigma_1}$ for some constant $c$. 
The proof of Theorem \ref{thm:lowerbd} can be found in Appendix \ref{app}. This result can be extended to more general random matrices with independent entries and matrix ensembles satisfying rotational invariance.
Due to space limitation,  we do not pursue such generalizations here.

\section*{Acknowledgements}
The authors are grateful to the anonymous referees for valuable feedback and corrections.

\section{Main results} \label{sec:main}

We state all of our main results in non-asymptotic forms (without using any asymptotic notation) and state all constants explicitly so that the results can be applied to matrices of any dimension. 

\subsection{Individual singular vector bounds} 
We first state the technical version of Theorem \ref{thm:gaussian-friendly} with precise dependences between the parameters.  
 
 \begin{theorem}[Perturbation with Gaussian noise] \label{thm:gaussian} 
Let $A$ and $E$ be $N \times n$ real matrices, where $A$ is deterministic with rank $r \geq 1$ and the entries of $E$ are jointly independent standard Gaussian random variables.  Let $K$ be an arbitrary positive constant, and denote $\eta:= 54 r\sqrt{(K+8) \log (N+n)}$. If $\sigma_{1} \ge 4(\sqrt N+ \sqrt{n}) + 140\eta r$ and $\delta_{1} = \sigma_{1} - \sigma_{2}\geq 100\eta r$, then
with probability at least $1 - 15 (N+n)^{-K} $
\begin{equation} \label{eq:gaussian}
	\sin \angle(u_1,\widetilde u_1) \le  54^2\sqrt{K+8}\frac{ r \sqrt{\log(N+n)}}{\delta_1} +2\frac{\|E\|}{\sigma_1}  
\end{equation} 
whenever $\frac{(\sqrt N+\sqrt n)^2}{\log(N+n)}>64(K+9)$. 
The same bound holds for $\sin \angle(v_1,\widetilde v_1) $. 
\end{theorem}

\begin{remark} \label{rem:constants}
Note that the constants in this theorem (such as $54$ and $100$) and our theorems below are chosen for convenience.  In order to keep the proof presentable, we have not tried to optimize these values.  However, the values can be significantly improved by tracking the constants throughout the proof.  For example, if one replaces the assumptions $\sigma_{1} \ge 4(\sqrt N+ \sqrt{n}) + 140\eta r$ and $\frac{(\sqrt N+\sqrt n)^2}{\log(N+n)}>64(K+9)$ by $\sigma_{1} \ge 11(\sqrt N+ \sqrt{n}) + 150\eta r$ and $\frac{(\sqrt N+\sqrt n)^2}{\log(N+n)}>200(K+9)$, then the constant $54^2$ appearing on the right-hand side of \eqref{eq:gaussian} can be replaced by $180$.  
\end{remark}

Theorem \ref{thm:gaussian} improves upon \eqref{eq:OVW} and Theorem \ref{thm:wedin} when the rank $r$ is sufficiently small and $\delta_1 \ll \|E\|$. As discussed in the previous section, the bound given in \eqref{eq:gaussian} is optimal, up to the choice of constants and the particular polynomial dependence on $r$ and $\log(N + n)$.

\subsection{Singular subspace bounds}
The results stated so far have focused on the singular vectors corresponding to the largest singular value.  More generally, we will consider the singular subspaces spanned by the first $j$ ($1\le j \le r$) singular vectors.  Define
\begin{equation}\label{def:subspace}
\begin{aligned}
&U_j := \Span\{u_1,\ldots, u_j \}, \quad V_j := \Span\{v_1,\ldots, v_j \},\\
&\widetilde U_j := \Span\{\widetilde u_1,\ldots, \widetilde u_j \}, \quad \widetilde V_j := \Span\{\widetilde v_1,\ldots, \widetilde v_j \}.
\end{aligned}
\end{equation}
Even more generally, for any $1\le k \le s \leq r$, let us denote
\[ U_{k,s} := \Span\{u_k,\ldots, u_s \}, \quad \widetilde U_{k,s} := \Span\{\widetilde u_k,\ldots, \widetilde u_s \} \]
and analogously for $V_{k,s}$ and $\widetilde V_{k,s}$. 

Recall that if $U$ and $V$ are two subspaces of the same dimension, then the largest principal angle $\angle(U,V)$ between them is given by 
\begin{equation} \label{def:UV}
	\sin \angle(U,V) := \max_{u \in U; u \neq 0} \min_{v \in V; v \neq 0} \sin \angle(u,v) = \|P_U - P_V \| = \|P_{U^\perp} P_{V} \|, 
\end{equation}
where $P_W$ is the orthogonal projection matrix onto subspace $W$.  We define the gaps (or separations) between the singular values of $A$ by \[ \delta_i = \sigma_i - \sigma_{i+1} \] for $1 \leq i \leq r$, where we use the convention that $\sigma_{r+1} = 0$.

Theorem \ref{thm:gaussian-friendly} can be generalized to the following.  

\begin{theorem}[Singular subspace bounds; simplified asymptotic version] \label{thm:subspace-friendly}
Let $A$ and $E$ be $N \times n$ real matrices, where $A$ is deterministic with rank $r \geq 1$ and the entries of $E$ are jointly independent standard Gaussian random variables.  For any $1\le r_0 \le r$,  if $\min_{\sigma_l \neq \sigma_j, 1\le l,j \le r_0} |\sigma_l - \sigma_j| \ge C(r , \log (N+n) )$, then, with probability $1 -o(1)$,  
\[ \sin \angle(U_{r_0}, \widetilde{U}_{r_0})  \lesssim \frac{C(r , \log (N+n) )}{ \delta_{r_0}}   + \frac{ \|E \|}{\sigma_{r_0}}, \] 
where $C(s, t)$ is a positive parameter that grows at most polynomially in $s$ and $t$.
\end{theorem}

Here, the minimum $\min_{\sigma_l \neq \sigma_j, 1\le l,j \le r_0} |\sigma_l - \sigma_j| $ is over all distinct singular values $\sigma_l \neq \sigma_j$, $j, l \leq r_0$.  In particular, this includes the case when some of the singular values of $A$ may be repeated.  Since the minimum is only over distinct singular values, even if some singular values occur with multiplicity, it is always the case that $\min_{\sigma_l \neq \sigma_j, 1\le l,j \le r_0} |\sigma_l - \sigma_j| > 0$.  The technical version of Theorem \ref{thm:subspace-friendly} is given below.  

\begin{theorem}[Singular subspace bounds] \label{thm:subspace}
Let $A$ and $E$ be $N \times n$ real matrices, where $A$ is deterministic and the entries of $E$ are jointly independent standard Gaussian random variables. Assume $A$ has rank $r \geq 1$. Let $K>0$ be any constant and denote $\eta:= 54 r\sqrt{(K+8) \log (N+n)}$. Assume $\frac{(\sqrt N+\sqrt n)^2}{\log(N+n)}>64(K+9)$. For any $1\le r_0 \le r$, if $ \sigma_{r_0} \ge 4(\sqrt N+ \sqrt{n}) + 140\eta r$, $\delta_{r_0}\geq 100\eta r$ and $\min_{\sigma_l \neq \sigma_j, 1\le l,j \le r_0} |\sigma_l - \sigma_j| \ge 100\eta r$, then
\begin{equation}\label{eq:subspacebd}
\sin \angle (U_{r_0}, \widetilde{U}_{r_0}) \leq  21\sqrt{2} \eta\sqrt{\sum_{j=1}^{r_0} \frac{1 }{(\sigma_j-\sigma_{r_0+1})^2}} + 2 \frac{ \|E \|}{\sigma_{r_0} } 
\end{equation}
with probability at least $1 - 15(N+n)^{-K}$. The same conclusion also holds for $\sin \angle (V_{r_0}, \widetilde{V}_{r_0})$. 

In addition, for any $1< k \le s \le r_0$, if $\min\{\delta_{k-1}, \delta_s\} \ge 100\eta r$, then 
\begin{align*}
\sin \angle (U_{k,s}, \widetilde{U}_{k,s}) \leq &21\sqrt{2} \eta\left(\sqrt{\sum_{j=1}^{k-1} \frac{1 }{(\sigma_j-\sigma_{k})^2}} +  \sqrt{\sum_{j=k}^{s} \frac{1 }{(\sigma_j-\sigma_{s+1})^2}}\right) \\
&\qquad+ 2 \left(\frac{ \| E \|} { \sigma_{k-1} }+ \frac{ \| E \|} { \sigma_{s} }\right)
\end{align*} with probability at least $1 - 15(N+n)^{-K}$.
The same conclusion also holds for $\sin \angle (V_{k,s}, \widetilde{V}_{k,s})$. 
\end{theorem}
The choice of constants, such as 100 and 140, in Theorem \ref{thm:subspace} is for convenience; an inspection of the proof will reveal exactly how much these constants can be optimized (see Remark \ref{rem:constants}).


The key to proving our main results is a precise prediction for the location of the singular values of $A+E$.  In order to obtain optimal control of the singular values, one cannot simply compare $\widetilde{\sigma}_j$ to $\sigma_j$ (or more conveniently $\widetilde{\sigma}^2_j$ to $\sigma^2_j$).  Instead, we compare $\sigma_j^2$ to $\widetilde{\sigma}^2_j + \eps$, where $\eps$ is a random correction term (depending only on the matrix $E$).  This random correction term allows us to obtain a more precise prediction for the singular values. The precise result is given in Theorem \ref{thm:singularlocation}, which can be found in Section \ref{subsec:location}, after appropriate notations have been introduced.

\subsection{Comparison to other results in the literature}

Many classical results compare the singular vectors (alternatively, eigenvectors) of $A + E$ to those of $A$.  The study of eigenvector perturbations dates back to at least Rayleigh \cite{MR0016009} and Schr\"{o}dinger \cite{https://doi.org/10.1002/andp.19263840404}.  More recently, these results include the Davis--Kahan--Wedin $\sin \Theta$ theorem \cite{DK, Wedin}.  For the singular values (alternatively, eigenvalues), there are many classical results, including Weyl's bound \cite{Bhatia}.  In contrast to this work, all of these classical results focus on the case when $A$ and $E$ are deterministic. We refer the reader to the classical texts \cite{MR1061154, Bhatia, MR2978290} for further details and generalizations.  

The case when $E$ is random has only been studied more recently.  As discussed above, our main results in this paper improve upon the works \cite{OVW2,MR2846302}.  A number of similar results have focused on the case when $E$ has Gaussian entries.  For example, in \cite{MR3565274}, Koltchinskii and Xia derive concentration bounds for linear forms involving the singular vectors and this was later extended to tensors by Xia and Zhou in \cite{MR3960915}.  The non-asymptotic distribution of the singular vectors, up to rotation, is studied by R. Wang in \cite{MR3310977}.  A perturbative expansion of the coordinates of the eigenvectors is given in \cite{MR4260218}.  Allez and Bouchaud studied the eigenvector dynamics of $A + E$ when both $A$ and $E$ are real symmetric matrices and the entries of $E$ are constructed from a family of independent real Brownian motions \cite{MR3256861}.  

In the random matrix theory literature, there are a number of perturbation results; in contrast to this work, many of these results focus on the case when $\|A\|$ and  $\|E\|$ are proportional.  The works of Benaych--Georges and Nadakuditi \cite{MR2782201,MR2944410} have influenced this paper (and we discuss these works more below). The results in \cite{MR2782201,MR2944410} establish the almost sure convergence of the projection of the outlier singular vectors (resp. eigenvectors) onto the $r$-dimensional singular vector subspace (resp. eigenspace) of $A$, assuming $r$, the rank of $A$, is fixed and the dimensions $N,n$ tend to infinity. The limiting distribution of such projections is explicitly given in \cite{BDW21}.  In these papers, the norm of 
$A$ and $E$ must be comparable. We make no such assumption here. (In fact,  in applications, the intensity of the noise is expected to be much smaller than the key signals.)  Several related results for eigenvectors of random matrices are also discussed in the survey \cite{OVW}.  A different yet closely related type of perturbation comes from the spiked covariance model (see \cite{MR3334281,MR2485013,MR2165575, BKY16, BDWW20} and many references therein). 

Another class of results in the literature is motivated by applications.  Motivated by statistical machine learning, Abbe, Fan, Wang, and Zhong \cite{MR4124330} provide entry-wise bounds between the eigenvectors of a random matrix and those of its expectation.  With similar motivations, the geometry of the singular subspaces are studied using the two-to-infinity subordinate vector norm on matrices in \cite{MR3988761}.  In the real symmetric case, when the matrix $A$ is incoherent and has low rank, $\ell^\infty$-norm bounds for the eigenvectors are given in \cite{MR3827095}.  
Similar entrywise-type behaviors for the eigenvectors are studied in \cite{1702.00139, MR3782406}.  Both deviation and fluctuation results for the eigenvectors are presented in \cite{MR3912394} based on statistical motivations.  Applications of principal components analysis have also motivated a number of similar works, including \cite{MR4460578} and references therein.  

The stochastic block model has been extensively studied in recent years, especially in connection with spectral algorithms, which often take advantage of eigenvector perturbation results.  For example, motivated by the stochastic block model, Eldridge, Belkin, and Wang \cite{MR3857310} investigated random perturbations of real symmetric matrices.  In particular, their results focus on the eigenvalues and $\ell^\infty$-norm bounds for the eigenvectors, which improve upon classical bounds.  Additionally, we highlight the works \cite{MR2893856,1710.10936,MR1948742,MR3734334,1412.7335,MR3453283,pmlr-v40-Chin15,NIPS2016_a8849b05,MR3520025} concerning the stochastic block model, which are perhaps the most relevant to this paper.  

The list of works discussed above is far from complete and represents only a small fraction of the literature.  

\subsection{Outline and notation}
The paper is organized as follows.  Section \ref{sec:prelim} establishes the preliminary tools needed for the proofs of our main results. We introduce a key lemma, Lemma \ref{lemma:assumpGOE}, that shows that the resolvent of the random noise can be well approximated by a diagonal matrix. The proof of Lemma \ref{lemma:assumpGOE} is deferred to Section \ref{sec:lem:GOE}. Section \ref{sec:prelim}  also contains a brief overview of our proof techniques.  In Section \ref{sec:thm:subspace}, we prove Theorem \ref{thm:subspace} using Lemma \ref{lemma:assumpGOE}. Section \ref{sec:thm:singularlocation} contains the proof of Theorem \ref{thm:singularlocation}, which describes the precise location of the perturbed singular values of $\widetilde A$.   The proof of Theorem \ref{thm:lowerbd} is presented in Appendix \ref{app}. In Appendix \ref{app:B}, we collect the proofs of Propositions \ref{prop:spacebd}, \ref{prop:blockphi}, \ref{prop:evblock}, \eqref{eq:difference} and Lemma \ref{lem:svloc-large}.

Without loss of generality, we always assume $N\le n$, for if not, one can simply apply the results to the transposes of the matrices. We introduce the following notation.    For a finite set $S$, $|S|$ denotes the cardinality of $S$.  Recall that $\| M \|$ denotes the spectral norm of the matrix $M$, and let $\|M\|_F$ denote the Frobenius norm.  For a vector $x$, $\|x\|$ will be its Euclidean norm.  The matrix $I_n$ is the $n \times n$ identity matrix.  We will often simply write $I$ when the size can be deduced from context.  For integers $m_2\geq m_1 \geq 1$, we let $\llbracket m_1,m_2\rrbracket := \{m_1, \ldots, m_2\}$ denote the discrete interval.  The distance between a point $z\in \mathbb C$ and a set $\mathcal G \subset \mathbb{C}$ is $ \dist(z, \mathcal{G}) := \inf_{w \in \mathcal{G}} |z - w|.$ For two sets $\mathcal F, \mathcal G \subset \mathbb{C}$, the distance between them is $ \dist(\mathcal{F}, \mathcal{G}) := \inf_{z \in \mathcal{F}, w \in \mathcal{G}} |z - w| $.  For two random elements $x$ and $y$, we write $x \sim y$ if $x$ and $y$ have the same distribution.  The function $\log (\cdot)$ will always denote the natural logarithm.

\section{Basic tools and an overview of the proof} \label{sec:prelim}

In this section, we develop the preliminary tools needed to establish our main results.  Section \ref{sec:overview} contains a brief overview of our main proof techniques.

\subsection{Linear algebra}\label{sec:linearalgebra}
We first apply a linearization trick, which allows us to consider the eigenvalues and eigenvectors of a symmetric matrix instead of studying the singular values and vectors of a non-symmetric matrix.  

Consider the $(N+n)\times (N+n)$ matrices 
\begin{align*}
\mathcal A := \begin{pmatrix}
0& A\\
A^\T& 0
\end{pmatrix}
\quad\text{and} \quad \mathcal E:=\begin{pmatrix}
0& E\\
E^\T& 0
\end{pmatrix}
\end{align*}
in block form.  
Define
\[ \widetilde{\mathcal A}:=\mathcal A + \mathcal E. \]
The non-zero eigenvalues of $\mathcal A$ are given by $\pm \sigma_1, \ldots, \pm \sigma_r$.  Indeed, $\mathcal A (u_j^\T, v_j^\T)^\T = \sigma_j (u_j^\T, v_j^\T)^\T$ and $\mathcal A (u_j^\T, -v_j^\T)^\T = -\sigma_j (u_j^\T, -v_j^\T)^\T$. Denote these eigenvalues by $\lambda_j= \sigma_j$ and $\lambda_{j+r} = -\sigma_j$ for $1\le j \le r$. Then $\mathbf u_j:=\frac{1}{\sqrt 2}(u_j^\T, v_j^\T)^\T$  and $\mathbf u_{j+r}:=\frac{1}{\sqrt 2}(u_j^\T, -v_j^\T)^\T$ for $1\le j \le r$ are their corresponding orthonormal eigenvectors. The spectral decomposition of $\mathcal A$ is given by 
\begin{equation}\label{eq:A}
 \mathcal A = \mathcal{U} \mathcal{D} {\mathcal U}^\T, 
 \end{equation}
where $\mathcal U:=(\mathbf{u}_1,\ldots,\mathbf{u}_{2r})$ and $\mathcal D := \diag(\lambda_1,\cdots, \lambda_{2r})$. It follows that $\mathcal U^\T \mathcal U = I_{2r}$. Similarly, the non-zero eigenvalues  
of $\widetilde{\mathcal A}$ are denoted by $\widetilde\lambda_j=\widetilde\sigma_j$ and  $\widetilde\lambda_{j+\min\{N, n\}}=-\widetilde\sigma_j$ for $1\le j \le \min\{N, n\}$.  The eigenvector corresponding to $\widetilde\lambda_j$ is denoted by $\mathbf{\widetilde u}_j$ and is formed by the right and left singular vectors of $\widetilde{A}$.  

For $J \subset \llbracket 1, 2r\rrbracket$, we introduce the notation $\mathcal U_J$ to denote the $(N+n) \times |J|$ matrix formed from $\mathcal U$ by removing the columns containing ${\mathbf u}_i$ for $i \not\in J$.  Similarly, $\mathcal{D}_J$ will denote the $|J| \times |J|$ matrix formed from $\mathcal D$ by removing the rows and columns containing ${\lambda}_i$ for $i \not\in J$. Let $I:= \llbracket 1, 2r\rrbracket \setminus J$. In this way, we can decompose $A$ as
\begin{equation} \label{eq:Adecomp}
	\mathcal A = \mathcal U \mathcal D \mathcal U^T = {\mathcal U}_J {\mathcal D}_J {\mathcal U}_J^\T + {\mathcal U}_{I} {\mathcal D}_{I} {\mathcal U}_{I}^\T.   
\end{equation} 
With a slight abuse of notation, we also denote the subpace
\begin{align*}
\mathcal U_J: =\Span\{\mathbf u_{k}: {k\in J}\}.
\end{align*}
Let $P_{J}$ be the orthogonal projection onto the subspace $ \mathcal U_J$. Clearly, $P_J = \mathcal U_J \mathcal U_J^\T.$ 
Analogous notations $ \widetilde{\mathcal U}_J, \widetilde{P}_J, \widetilde{\mathcal D}_J$ are also defined for $\widetilde{\mathcal A}$.

For the remainder of the paper, it suffices to derive results on the eigenspaces of $\widetilde{\mathcal A}$ by noting the following linear algebra fact.

\begin{proposition}\label{prop:spacebd} Let $U, \widetilde U \subset \mathbb R^N$ and $V, \widetilde V \subset \mathbb R^n$ be subspaces of the same dimension $p$. Let $W$ and $\widetilde W$ be subspaces in $\mathbb R^{N+n}$ obtained by concatenating vectors from $U, V$ and $\widetilde U, \widetilde V$ respectively, i.e. $W=\{ w \in \mathbb R^{N+n} : w=(u^T, v^T)^T, u\in U, v\in V\}$ and $\widetilde W=\{ \widetilde w \in \mathbb R^{N+n} : \widetilde w=(\widetilde u^T, \widetilde v^T)^T, \widetilde u\in \widetilde U, \widetilde v\in \widetilde V\}$. Then
$$\max \{\sin\angle(U, {\widetilde U}), \sin\angle(V, {\widetilde V}) \} = \sin\angle(W, {\widetilde W}).$$
\end{proposition}

The proof of Proposition \ref{prop:spacebd} is deferred to  Appendix \ref{appB:prop7}. In particular, as a special case of Proposition \ref{prop:spacebd} one has
\begin{align}\label{eq:unformcontrol} 
\max \{\sin\angle(U_{r_0}, {\widetilde U}_{r_0}), \sin\angle(V_{r_0}, {\widetilde V}_{r_0}) \} = \sin\angle(\mathcal U_I, {\widetilde{\mathcal U}}_I) 
\end{align} for the index set $I:=\llbracket 1, r_0\rrbracket \cup \llbracket r+1, r+r_0\rrbracket$
and
\begin{align}\label{eq:unformcontrol1} 
\max \{ \sin \angle (U_{k,s}, \widetilde{U}_{k,s}), \sin \angle (V_{k,s}, \widetilde{V}_{k,s}) \} = \sin\angle(\mathcal U_{\mathsf I}, {\widetilde{\mathcal U}}_{\mathsf I}) 
\end{align}
where $\mathsf I:=\llbracket k,s\rrbracket \cup \llbracket r+k,r+s\rrbracket$, $1 \leq r_0 \leq r$, and $1 \leq k \leq s \leq r_0$.

We also recall the Weinstein-Aronszajn identity (also called Sylvester's determinant identity), see page 271 of \cite{Pggn}: if $B$ is an $n \times k$ matrix and $C$ is a $k \times n$ matrix then
\begin{equation} \label{eq:sylvester}
	\det(I_n - BC) = \det(I_k - CB).   
\end{equation} 
When $k < n$, \eqref{eq:sylvester} allows us to reduce an $n \times n$ determinant to a smaller $k \times k$ determinant.  

Weyl's inequality (see \cite[Corollary III.2.6]{Bhatia}) states that if $B$ and $C$ are $n \times n$ real symmetric matrices with eigenvalues $\lambda_1(B) \geq \cdots \geq \lambda_n(B)$ and $\lambda_1(C) \geq \cdots \geq \lambda_n(C)$, then
\begin{equation} \label{eq:weyl}
	\max_{1 \leq j \leq n} | \lambda_j(B) - \lambda_j(C) | \leq \| B - C \|. 
\end{equation} 

The resolvent identity 
\begin{equation} \label{eq:resolvent}
	B^{-1} - C^{-1} = B^{-1} (C - B) C^{-1} 
\end{equation}
holds for invertible matrices $B$ and $C$. 

\subsection{Resolvent}\label{subsec:resolvent}
For $z \in \mathbb{C}$ with $|z| > \|\mathcal E\|$, we define the resolvent of $\mathcal E$ as
\[ G(z) := (zI - \mathcal E)^{-1}. \]
Often we will drop the identity matrix and simply write $(z - \mathcal E)^{-1}$ for this matrix. We use $G_{ij}(z)$ to denote the $(i,j)$-entry of $G(z)$.  

The resolvent $G(z)$ is a heavily studied object in random matrix theory.  For example, one can study the eigenvalues of $\mathcal{E}$ (and hence the singular values of $E$) by analyzing the trace, $\tr G(z)$, since the trace is a meromorphic function with poles precisely at the eigenvalues.  In addition, the matrix $G(z)$ encodes the behavior of the eigenvectors of $\mathcal{E}$ (and hence the singular vectors of $E$).  
For many matrix models (see \cite{BEKYY14,MR3103909,HKR18,KY17,EYY,EKY13,EYY12,LSSY16,LS18} and references therein), the resolvent $G(z)$ can be approximated by a diagonal matrix. Here, we consider a random diagonal matrix 
\begin{equation}\label{eq:Phi}
 \Phi(z):=\begin{pmatrix}
\frac{1}{\phi_1(z)}I_N & 0 \\
0 &\frac{1}{\phi_2(z)}I_n
\end{pmatrix},
\end{equation}
where 
\begin{equation}\label{eq:phi12}
\phi_1(z):= z - \sum_{t \in \llbracket N+1,N+n \rrbracket} G_{tt}(z),\qquad\phi_2(z):= z - \sum_{s \in \llbracket 1,N \rrbracket} G_{ss}(z).
\end{equation}

Under the assumptions of Theorem \ref{thm:subspace}, where $\eta= 54 r\sqrt{(K+8) \log (N+n)}$, define a set in the complex plane in the neighborhood of any $\sigma\in \mathbb R$ by
\begin{equation} \label{eq:resolventset}
	S_{\sigma} : = \{w \in \mathbb{C} : | \Im(w) | \leq 20 \eta r, \sigma - 20 \eta r \leq \Re(w) \leq \frac{8}{7}\sigma +  20 \eta r \}.
\end{equation}

In the remainder of this paper, we define an index 
\begin{equation}\label{eq:index}
i_0:=\min\{ j \in \llbracket 1, r_0 \rrbracket : \sigma_j \le n^2 \}.
\end{equation}
Hence, for any the index $l<i_0$,  $\sigma_l >n^2$. Note that $i_0$ may not exists; in this case, $\sigma_j > n^2$ for all $1\le j \le r_0$.

\begin{lemma} \label{lemma:assumpGOE}
Under the assumptions of Theorem \ref{thm:subspace} and the additional assumption that $i_0$ defined in \eqref{eq:index} exists, one has
\[ \max_{ j \in \llbracket i_0, r_0\rrbracket }\max_{ z \in S_{\sigma_j}} |z|^2 \left\|  \mathcal U^\T \left(G(z)-\Phi(z) \right) \mathcal U \right\| \leq 54 r \sqrt{(K+8)\log (N+n)}\] 
with probability at least $1 - 10(N+n)^{-(K+1)}$.
\end{lemma}
Lemma \ref{lemma:assumpGOE} is similar to many isotropic laws for random matrices; see, for instance, \cite{BEKYY14,MR3103909,HKR18,KY17} and references therein.  Roughly speaking, Lemma \ref{lemma:assumpGOE} quantitatively controls how close $G(z)$ is to the diagonal matrix $\Phi(z)$ and will be a fundamental tool in our proofs.  Unfortunately, we are not aware of any results in the literature that imply Lemma \ref{lemma:assumpGOE} as stated due to the block structure that $\mathcal{E}$ takes and the particular spectral domain $S_\sigma$ we are interested in.  We present the proof of this lemma in Section \ref{sec:lem:GOE}.  The assumption $\sigma_j \le n^2$ is purely technical: it is used in the proof for carrying out a volume argument on a bounded set. The cutoff $n^2$ is chosen for convenience. An inspection of the proof reveals the conclusion of Lemma \ref{lemma:assumpGOE} holds for all $\sigma_j$'s satisfying $\sigma_j \le n^{c}$ for a fixed $c>1$ by adjusting the tail probability accordingly.

Finally, we collect some basic facts about the matrix $\Phi(z)$. By setting 
\[
\mathcal I^\mathrm{u} := \begin{pmatrix}
I_N & 0\\
0& 0
\end{pmatrix} 
\quad\text{and}\quad
\mathcal I^\mathrm{d} := \begin{pmatrix}
0 & 0\\
0& I_n
\end{pmatrix},
 \]
one can rewrite \eqref{eq:phi12} as 
\begin{align}\label{eq:defphi}
\phi_1(z) = z - \tr \mathcal I^\mathrm{d} G(z), \qquad \phi_2(z) = z- \tr \mathcal I^\mathrm{u} G(z).
\end{align}
It can be derived by elementary linear algebra (see Appendix \ref{app:diff} for the proof) that 
\begin{align}\label{eq:difference}
\phi_1(z) = \phi_2(z) - \frac{1}{z}(n-N).
\end{align}

From the definition of $ \mathcal U$ in \eqref{eq:A}, it is easy to verify that  
\begin{align}\label{eq:productPhi}
 \mathcal U^\T \Phi(z) \mathcal U = \begin{pmatrix}
\alpha(z) I_r & \beta(z) I_r\\
\beta(z) I_r & \alpha(z) I_r
\end{pmatrix}, 
\end{align}
where we denote $$\alpha(z):=\frac{1}{2}\left(\frac{1}{\phi_1(z)} + \frac{1}{\phi_2(z)} \right)\quad \text{and}\quad \beta(z):= \frac{1}{2}\left(\frac{1}{\phi_1(z)} - \frac{1}{\phi_2(z)} \right)$$ for notational brevity. It follows that 
\begin{align}\label{eq:Phinorm}
\|\mathcal U^\T \Phi(z) \mathcal U\| = \max \left\{\frac{1}{|\phi_1(z)|}, \frac{1}{|\phi_2(z)|} \right\}.
\end{align}

Sometimes, we drop the $z$-dependence of $\alpha(z), \beta(z),\phi_1(z), \phi_2(z)$ and simply write $\alpha,\beta,\phi_1,\phi_2$ when the context is clear.

The following technical results can be derived via basic linear algebra and the proofs are deferred to Appendix \ref{app:B}. 
\begin{proposition}\label{prop:blockphi}For $1\le r_0 < r$, denote the index sets $I:=\llbracket 1, r_0\rrbracket \cup \llbracket r+1, r+r_0\rrbracket$ and $J:=\llbracket 1, 2r\rrbracket\setminus I$. For any $x\in \mathbb R$ satisfying $|x|>\|\mathcal E\|$, the singular values of $I_{2r-2r_0} -\mathcal U_J^\T \Phi(x) {\mathcal U}_J \mathcal D_J$ are given by
\begin{align*}
\left|\sqrt{1+\beta(x)^2 \sigma_t^2} \pm |\alpha(x)| \sigma_t \right|
\end{align*}
for $r_0+1 \le t \le r$.
\end{proposition}

\begin{proposition}\label{prop:evblock}For any $z\in \mathbb C$ satisfying $|z|>\|\mathcal E \|$, the matrix 
\[ \mathcal D^{-1} - \mathcal U^\T \Phi(z) \mathcal U \]
is invertible with probability 1, and one has
\begin{align*}
\left \| \left( \mathcal D^{-1} - \mathcal U^\T \Phi(z) \mathcal U \right)^{-1} \right \|= \max_{1\le l \le r} \frac{\sigma_l}{|\sigma_l^2 - \phi_1 \phi_2|} \mathcal Q^{1/2},
\end{align*}
where $$\mathcal Q:=|\phi_1 \phi_2|^2 + \frac{1}{2}\sigma_l^2 (|\phi_1|^2 + |\phi_2|^2) + \frac{1}{2}\sigma_l \left[ 4 |\phi_1 \phi_2|^2 |\phi_1 + \bar{\phi}_2|^2 + \sigma_l^2(|\phi_1|^2 - |\phi_2|^2)^2\right]^{1/2}.$$
\end{proposition}

\subsection{Singular value locations} \label{subsec:location} In this subsection, we introduce results that describe the precise locations of the singular values of $\mathcal A + \mathcal E$, and they will play a key role in the proof of Theorem \ref{thm:subspace}. 

To this end, consider the random function
\begin{equation} \label{eq:def:varphi}
	\varphi(z):=\phi_1(z)\phi_2(z),
\end{equation} 
where $\phi_1(z)$ and $\phi_2(z)$ are defined in \eqref{eq:phi12}, and recall 
the set $S_\sigma$ for any $\sigma \in \mathbb R$ defined in \eqref{eq:resolventset}.

\begin{theorem}[Singular value locations] \label{thm:singularlocation}
Under the assumptions of Theorem \ref{thm:subspace} and the additional assumption that $i_0$ defined in \eqref{eq:index} exists, for any $i_0 \le j \le r_0$, if $\sigma_j$ has multiplicity $\alpha_j$ and is denoted by $ \sigma_j= \sigma_{j+1}=\ldots=\sigma_{j+\alpha_j-1}$, then $\widetilde \sigma_j, \widetilde \sigma_{j+1},\ldots, \widetilde \sigma_{j+\alpha_j-1}$ are in the set $S_{\sigma_j}$ specified in \eqref{eq:resolventset}, and
\begin{align}\label{eq:location}
|\varphi(\widetilde{\sigma}_{j+s}) - \sigma_{j+s}^2| \leq \frac{115}{4} \eta r \left(\widetilde{\sigma}_{j+s}+\frac{8}{7}\sigma_{j+s} \right) \quad \text{for } 0\le s \le \alpha_j-1
\end{align}
with probability at least $1-10(N+n)^{-K}$.
\end{theorem}

Since $\varphi(z) = z^2 + \eps(z)$, where $\eps(z)$ is a random term depending on $z$ and the resolvent $G(z)$, Theorem \ref{thm:singularlocation} allows us to approximate the (squared) singular values of $\widetilde A$ with those of $A$, up to the random $\eps(z)$ correction term.  While this random correction term may seem odd at first, it allows for the much sharper bound appearing on the right-hand side of \eqref{eq:location}, which in many cases is a significant improvement over classic, deterministic bounds (such as Weyl's inequality).  Intuitively, since the singular values of $\widetilde{A}$ are random, it makes sense that one cannot only use deterministic values to accurately predict their locations.  One of the key differences between the techniques in this paper and those in \cite{OVW2} is that we take into account the precise behavior of this random correction term.  

The singular values considered in Theorem \ref{thm:singularlocation} are no larger than $n^2$. If a singular value $\sigma_l$ is sufficiently large, the effect of the noise $\mathcal E$ is negligible compared to the strong signal and consequently, the location of $\widetilde \sigma_l$ is very close to $\sigma_l$. The next lemma provides the perturbed singular value locations for large singular values. We defer its proof to Appendix \ref{appb:lem}.
\begin{lemma}\label{lem:svloc-large}Under the assumptions of Theorem \ref{thm:subspace}, 
$$\max_{l \in \llbracket 1, r_0\rrbracket : \sigma_l >\frac{1}{2}n^2} |\widetilde \sigma_l - \sigma_l |\le \eta r$$
with probability at least $1- (N+n)^{-2r^4(K+8)}$.
\end{lemma}

\subsection{Additional tools}
We now present a few additional tools we will need in the proofs.  
The first lemma captures the tail behavior for sub-exponential random variables. A random variable $X$ with mean $\mu = \E X$ is \emph{sub-exponential} with non-negative parameters $(\nu^2,\alpha)$ if $$\E \left( e^{\lambda(X-\mu)}\right) \le e^{\frac{\nu^2 \lambda^2}{2}} \qquad\text{for all } |\lambda|<\frac{1}{\alpha}.$$ It is easy to verify that for a standard Gaussian random variable $Z$, $Z^2$ is sub-exponential with parameters $(4,4)$, i.e. 
\begin{align}\label{eq:chi1sub}
\E\left( e^{\lambda(Z^2-1)} \right) \le e^{\frac{4\lambda^2}{2}},\quad \text{for all } |\lambda|<\frac{1}{4}. 
\end{align}
\begin{lemma}[Bernstein's inequality for sub-exponential random variables; see Proposition 2.9 in \cite{Wain19}]\label{lem:Bernstein} 
Suppose that $X$ is sub-exponential with parameters $(\nu^2,\alpha)$. Then $$\Prob(|X-\E X| \ge t) \le 2\exp\left( -\frac{1}{2}\min\left\{\frac{t^2}{\nu^2},\frac{t}{\alpha} \right\} \right).$$
\end{lemma}

The following lemma provides a non-asymptotic bound for the spectral norm of Gaussian matrices. 
\begin{lemma}[Spectral norm bound; see (2.3) from \cite{MR2827856}] \label{lemma:norm}
Let $E$ be an $N \times n$ matrix whose entries are independent standard Gaussian random variables.  Then
\[ \|E \| \leq 2 (\sqrt{N}+\sqrt{n}) \]
with probability at least $1 - 2 e^{-(\sqrt{N}+\sqrt{n})^2/2}$.  More generally,
$$\Prob (\|E\| \le \sqrt{N}+\sqrt{n} + t)\ge 1- 2 e^{-t^2/2}.$$
\end{lemma}
The next lemma bounds the operator norms of the resolvent and is used frequently in the proof. Let $\mathcal E^{(k)}$ be the minor of $\mathcal E$ with the $k$th row and column replaced by zeros and $G^{(k)}(z)$ the resolvent of $\mathcal E^{(k)}$ (see the precise definition at the beginning of Section \ref{sec:lem:GOE}). The proof is given in Appendix \ref{app:C}.
\begin{lemma} \label{lemma:resolventnorm}
On the event where $\|E \| \leq 2 (\sqrt{N}+\sqrt{n})$,
\[ \| G(z) \| \leq \frac{2}{|z|}, \qquad \|G^{(k)}(z) \| \leq \frac{2}{|z|} \]
and
\begin{equation}\label{eq:phibd}
\frac{7}{8}|z| \le |\phi_i(z)| \le \frac{9}{8} |z| \quad\text{for } i=1,2
\end{equation}
for any $z \in \mathbb{C}$ with $|z| \geq 4 (\sqrt{N}+\sqrt{n})$ and for any $k \in \llbracket 1,N+n \rrbracket$.  
\end{lemma}

The following lemma suggests that when $|z|$ is large, the resolvent $G(z)$ can be well approximated by a simple matrix. We defer its proof to Appendix \ref{app:C}.
\begin{lemma} \label{lem:simpleapprox}On the event where $\|E \| \leq 2 (\sqrt{N}+\sqrt{n})$,
\begin{equation*}
\left\| G(z) - \frac{1}{z} I_{N+n} - \frac{\mathcal E}{z^2}\right\| \le \frac{2\|\mathcal E\|^2}{|z|^3}
\end{equation*}
for any $z \in \mathbb{C}$ with $|z| \geq 4 (\sqrt{N}+\sqrt{n})$.
\end{lemma}

The next lemma bounds the operator norm of the random matrix $\mathcal U^T \mathcal E \mathcal U$ of size $2r\times 2r$ where $\mathcal U$ is defined in \eqref{eq:A}. Its proof can be found in Appendix \ref{app:C}.
\begin{lemma}\label{lem:Unoise}
Let $K$ be an arbitrary positive constant. With probability at least $1-2(N+n)^{-K}$, we have 
$$\| \mathcal U^T \mathcal E \mathcal U \| \le 2\sqrt{r} + \sqrt{2K\log(N+n)}.$$
\end{lemma}

\subsection{Overview of the proofs} \label{sec:overview}
We now provide a brief overview of the proofs of our main results.  For simplicity, we focus on the proof of Theorem \ref{thm:gaussian}, although the proof of Theorem \ref{thm:subspace} is similar.  

As noted above, Proposition \ref{prop:spacebd} allows us to focus on the eigenvectors of $\mathcal{A}$ and $\widetilde{\mathcal{A}}$.  In this case, we have
\[ \sin^2 \angle (\mathbf{u}_1, \widetilde{\mathbf{u}}_1) = 1 - | \mathbf{u}_1^\T \widetilde{\mathbf{u}}_1|^2 = \sum_{j \neq 1} | \mathbf{u}_1^\T \widetilde{\mathbf{u}}_j|^2. \]
Thus, our task reduces to controlling the inner products $\mathbf{u}_1^\T \widetilde{\mathbf{u}}_j$ for $j \neq 1$.  For simplicity, let us focus on the case when $j = 2$.  We start with the eigenvector-eigenvalue equation 
\[ \widetilde{\mathcal{A}} \widetilde{\mathbf{u}}_2 = \widetilde{\lambda}_2 \widetilde{\mathbf{u}}_2. \]
Rearranging this equation and using that $\widetilde{\mathcal{A}} = \mathcal{A} + \mathcal{E}$, we arrive at
\begin{equation} \label{eq:resolventdefeq}
	\widetilde{\mathbf{u}}_2 = G(\widetilde{\lambda}_2) \mathcal{A} \widetilde{\mathbf{u}}_2, 
\end{equation}
provided $|\widetilde{\lambda}_2| > \| \mathcal{E} \|$.  Multiplying \eqref{eq:resolventdefeq} by $\mathbf{u}_1^{\mathrm{T}}$, we find
\begin{equation} \label{eq:resolventdefeq2}
	\mathbf{u}_1^\T \widetilde{\mathbf{u}}_2 =  \mathbf{u}_1^{\mathrm{T}} G(\widetilde{\lambda}_2) \mathcal{U} \mathcal{D} \mathcal{U}^{\T} \widetilde{\mathbf{u}}_2 , 
\end{equation} 
where we used the spectral decomposition given in \eqref{eq:A}.  Using Lemma \ref{lemma:assumpGOE}, the resolvent matrix on the right-hand side of \eqref{eq:resolventdefeq2} can be approximated by the matrix $\Phi(\widetilde{\lambda}_2)$. Accordingly, we split the right-hand side of \eqref{eq:resolventdefeq2} into two terms
\begin{equation} \label{eq:resolventdefeq3}
\mathbf{u}_1^\T \widetilde{\mathbf{u}}_2 =  \mathbf{u}_1^{\mathrm{T}} \Phi(\widetilde{\lambda}_2) \mathcal{U} \mathcal{D} \mathcal{U}^{\T} \widetilde{\mathbf{u}}_2 + \mathbf{u}_1^{\T} \left(G(\widetilde{\lambda}_2)-\Phi(\widetilde{\lambda}_2) \right) \mathcal{U} \cdot\mathcal{D} \mathcal{U}^{\T} \widetilde{\mathbf{u}}_2, 
\end{equation} 
and Lemma \ref{lemma:assumpGOE} facilitates precise control over the second term appearing on the right-hand side of \eqref{eq:resolventdefeq3}.  In addition, we see that the term $\mathcal{U}^{\mathrm{T}} \widetilde{\mathbf{u}}_2$ appearing in the first term on the right-hand side of \eqref{eq:resolventdefeq3} contains a copy of the inner product $\mathbf{u}_1 \cdot \widetilde{\mathbf{u}}_2$ as one of its entries.  This will allow us to turn \eqref{eq:resolventdefeq2} into a recursive equation, with the inner product appearing on both sides of the equation.  More precisely, from the spectral decomposition $$\mathcal A = \mathcal{U} \mathcal{D} \mathcal{U}^{\T}=\lambda_1 \mathbf{u}_1 \mathbf{u}_1^\T + \lambda_{r+1} \mathbf{u}_{r+1} \mathbf{u}_{r+1}^\T + \mathcal U_0 \mathcal D_0 \mathcal U_0^\T,$$ where $\mathcal U_0$ is obtained from $\mathcal U$ by removing the columns $\mathbf{u}_1$ and $\mathbf{u}_{r+1}$ and $\mathcal D_0$ from $\mathcal D$ by removing the rows/columns containing $\lambda_1$ and $\lambda_{r+1}$. Hence, from  \eqref{eq:productPhi}, we arrive at 
$$\mathbf{u}_1^{\T} \Phi(\widetilde{\lambda}_2) \mathcal{U} \mathcal{D} \mathcal{U}^{\T} \widetilde{\mathbf{u}}_2 = \lambda_1  \alpha(\widetilde{\lambda}_2)  \mathbf{u}_1^\T \widetilde{\mathbf{u}}_2 + \lambda_{r+1}  \beta(\widetilde{\lambda}_2) \mathbf{u}_{r+1}^\T \widetilde{\mathbf{u}}_2.$$ Plugging the above equation back into \eqref{eq:resolventdefeq3} and rearranging the terms, we see that
\begin{equation} \label{eq:resolventdefeq4}
\left( 1- \lambda_1  \alpha(\widetilde{\lambda}_2) \right)\mathbf{u}_1^\T \widetilde{\mathbf{u}}_2 = \lambda_{r+1}  \beta(\widetilde{\lambda}_2) \mathbf{u}_{r+1}^\T \widetilde{\mathbf{u}}_2 +\mathbf{u}_1^{\T} \left(G(\widetilde{\lambda}_2)-\Phi(\widetilde{\lambda}_2) \right) \mathcal{U} \cdot\mathcal{D} \mathcal{U}^{\T} \widetilde{\mathbf{u}}_2.
\end{equation}
Understanding the location of $\widetilde{\lambda}_2$ is key to analyzing the left-hand side of \eqref{eq:resolventdefeq4}, and Theorem \ref{thm:singularlocation} will allow us to accurately estimate $$1- \lambda_1  \alpha(\widetilde{\lambda}_2) = 1- \sigma_1  \alpha(\widetilde{\sigma}_2)\approx 1- {\sigma_1}/{\sigma_2}; $$
this estimate explains the appearance of the spectral gap $\delta_1$ in the results. 

Actually, the term $\mathbf{u}_{r+1}^\T \widetilde{\mathbf{u}}_2$ on the right-hand side of \eqref{eq:resolventdefeq4} also encompasses information regarding $\mathbf{u}_1^\T \widetilde{\mathbf{u}}_2$ by the definition of the $\widetilde{\mathbf{u}}_j$'s and should not be overlooked. Besides, it is also wasteful to 
estimate each inner product $|\mathbf{u}_1^\T \widetilde{\mathbf{u}}_j|$ separately.  Instead, when proving Theorem \ref{thm:subspace}, we will group the inner products $| \mathbf{u}_1^\T \widetilde{\mathbf{u}}_j|$ into two categories: when $j \in  \llbracket r_0+1, r\rrbracket \cup \llbracket r+r_0+1, 2r\rrbracket$ and when $j > 2r$.  The arguments for the first category are similar to the methods discussed above (see Lemma \ref{lemma:close}).  The bound for the second category is much simpler and results in the signal-to-noise ratio term appearing in Theorem \ref{thm:subspace} (see Lemma \ref{lemma:twobound}).

\section{Proof of Theorem \ref{thm:subspace}} \label{sec:thm:subspace}

In this section, we prove Theorem \ref{thm:subspace} using Theorem \ref{thm:singularlocation} and Lemma \ref{lemma:assumpGOE}.  In view of \eqref{eq:unformcontrol}, it suffices to bound $\sin\angle(\mathcal U_I, {\widetilde{\mathcal U}}_I)$ for $I:=\llbracket 1, r_0\rrbracket \cup \llbracket r+1, r+r_0\rrbracket$. Assume the null space of $\mathcal A$ is spanned by the orthonormal basis $\{\mathbf u_{2r+1}, \ldots, \mathbf u_{N+n} \}$; in general, there will be many choices for this orthonormal basis and any choice will suffice for our purposes.  

The main idea of the proof is to divide the bound on $\sin\angle(\mathcal U_I, {\widetilde{\mathcal U}}_I)$ into two parts.  The first part involves projections of the vectors $\mathbf{u}_{r_0+1}, \ldots, \mathbf{u}_{r}, \mathbf{u}_{r+r_0+1}, \ldots, \mathbf{u}_{2r}$ while the second part involves projections of the vectors $\mathbf{u}_{2r + 1}, \ldots, \mathbf{u}_{n+N}$.  The latter terms can be controlled using the noise-to-signal ratio $\|E\|/\sigma_{r_0}$. The main argument is the estimate of the first term, which reflects that when the signal is stronger than the noise, the action of $\mathcal E$ is essentially on the $2r$-dimensional subspace spanned by the eigenvectors of $\mathcal A$ corresponding to the non-trivial eigenvalues. We use the resolvent $G(z)$ to extract information about the perturbed eigenvectors with the aid of Lemma \ref{lemma:assumpGOE} and Theorem \ref{thm:singularlocation}.

If $r_0=r$, then $\delta_r=\sigma_r$ and the conclusion follows from the Davis--Kahan--Wedin sine $\Theta$ 
theorem \cite{DK,Wedin}, which implies that 
$$\max \{\sin\angle(U_{r}, {\widetilde U}_{r}), \sin\angle(V_{r}, {\widetilde V}_{r}) \} \le 2\frac{\|E\|}{\sigma_r}.$$

Thus we assume $1\le r_0 \le  r-1$ throughout the proof. For the remainder of this section, we fix the index sets
$$I:=\llbracket 1, r_0\rrbracket \cup \llbracket r+1, r+r_0\rrbracket$$
and $$J:=\llbracket 1, 2r \rrbracket \setminus I= \llbracket r_0+1, r\rrbracket \cup \llbracket r+r_0+1, 2r\rrbracket.$$

\begin{lemma} \label{lemma:twobound}
Under the assumptions of Theorem \ref{thm:subspace}, 
\begin{align*}
\sin\angle(\mathcal U_I, {\widetilde{\mathcal U}}_I) &:= \| P_I - \widetilde{P}_I \| \leq \| P_J \widetilde{P}_I \| + 2 \frac{ \| E \|} { \sigma_{r_0} }
\end{align*}
with probability $1$.  
\end{lemma}
\begin{proof}
We assume $\sigma_{r_0} > 2 \|E\|$ as the bound is trivial otherwise.  
In view of \cite[Exercise VII.1.11]{Bhatia}, 
\begin{align*}
\| P_I - \widetilde{P}_I \|&=\|P_{I^c}  \widetilde{P}_I\| \\
&\le  \|P_J  \widetilde{P}_I\| + \|P_{\llbracket 2r+1, N+n\rrbracket}  \widetilde{P}_I\|.
\end{align*}
We aim to bound $$\|P_{\llbracket 2r+1, N+n\rrbracket}  \widetilde{P}_I\|\le \| \mathcal U_{\llbracket 2r+1, N+n\rrbracket}^\T  \widetilde{\mathcal U}_I \|.$$  From the spectral decomposition of $\widetilde{\mathcal A}$, we have
\begin{align*}
(\mathcal A + \mathcal E) \widetilde{\mathcal U}_I = \widetilde{\mathcal U}_I \widetilde{\mathcal D}_I.
\end{align*}
Multiplying by $\mathcal U_{\llbracket 2r+1, N+n\rrbracket}^\T$ on the left of the equation above, we further have
\begin{align*}
\mathcal U_{\llbracket 2r+1, N+n\rrbracket}^\T \mathcal E \widetilde{\mathcal U}_{I} = \mathcal U_{\llbracket 2r+1, N+n\rrbracket}^T  \widetilde{\mathcal U}_I \widetilde{\mathcal D}_I.
\end{align*}
As $\sigma_{r_0} > 2 \| E \|$ by supposition, Weyl's inequality \eqref{eq:weyl} implies that 
\begin{equation} \label{eq:hatlambdabnd}
	\widetilde{\sigma}_{i} \geq \sigma_{i} - \|E \| \geq \frac{1}{2} \sigma_{i} 
\end{equation}  
for $1\le i \le r_0$. 
Hence, $\widetilde{\mathcal D}_I$ is invertible since $|\widetilde\lambda_i| =\widetilde \sigma_i \ge \sigma_i -\|E\| >0$ for  $i\in \llbracket 1, r_0\rrbracket$ and $|\widetilde\lambda_i| = \widetilde \sigma_{i-r}>0$ for  $i\in \llbracket r+1, r+r_0\rrbracket$. It follows that 
\begin{align*}
\| \mathcal U_{\llbracket 2r+1, N+n\rrbracket}^\T  \widetilde{\mathcal U}_I \| &= \|\mathcal U_{\llbracket 2r+1, N+n\rrbracket}^\T \mathcal E \widetilde{\mathcal U}_I \widetilde{\mathcal D}_I^{-1}\|\le \|\mathcal E\| \|\widetilde{\mathcal D}_I^{-1}\| = \frac{\|E\|}{\widetilde{\sigma}_{r_0}}.
\end{align*}
Thus by another application of \eqref{eq:hatlambdabnd}, we get
\begin{align}\label{eq:conclusionbd1}
\|P_{\llbracket 2r+1, N+n\rrbracket}  \widetilde{P}_I\| \le \frac{\|E\|}{\widetilde{\sigma}_{r_0}} \le 2 \frac{\|E\|}{{\sigma}_{r_0}}.
\end{align}
as desired.  
\end{proof}

It remains to bound $\| P_J \widetilde{P}_I \| $, which is the content of the following lemma.      

\begin{lemma} \label{lemma:close}
Under the assumptions of Theorem \ref{thm:subspace}, we have\[  \| P_J \widetilde{P}_I \|  \leq  21 \sqrt{2}\eta \sqrt{\sum_{j=1}^{r_0} \frac{1 }{(\sigma_j-\sigma_{r_0+1})^2}} \]
with probability at least $1 - 15(N+n)^{-K}$.
\end{lemma}
\begin{proof}
In the following, we work on the event where $\|E\|\le 2(\sqrt{N}+\sqrt n)$; Lemma \ref{lemma:norm} shows this event holds with probability at least $1 - 2 e^{-(\sqrt{N}+\sqrt{n})^2/2}\ge 1 - 2(N+n)^{-32(K+2)}$ since $(\sqrt N+\sqrt n)^2>64(K+2)\log(N+n)$ by assumption. We start with the bound
\begin{align}\label{eq:lembd}
\| P_J \widetilde{P}_I \| &\le  \| \mathcal U_J^\T \widetilde{\mathcal U}_I \| 
\le \| \mathcal U_J^\T \widetilde{\mathcal U}_I \|_F = \sqrt{\sum_{i\in I} \| \mathcal U_J^\T \widetilde{\mathbf u}_i \|^2}.
\end{align}
Now we estimate $\| \mathcal U_J^\T \widetilde{\mathbf u}_i \|$ for each $i \in I$. 
We split the index set $I$ into two disjoint sets: 
$$\mathcal I_s := \{ i \in I: | \lambda_i| \le n^2\} \quad\text{and}\quad \mathcal I_b := \{ i \in I: | \lambda_i| > n^2\}.$$
Note that $\mathcal I_s$ or  $\mathcal I_b$ could be the empty set.

\medskip

\noindent{\bf Case 1: estimate $\| \mathcal U_J^\T \widetilde{\mathbf u}_i \|$ for $i \in \mathcal I_s$.} We first obtain an identity for the eigenvector $\widetilde{\mathbf u}_i$. By Weyl's inequality, $|\widetilde \lambda_i| \ge \widetilde \sigma_{r_0}\ge \sigma_{r_0}-\|E\|>\|E\|=\|\mathcal E\|$ by supposition on $\sigma_{r_0}$ and thus $G(\widetilde \lambda_i)$ and $\Phi(\widetilde \lambda_i)$ are well-defined. As $(\mathcal A + \mathcal E)\widetilde{\mathbf u}_i = \widetilde \lambda_i \widetilde{\mathbf u}_i$, we solve for $\widetilde{\mathbf u}_i$  to obtain 
\begin{equation*}
	\widetilde{\mathbf u}_i =(\widetilde \lambda_i I - \mathcal E)^{-1} \mathcal A \widetilde{\mathbf u}_i=G(\widetilde \lambda_i) \mathcal A \widetilde{\mathbf u}_i.
\end{equation*}

Rewrite the above equation
\begin{equation*}
\widetilde{\mathbf u}_i=\Phi(\widetilde \lambda_i) \mathcal A \widetilde{\mathbf u}_i + \left( G(\widetilde \lambda_i)-\Phi(\widetilde \lambda_i) \right) \mathcal A \widetilde{\mathbf u}_i
\end{equation*}
and multiply on the left by $\mathcal U_J^\T$ to get
\begin{equation}\label{eq:middle}
\mathcal U_J^\T \widetilde{\mathbf u}_i = \mathcal U_J^\T \Phi(\widetilde \lambda_i) \mathcal A \widetilde{\mathbf u}_i + \mathcal U_J^\T \left( G(\widetilde \lambda_i)-\Phi(\widetilde \lambda_i) \right) \mathcal A \widetilde{\mathbf u}_i.
\end{equation}
Plugging in \eqref{eq:Adecomp}, we further have
\begin{equation*}
\mathcal U_J^\T \widetilde{\mathbf u}_i = \mathcal U_J^\T \Phi(\widetilde \lambda_i) {\mathcal U}_J {\mathcal D}_J {\mathcal U}_J^\T \widetilde{\mathbf u}_i + \mathcal U_J^\T \left( G(\widetilde \lambda_i)-\Phi(\widetilde \lambda_i) \right) {\mathcal U} {\mathcal D} {\mathcal U}^\T \widetilde{\mathbf u}_i,
\end{equation*}
where we used $\mathcal U_J^\T \Phi(\widetilde \lambda_i) \mathcal U_{I}=0$.  Hence, 
\begin{equation} \label{eq:recursive}
\left( I_{2r-2r_0} - \mathcal U_J^\T \Phi(\widetilde \lambda_i) {\mathcal U}_J {\mathcal D}_J \right) \mathcal U_J^\T \widetilde{\mathbf u}_i = \mathcal U_J^\T \left( G(\widetilde \lambda_i)-\Phi(\widetilde \lambda_i) \right) {\mathcal U} {\mathcal D} {\mathcal U}^\T \widetilde{\mathbf u}_i.
\end{equation}

We are now in position to bound $\| \mathcal U_{J}^\T \widetilde{\mathbf u}_i \|$. This can be achieved by obtaining an upper bound for the right-hand side of \eqref{eq:recursive} and estimating the smallest singular value of the matrix 
\begin{equation}\label{eq:bigmatrix}
 I_{2r-2r_0} - \mathcal U_J^\T \Phi(\widetilde \lambda_i) {\mathcal U}_J {\mathcal D}_J 
\end{equation}
on the left-hand side of \eqref{eq:recursive}. We establish these estimates in the following two steps. Recall the index $i_0$ from \eqref{eq:index}. We will work on the event $\mathsf E:=\cap_{i=i_0}^{r_0} \mathsf E_i$ where  
\begin{align}\label{event}
\mathsf E_i&:=\left\{ \widetilde \sigma_i \in S_{\sigma_i}, \left\|  \mathcal U^\T \left(G(\widetilde \sigma_i)-\Phi(\widetilde \sigma_i) \right) \mathcal U \right\| \leq \frac{\eta}{\widetilde \sigma_i^2}, |\phi_1(\widetilde\sigma_i)\phi_2(\widetilde\sigma_i) -\sigma_i^2| \le \frac{115}{4} \eta r (\widetilde{\sigma}_{i}+\frac{8}{7}\sigma_{i})\right\}.
\end{align}
By Theorem \ref{thm:singularlocation} and Lemma \ref{lemma:assumpGOE}, the event $\mathsf E$ holds with probability at least $1-12(N+n)^{-K}$.

\medskip
\noindent\emph{Step 1. Upper bound for the right-hand side of \eqref{eq:recursive}}. Recall $I=\llbracket 1, r_0\rrbracket \cup \llbracket r+1, r+r_0\rrbracket$. We first consider the case when $i\in \llbracket 1, r_0\rrbracket$ and $\widetilde \lambda_i = \widetilde \sigma_i$. Note that $\mathcal U_J^\T \left( G(\widetilde \sigma_i)-\Phi(\widetilde \sigma_i) \right) {\mathcal U}$ is a sub-matrix of $\mathcal U^\T \left( G(\widetilde \sigma_i)-\Phi(\widetilde \sigma_i) \right) {\mathcal U}$. Thus, using \eqref{event} and the fact that the spectral norm of any sub-matrix is bounded by the spectral norm of the full matrix, we deduce that 
\begin{align*}
\| \mathcal U_J^\T \left( G(\widetilde \sigma_i)-\Phi(\widetilde \sigma_i) \right) {\mathcal U} {\mathcal D} {\mathcal U}^\T \widetilde{\mathbf u}_i \| &\le \frac{\eta}{\widetilde \sigma_i^2}\|{\mathcal D} {\mathcal U}^\T \widetilde{\mathbf u}_i \|.
\end{align*}
Observe from $(\mathcal A + \mathcal E)  \widetilde{\mathbf u}_i =  \widetilde{\sigma}_i \widetilde{\mathbf u}_i$ that $\mathcal U \mathcal D \mathcal U^\T \widetilde{\mathbf u}_i=( \widetilde{\sigma}_i I -\mathcal E)\widetilde{\mathbf u}_i$. Multiplying $\mathcal U^\T$ on both sides, we get the bound 
\begin{align}\label{eq:bdonev}
\|{\mathcal D} {\mathcal U}^\T \widetilde{\mathbf u}_i \| \le \|E\|+\widetilde{\sigma}_i \le 2\sigma_i
\end{align}
using the assumption $\|E\| \le \frac{1}{2}\sigma_i$ and Weyl's inequality. Hence, 
\begin{align}\label{eq:upbd1}
\left\| \mathcal U_J^\T \left( G(\widetilde \lambda_i)-\Phi(\widetilde \lambda_i) \right) {\mathcal U} {\mathcal D} {\mathcal U}^\T \widetilde{\mathbf u}_i \right\| &\le 2\frac{\eta\sigma_i}{\widetilde \sigma_i^2}.
\end{align}
For the case when $i\in \llbracket r+1, r+r_0\rrbracket$, $\widetilde \lambda_i = -\widetilde \sigma_{i-r}$. Observe that $$ G(-\widetilde \sigma_{i-r}) = (-\widetilde \sigma_{i-r}-\mathcal E)^{-1} = - (\widetilde \sigma_{i-r}+\mathcal E)^{-1}\sim -(\widetilde \sigma_{i-r}-\mathcal E)^{-1}=-G(\widetilde \sigma_{i-r}) $$ because the distribution of $\mathcal E$ is symmetric. Hence $$\Phi(-\widetilde \sigma_{i-r})\sim -\Phi(\widetilde \sigma_{i-r})$$ by the definition \eqref{eq:Phi}. Repeating the arguments from the previous case, we see that
\begin{align}\label{eq:upbd2}
\left\| \mathcal U_J^\T \left( G(\widetilde \lambda_i)-\Phi(\widetilde \lambda_i) \right) {\mathcal U} {\mathcal D} {\mathcal U}^\T \widetilde{\mathbf u}_i \right\| &\le 2\frac{\eta\sigma_{i-r}}{\widetilde \sigma_{i-r}^2}.
\end{align}
\medskip

\noindent\emph{Step 2. Lower bound for the smallest singular value of the matrix \eqref{eq:bigmatrix}}.
By Proposition \ref{prop:blockphi}, the singular values of $I_{2r-2r_0} - \mathcal U_J^\T \Phi(\widetilde \lambda_i) {\mathcal U}_J {\mathcal D}_J$ are given by \begin{align}\label{eq:svblk}
\left|\sqrt{1+\beta(\widetilde \lambda_i)^2 \sigma_t^2} \pm |\alpha(\widetilde \lambda_i)| \sigma_t \right|
\end{align}
for $r_0+1 \le t \le r$.

In order to bound the singular values, we first estimate $\phi_1(\widetilde\sigma_i)\phi_2(\widetilde\sigma_i)$, $\phi_1(\widetilde\sigma_i)$ and $\phi_2(\widetilde\sigma_i)$ for $i_0 \le i \le r_0$. Since $\widetilde \sigma_i \in S_{\sigma_i}$ by \eqref{event} where $S_{\sigma_i}$ is defined in \eqref{eq:resolventset}, we have $$ \widetilde \sigma_i \ge \sigma_i -20\eta r\ge  4(\sqrt N+\sqrt n)$$ 
and $$\frac{9}{7}\sigma_i \ge \frac{8}{7}\sigma_i + 20 \eta r\ge \widetilde \sigma_i \ge \sigma_i -20\eta r\ge \frac{6}{7}\sigma_i$$
by the supposition $\sigma_i \ge 4(\sqrt N+\sqrt n) + 140\eta r$.  

Continuing from \eqref{event}, we have
\begin{equation}\label{eq:phiprodbd}
|\phi_1(\widetilde\sigma_i)\phi_2(\widetilde\sigma_i) -\sigma_i^2| \le \frac{1955}{28} \eta r \sigma_i
\end{equation} and  consequently, 
$\frac{393}{784}\sigma_i^2\le \phi_1(\widetilde\sigma_i)\phi_2(\widetilde\sigma_i) \le \frac{1175}{784}\sigma_i^2
$
 from the supposition that $\sigma_i \ge 140\eta r$. 
Observe from \eqref{eq:phibd} that 
\begin{equation}\label{eq:phiprodbd1}
\frac{7}{8} \widetilde\sigma_i \le \phi_1(\widetilde\sigma_i) \le \frac{9}{8} \widetilde\sigma_i.
\end{equation}
 Thus
$$\frac{3}{4} \sigma_i \le \phi_1(\widetilde\sigma_i) \le \frac{81}{56} \sigma_i.$$ The same bound also holds from $\phi_2(\widetilde\sigma_i)$. Using these estimates, we crudely bound $$0< \alpha(\widetilde \sigma_i)=\frac{1}{2}\left(\frac{1}{\phi_1(\widetilde \sigma_i)} + \frac{1}{\phi_2(\widetilde \sigma_i)} \right) \le \frac{4}{3\sigma_i}, $$ and with \eqref{eq:difference} 
\begin{align*}
\beta(\widetilde \sigma_i)&=\frac{1}{2}\left(\frac{1}{\phi_1(\widetilde \sigma_i)} - \frac{1}{\phi_2(\widetilde \sigma_i)} \right) = \frac{\phi_2(\widetilde \sigma_i) - \phi_1(\widetilde \sigma_i)}{2 \phi_1(\widetilde \sigma_i)  \phi_2(\widetilde \sigma_i) }= \frac{n-N}{\widetilde \sigma_i}\frac{1}{ 2\phi_1(\widetilde \sigma_i)  \phi_2(\widetilde \sigma_i)} \le \frac{49}{786\sigma_i}
\end{align*}
by noting that $\widetilde \sigma_i,\sigma_i \ge 4(\sqrt N + \sqrt n)$. 

We are ready to bound the singular values of $I_{2r-r_0} - \mathcal U_J^\T \Phi(\widetilde \sigma_i) {\mathcal U}_J {\mathcal D}_J$. We start with the case when $i\in \llbracket 1, r_0\rrbracket$ and $\widetilde \lambda_i = \widetilde \sigma_i$. In view of \eqref{eq:svblk}, the goal is to bound
\begin{align*}
\min_{r_0+1\le t \le r} \left| \sqrt{1+\beta(\widetilde \sigma_i)^2 \sigma_t^2} \pm |\alpha(\widetilde \sigma_i)| \sigma_t \right|&=\min_{r_0+1\le t \le r} \left| \sqrt{1+\beta(\widetilde \sigma_i)^2 \sigma_t^2} - \alpha(\widetilde \sigma_i) \sigma_t \right|\\
&=\min_{r_0+1\le t \le r} \left| \frac{1-(\alpha(\widetilde \sigma_i)^2 -\beta(\widetilde \sigma_i)^2) \sigma_t^2}{\sqrt{1+\beta(\widetilde \sigma_i)^2 \sigma_t^2} + \alpha(\widetilde \sigma_i) \sigma_t}\right| \\
&=\min_{r_0+1\le t \le r} \frac{\left| 1- \frac{\sigma_t^2}{\phi_1(\widetilde\sigma_i)\phi_2(\widetilde\sigma_i)}\right|}{\sqrt{1+\beta(\widetilde \sigma_i)^2 \sigma_t^2} + \alpha(\widetilde \sigma_i) \sigma_t}.
\end{align*}
The upper bounds of $\alpha(\widetilde \sigma_i)$ and $\beta(\widetilde \sigma_i) $ obtained above, together with $\sigma_t/\sigma_i \le 1$, yield that
\begin{align*}
\sqrt{1+\beta(\widetilde \sigma_i)^2 \sigma_t^2} + \alpha(\widetilde \sigma_i) \sigma_t \le \sqrt{1+ \left(\frac{49}{786}\right)^2 \frac{\sigma_t^2}{\sigma_i^2}} + \frac{4\sigma_t}{3\sigma_i} \le 2.5
\end{align*}
for any $r_0+1\le t \le r$. Hence, 
\begin{align*}
\min_{r_0+1\le t \le r} \left| \sqrt{1+\beta(\widetilde \sigma_i)^2 \sigma_t^2} \pm \alpha(\widetilde \sigma_i) \sigma_t \right|& \ge \frac{1}{2.5}\min_{r_0+1\le t \le r}\left|  \frac{\phi_1(\widetilde\sigma_i)\phi_2(\widetilde\sigma_i)-\sigma_t^2}{\phi_1(\widetilde\sigma_i)\phi_2(\widetilde\sigma_i)}\right|\\
&\ge \frac{1}{2.5} \frac{\sigma_i^2 - \frac{1955}{28}\eta r \sigma_i -\sigma_{r_0+1}^2}{\phi_1(\widetilde\sigma_i)\phi_2(\widetilde\sigma_i)}.
\end{align*}
In the last inequality above, we invoked \eqref{eq:phiprodbd} which implies $$\phi_1(\widetilde\sigma_i)\phi_2(\widetilde\sigma_i)-\sigma_t^2 \ge \sigma_i^2 - \frac{1955}{28}\eta r \sigma_i -\sigma_{r_0+1}^2 >0$$ by the supposition $\sigma_i -\sigma_{r_0+1} \ge \delta_{r_0} \ge 100 \eta r$. Applying this supposition again, together with \eqref{eq:phiprodbd1}, we further deduce that
\begin{align}\label{eq:lwsv1}
\min_{r_0+1\le t \le r} \left| \sqrt{1+\beta(\widetilde \sigma_i)^2 \sigma_t^2} \pm \alpha(\widetilde \sigma_i) \sigma_t \right| &\ge \frac{1}{\widetilde\sigma_i^2}\frac{1}{2.5}\left(\frac{8}{9}\right)^2 \left(\sigma_i^2 - \sigma_{r_0+1}^2- \frac{1955}{28}\eta r \sigma_i \right)\nonumber\\
&\ge \frac{\sigma_i^2 - \sigma_{r_0+1}^2}{\widetilde\sigma_i^2}\frac{1}{2.5}\left(\frac{8}{9}\right)^2 \left(1-\frac{1955}{28} \frac{1}{100}\right)\nonumber\\
&\ge 0.0953\frac{\sigma_i^2 - \sigma_{r_0+1}^2}{\widetilde\sigma_i^2}.
\end{align}

For the case when $i\in \llbracket r+1, r+r_0\rrbracket$ and $\widetilde \lambda_i = -\widetilde \sigma_{i-r}$. Use the observation that $\alpha(\widetilde \lambda_i) \sim -\alpha(\widetilde \sigma_{i-r})$ and $\beta(\widetilde \lambda_i) \sim -\beta(\widetilde \sigma_{i-r})$ from the definitions \eqref{eq:defphi}. A simple modification of the proof for the first case shows that
\begin{equation}\label{eq:lwsv2}
\min_{r_0+1\le t \le r} \left| \sqrt{1+\beta(\widetilde \lambda_i)^2 \sigma_t^2} \pm |\alpha(\widetilde \lambda_i)| \sigma_t \right| \ge 0.0953\frac{\sigma_{i-r}^2 - \sigma_{r_0+1}^2}{\widetilde\sigma_{i-r}^2}.
\end{equation}

\medskip
\noindent\emph{Step 3. Combining the bounds above}.
With the estimates deduced in the previous two steps, we are in a position to bound $\| {\mathcal U}_J^\T \widetilde{\mathbf u}_i \|$. For $i\in \llbracket 1, r_0\rrbracket \cap \mathcal I_s$,
plugging \eqref{eq:upbd1} and \eqref{eq:lwsv1} and  into \eqref{eq:recursive}, we find that 
\begin{align}\label{eq:UJbd1}
\| {\mathcal U}_J^\T \widetilde{\mathbf u}_i \| \le \frac{2\eta \sigma_i}{0.0953(\sigma_i^2 - \sigma_{r_0+1}^2)}\le \frac{21\eta }{\sigma_i - \sigma_{r_0+1}}, 
\end{align}
and for $i\in \llbracket r+1, r+r_0\rrbracket \cap \mathcal I_s$, using  \eqref{eq:upbd2} and \eqref{eq:lwsv2}, we get
\begin{align}\label{eq:UJbd2}
\| {\mathcal U}_J^\T \widetilde{\mathbf u}_i \| \le \frac{21\eta }{\sigma_{i-r} - \sigma_{r_0+1}}.
\end{align}

\medskip

\noindent{\bf Case 2: estimate $\| \mathcal U_J^\T \widetilde{\mathbf u}_i \|$ for  $i \in \mathcal I_b$.} By Weyl's inequality, $|\widetilde \lambda_i| \ge n^2 - \|\mathcal E\| \ge 4(\sqrt N + \sqrt n)$ for every $i \in \mathcal I_b$. Hence, we apply Lemma \ref{lem:simpleapprox} to get
\begin{equation}\label{eq:case21}
\left\| G(\widetilde{\lambda}_i) - \Psi(\widetilde{\lambda}_i)\right\| \le \frac{2\|\mathcal E\|^2}{|\widetilde{\lambda}_i|^3}
\end{equation}
where $\Psi(z):=\frac{1}{z} I_{N+n} + \frac{1}{z^2}\mathcal E.$ Repeating the arguments as in the beginning of Case 1, we obtain the following equation similar to \eqref{eq:middle}: 
\begin{equation*}
\mathcal U_J^\T \widetilde{\mathbf u}_i = \mathcal U_J^\T \Psi(\widetilde \lambda_i) \mathcal A \widetilde{\mathbf u}_i + \mathcal U_J^\T \left( G(\widetilde \lambda_i)-\Psi(\widetilde \lambda_i) \right) \mathcal A \widetilde{\mathbf u}_i.
\end{equation*}
Plugging in \eqref{eq:Adecomp} and using the facts $\mathcal U_J^\T \mathcal U_I =0$ and $\mathcal U_J^\T \mathcal U_J =I$, we further get
\begin{align*}
\mathcal U_J^\T \widetilde{\mathbf u}_i = \frac{1}{\widetilde \lambda_i} \mathcal D_J \mathcal U_J^\T \widetilde{\mathbf u}_i +  \frac{1}{\widetilde \lambda_i^2} \mathcal U_J^\T \mathcal E \mathcal U \mathcal D \mathcal U^\T \widetilde{\mathbf u}_i + \mathcal U_J^\T \left( G(\widetilde \lambda_i)-\Psi(\widetilde \lambda_i) \right) \mathcal U \mathcal D \mathcal U^\T  \widetilde{\mathbf u}_i,
\end{align*}
which, by rearranging the terms, is reduced to
\begin{align*}
(\widetilde \lambda_i I- \mathcal D_J)\mathcal U_J^\T \widetilde{\mathbf u}_i = \frac{1}{\widetilde \lambda_i} \mathcal U_J^\T \mathcal E \mathcal U \mathcal D \mathcal U^\T \widetilde{\mathbf u}_i + \widetilde \lambda_i \mathcal U_J^\T \left( G(\widetilde \lambda_i)-\Psi(\widetilde \lambda_i) \right) \mathcal U \mathcal D \mathcal U^\T  \widetilde{\mathbf u}_i.
\end{align*}
Hence, 
\begin{align*}
\min_{j\in J} |\widetilde \lambda_i - \lambda_j| \cdot \|\mathcal U_J^\T \widetilde{\mathbf u}_i\| \le \frac{1}{|\widetilde \lambda_i|} \|  \mathcal U^\T \mathcal E \mathcal U\| \cdot \|\mathcal D \mathcal U^\T \widetilde{\mathbf u}_i\| + |\widetilde \lambda_i| \|G(\widetilde \lambda_i)-\Psi(\widetilde \lambda_i)\| \cdot \|\mathcal D \mathcal U^\T \widetilde{\mathbf u}_i\|.
\end{align*}
Note that $\|\mathcal D \mathcal U^\T \widetilde{\mathbf u}_i\| \le \|\mathcal E\| + |\widetilde \lambda_i|\le 2 |\widetilde \lambda_i|$ as in \eqref{eq:bdonev}. Inserting \eqref{eq:case21} into the above inequality, we arrive at
\begin{align}\label{eq:bdtwosides}
\min_{j\in J} |\widetilde \lambda_i - \lambda_j| \cdot \|\mathcal U_J^\T \widetilde{\mathbf u}_i\| \le 2 \|  \mathcal U^\T \mathcal E \mathcal U\| + 4\frac{\|E\|^2}{|\widetilde \lambda_i|}.
\end{align}
For the remaining arguments, we work on the event 
\begin{align}\label{eventF}
\mathsf F:=\big\{\|  \mathcal U^\T \mathcal E \mathcal U\| \le 2 \sqrt{r}+\sqrt{10(K+8)\log(N+n)}\big\}\cap \big\{ \max_{i\in \llbracket 1, r_0\rrbracket;\sigma_i>\frac{1}{2}n^2} |\widetilde \sigma_i - \sigma_i| \le \eta r \big\}.
\end{align}
By Lemma \ref{lem:svloc-large} and Lemma \ref{lem:Unoise}, the event $\mathsf F$ holds with probability at least $$1-2(N+n)^{-5(K+8)}-(N+n)^{-2r^4(K+8)} \ge 1-3(N+n)^{-2(K+8)}.$$

We continue the estimation of $\|\mathcal U_J^\T \widetilde{\mathbf u}_i\|$ from \eqref{eq:bdtwosides}.  Note from \eqref{eventF} that
$$\|  \mathcal U^\T \mathcal E \mathcal U\| \le 2 \sqrt{r}+\sqrt{10(K+8)\log(N+n)}\le \eta.$$ Also, ${\|E\|^2}/{|\widetilde \lambda_i|}\le 8(2\sqrt n)^2/n^2\le \eta$ where we used the crude bound $|\widetilde \lambda_i| \ge \frac{1}{2}n^2$ by Weyl's inequality. It follows that
\begin{equation}\label{eq:2ndtolast}
\min_{j\in J} |\widetilde \lambda_i - \lambda_j| \cdot \|\mathcal U_J^\T \widetilde{\mathbf u}_i\| \le 6 \eta.
\end{equation}
To bound the left-hand side of \eqref{eq:2ndtolast}, we first consider $i \in  \llbracket 1, r_0\rrbracket \cap \mathcal I_b$. Then $$\min_{j\in J} |\widetilde \lambda_i - \lambda_j| = \min_{r_0+1\le j\le r} |\widetilde \sigma_i - \sigma_j|=\widetilde \sigma_i -\sigma_{r_0+1}$$ by \eqref{eventF} and the supposition $\delta_{r_0}=\sigma_{r_0}-\sigma_{r_0+1}\ge 100 \eta r$. Next, applying $\delta_{r_0}\ge 100 \eta r$ again, we get
$$\min_{j\in J} |\widetilde \lambda_i - \lambda_j| = \sigma_i - \sigma_{r_0+1} + (\widetilde \sigma_i -\sigma_i) \ge 0.99 (\sigma_i - \sigma_{r_0+1}).$$ 
It follows from \eqref{eq:2ndtolast} that
\begin{align}\label{eq:UJbd3}
 \|\mathcal U_J^\T \widetilde{\mathbf u}_i\| \le \frac{7\eta}{\sigma_i - \sigma_{r_0+1}}
\end{align} 
for every $i \in \llbracket 1, r_0\rrbracket \cap \mathcal I_b$.  Finally, for $i \in \llbracket r+1, r+r_0\rrbracket \cap \mathcal I_b$, analogous arguments yield that 
\begin{align}\label{eq:UJbd4}
\| {\mathcal U}_J^\T \widetilde{\mathbf u}_i \| \le \frac{7\eta }{\sigma_{i-r} - \sigma_{r_0+1}}.
\end{align}
The proof is now complete by inserting \eqref{eq:UJbd1}, \eqref{eq:UJbd2}, \eqref{eq:UJbd3} and \eqref{eq:UJbd4} into \eqref{eq:lembd}.
\end{proof}

Combining Lemma \ref{lemma:twobound} and Lemma \ref{lemma:close} gives
\begin{align}\label{eq:eigenspacebd}
\sin \angle (\mathcal U_I, \widetilde{\mathcal U}_I) \le  21\sqrt{2} \eta \sqrt{\sum_{i=1}^{r_0} \frac{1}{(\sigma_i-\sigma_{r_0+1})^2}}+ 2 \frac{ \| E \|} { \sigma_{r_0} }.
\end{align}
The proof of \eqref{eq:subspacebd} now follows by \eqref{eq:unformcontrol}.  

Finally, the bound for $\sin \angle ( U_{\llbracket k,s\rrbracket}, \widetilde{ U}_{\llbracket k,s\rrbracket})$ and $\sin \angle ( V_{\llbracket k,s\rrbracket}, \widetilde{ V}_{\llbracket k,s\rrbracket})$ can be derived by an analogous procedure, and we briefly sketch the details below. First, recall from \eqref{eq:unformcontrol1} that it suffices to bound $\sin \angle (\mathcal U_{\mathsf I}, \widetilde{\mathcal U}_{\mathsf I})$ where
$\mathsf I:=\llbracket k,s\rrbracket \cup \llbracket r+k,r+s\rrbracket.$

Denote $$\mathsf J:=\llbracket 1,2r\rrbracket \setminus \mathsf I=\mathsf J_1 \cup \mathsf J_2, $$where
$$\mathsf J_1:= \llbracket 1,k-1\rrbracket \cup \llbracket r+1,r+k-1\rrbracket\quad\text{and} \quad \mathsf J_2:=\llbracket s+1,r\rrbracket\cup \llbracket r+s+1,2r\rrbracket.$$
Observe that
\begin{align*}
\sin \angle (\mathcal U_{\mathsf I}, \widetilde{\mathcal U}_{\mathsf I}) &= \| P_{\mathsf I} - \widetilde{P}_{\mathsf I} \|=\|P_{\mathsf I^c }  \widetilde{P}_{\mathsf I}\|\\
&\le \|P_{\mathsf J_1}  \widetilde{P}_{\mathsf I}\| + \|P_{\mathsf J_2 }  \widetilde{P}_{\mathsf I}\|+\|P_{\llbracket 2r+1, N+n\rrbracket }  \widetilde{P}_{\mathsf I}\|.
\end{align*}
Using the same method as in Lemma \ref{lemma:twobound}, one can bound
\begin{align*}
\|P_{\llbracket 2r+1, N+n\rrbracket }  \widetilde{P}_{\mathsf I}\| \le 2 \frac{ \| E \|} { \sigma_{s} }.
\end{align*}
Similarly, following the arguments in the proof of Lemma \ref{lemma:close}, one sees that 
\begin{align*}
\|P_{ \mathsf J_2 }  \widetilde{P}_{\mathsf I}\| \le  21\sqrt{2} \eta \sqrt{\sum_{i=k}^{s} \frac{1 }{(\sigma_i-\sigma_{s+1})^2}}.
\end{align*}

To conclude the result, we only need to bound 
\begin{align*}
\|P_{\mathsf J_1}  \widetilde{P}_{\mathsf I}\| &= \| (P_{\mathsf J_1 } -\widetilde{P}_{\mathsf J_1}) \widetilde{P}_{\mathsf I}\|\\
& \le \| P_{\mathsf J_1} -\widetilde{P}_{\mathsf J_1}\| =\sin \angle (\mathcal U_{\mathsf J_1}, \widetilde{\mathcal U}_{\mathsf J_1}) \le 21\sqrt{2} \eta \sqrt{\sum_{i=1}^{k-1} \frac{1}{(\sigma_i-\sigma_{k})^2}}+ 2 \frac{ \| E \|} { \sigma_{k-1} }
\end{align*}
by applying \eqref{eq:eigenspacebd}.
Hence, we obtain
\begin{align*}
&\sin \angle (\mathcal U_{\mathsf I}, \widetilde{\mathcal U}_{\mathsf I}) \\
& \le 21\sqrt{2} \eta \sqrt{\sum_{i=1}^{k-1} \frac{1 }{(\sigma_i-\sigma_{k})^2}} + 21\sqrt{2} \eta \sqrt{\sum_{i=k}^{s} \frac{1}{(\sigma_i-\sigma_{s+1})^2}} + 2\frac{ \| E \|} { \sigma_{k-1} }+ 2\frac{ \| E \|} { \sigma_{s} }.
\end{align*}
The proof is complete by \eqref{eq:unformcontrol1}.

\section{Proof of Theorem \ref{thm:singularlocation}} \label{sec:thm:singularlocation}

This section is devoted to the proof of Theorem \ref{thm:singularlocation}.  Throughout the proof, we work on the event where $\|\mathcal E\|=\|E\| \le 2(\sqrt N+ \sqrt{n})$; recall that Lemma \ref{lemma:norm} provides with probability at least $1 - 2 e^{-(\sqrt{N}+\sqrt{n})^2/2}\ge 1 - 2(N+n)^{-32(K+2)}$  that this event holds. For convenience, 
denote $$M:= 4(\sqrt N+ \sqrt{n}).$$ 
Note that the assumptions of Theorem  \ref{thm:singularlocation} guarantees that for any $z\in S_{\sigma_j}$ ($i_0\le j \le r_0$), $|z| \ge \Re(z) \ge \sigma_j - 20 \eta r > M$.

Our next lemma provides a way to locate the eigenvalues of a perturbed real symmetric matrix.  Similar results have been applied in the random matrix theory literature to study eigenvalues for both symmetric and non-symmetric random matrices \cite{MR3010398,MR2782201,MR2919200,MR2451053,MR2835249}.  

\begin{lemma}[Eigenvalue location criterion]
Assume $\mathcal A$ has rank $2r$ with the spectral decomposition $\mathcal A = \mathcal U \mathcal D \mathcal U^\T$, where $\mathcal U$ is an $(N+n) \times 2r$ matrix satisfying $\mathcal U^\T \mathcal U = I_{2r}$ and $\mathcal D$ is a $2r \times 2r$ diagonal matrix with non-zero $\lambda_1, \ldots, \lambda_{2r}$ on the diagonal.  Then the eigenvalues of $\mathcal A+\mathcal E$ outside of $[ - \|\mathcal E \||, \|\mathcal E\| ]$ are the zeros of the function 
\[ z \mapsto \det (\mathcal D^{-1} - \mathcal U^\T G(z) \mathcal U ). \]
Moreover, the algebraic multiplicity of each eigenvalue matches the corresponding multiplicity of each zero.  
\end{lemma}
\begin{proof}
The eigenvalues of $\mathcal A+\mathcal E$ are the zeros of the polynomial $\det(zI - \mathcal A - \mathcal E)$.  
For $|z| >  \|\mathcal E\|$, 
\begin{align*}
	\det(zI - \mathcal A - \mathcal E) &= \det(zI - \mathcal E) \det(I - G(z) \mathcal A) \\
	&= \det(zI - \mathcal E) \det( I - G(z) \mathcal U \mathcal D \mathcal U^\T) \\
	&= \det(zI - \mathcal E) \det( I - \mathcal U^\T G(z) \mathcal U \mathcal D)\\
	&=\det(zI - \mathcal E) \det(\mathcal D)\det( \mathcal D^{-1} - \mathcal U^\T G(z) \mathcal U).  
\end{align*}
by the Weinstein-Aronszajn identity \eqref{eq:sylvester}.  Since $\det(zI - \mathcal E) \neq 0$ for $|z| > \|\mathcal E\|$, the claim follows.  
\end{proof}

Define the functions
\[ f(z) := \det(\mathcal D^{-1} - \mathcal U^\T G(z) \mathcal U), \qquad g(z) := \det \left( \mathcal D^{-1} - \mathcal U^\T \Phi(z) \mathcal U \right), \]
where $\Phi(z)$ is given in \eqref{eq:Phi}.  Observe that, by Lemma \ref{lemma:resolventnorm},  $1/\phi_1(z)$, $1/\phi_2(z)$ and thus $\Phi(z)$ are well-defined for any $|z| > M$. Therefore, $f$ and $g$ are both complex analytic in the region $\{z \in \mathbb{C} : |z| > M\}$.  

An easy computation, together with \eqref{eq:productPhi}, yields that
\begin{align*}
g(z)=\prod_{l=1}^r \left( \frac{1}{\phi_1(z)\phi_2(z)} - \frac{1}{\sigma_l^2} \right)
\end{align*}
and thus the zeros of $g(z)$ are the values $z \in \mathbb{C}$ which satisfy the equations $\phi_1(z)\phi_2(z) = \sigma_l^2.$

Recall from \eqref{eq:def:varphi} and \eqref{eq:defphi} that 
\[ \varphi(z)= \phi_1(z)\phi_2(z)=(z-\tr \mathcal I^{\mathrm d} G(z))(z-\tr \mathcal I^{\mathrm u} G(z)). \]
The next lemma establishes several properties of $\varphi$ in the complex plane and on the real line.  

\begin{lemma} \label{lemma:varphi}
The function $\varphi$ satisfies the following properties.  
\begin{enumerate}[(i)]
\item  For $z, w \in \mathbb{C}$ with $|z|, |w|, |z+w| > M$, 
\begin{equation} \label{eq:lipschitz}
	\frac{9}{16} |z^2-w^2| \leq | \varphi(z) - \varphi(w)| \leq \frac{23}{16} |z^2- w^2|. 
\end{equation}
\item (Monotone) $\varphi$ is real-valued and strictly increasing on $[M, \infty)$.  
\item \label{item:crude} (Crude bounds) $0 < \varphi(z) < z^2$ for any $z \in [M, \infty)$.  
\end{enumerate}
\end{lemma}
\begin{proof}
Since $\varphi(z)=z^2 - z \tr G(z) + \tr \mathcal I^{\mathrm u} G(z)\tr \mathcal I^{\mathrm d} G(z)$, we first have
\begin{align}\label{eq:lemvarphi-eq1}
\varphi(z)- \varphi(w) &= (z^2 - w^2) - (z \tr G(z) - w \tr G(w) ) \nonumber\\
&+(\tr \mathcal I^{\mathrm u} G(z)\tr \mathcal I^{\mathrm d} G(z) - \tr \mathcal I^{\mathrm u} G(w)\tr \mathcal I^{\mathrm d} G(w)).
\end{align}
To establish \eqref{eq:lipschitz}, observe that
\begin{align*}
z \tr G(z) - w \tr G(w) &= (z \tr G(z) - w \tr G(z)) + w (\tr G(z) - \tr G(w))\\
&=(z-w) \tr G(z) + w(w-z) \tr G(z) G(w)
\end{align*}
by the resolvent identity \eqref{eq:resolvent}.  For  $|z|, |w|, |z+w| > M $, using Lemma \ref{lemma:resolventnorm}, we get
\begin{align}\label{eq:lemvarphi-eq2}
| z \tr G(z) - w \tr G(w) |&=|z-w| ( |\tr G(z)| + |w| |\tr G(z) G(w)| )\nonumber \\
&\le |z-w| \left( (N+n) \|G(z)\| + |w| (N+n)\|G(z) G(w)\| \right)\nonumber\\
&\le |z-w| (N+n)  \left( \frac{2}{|z|} + |w| \frac{4}{|z| |w|} \right) \nonumber \\
&= |z-w| (N+n) \frac{6}{|z|}\nonumber\\
&\le \frac{3}{8}|z^2 - w^2|.
\end{align}
Analogously,  one can split
\begin{align}\label{eq:estimate1}
| \tr \mathcal I^{\mathrm u} G(z)&\tr \mathcal I^{\mathrm d} G(z) - \tr \mathcal I^{\mathrm u} G(w)\tr \mathcal I^{\mathrm d} G(w) |\nonumber\\
& \le |\tr \mathcal I^{\mathrm u} G(z)\tr \mathcal I^{\mathrm d} G(z) - \tr \mathcal I^{\mathrm u} G(w)\tr \mathcal I^{\mathrm d} G(z)| \nonumber\\
&\qquad+ |\tr \mathcal I^{\mathrm u} G(w)\tr \mathcal I^{\mathrm d} G(z) - \tr \mathcal I^{\mathrm u} G(w)\tr \mathcal I^{\mathrm d} G(w)|.
\end{align}
For the first term on the right-hand side of \eqref{eq:estimate1}, by \eqref{eq:resolvent} and Lemma \ref{lemma:resolventnorm}, we have
\begin{align*}
|\tr \mathcal I^{\mathrm u} G(z)\tr \mathcal I^{\mathrm d} G(z) - \tr \mathcal I^{\mathrm u} G(w)\tr \mathcal I^{\mathrm d} G(z)| & = |\tr \mathcal I^{\mathrm d} G(z) | |\tr \mathcal I^{\mathrm u} G(z)- \tr \mathcal I^{\mathrm u} G(w) | \\
&=|\tr \mathcal I^{\mathrm d} G(z) | |(z-w)\tr\mathcal I^{\mathrm u} G(z)G(w) |\\
&\le \frac{1}{8} |z| |z-w| |\tr \mathcal I^{\mathrm u} G(z) G(w)| \\
&\le \frac{1}{8}|z| |z-w| (N + n) \|G(z)\| \|G(w)\|\\
&\le \frac{1}{2}\frac{N+n}{|w|}|z-w| \\
&\le \frac{1}{32}|z^2 - w^2|,
\end{align*}
where in the last inequality we used $\frac{N+n}{|w|} \le \frac{1}{16}|z+w|$ since $|w|, |z+w|>M$.
The second term on the right-hand side of \eqref{eq:estimate1} can be estimated likewise, and we conclude that  
$$| \tr \mathcal I^{\mathrm u} G(z)\tr \mathcal I^{\mathrm d} G(z) - \tr \mathcal I^{\mathrm u} G(w)\tr \mathcal I^{\mathrm d} G(w) | \le  \frac{1}{16}|z^2 - w^2|.$$
Combining the above bound and \eqref{eq:lemvarphi-eq2} with \eqref{eq:lemvarphi-eq1} yields \eqref{eq:lipschitz}.

To prove property (ii), it suffices to show that $\phi_1(z), \phi_2(z)$ are both positive and strictly increasing on $[M, \infty)$. By \eqref{eq:defphi} and the bound in \eqref{eq:bdid}, $\phi_1(z)>0$ for $z\ge M$. Using the expression in \eqref{eq:traceid}, we observe that 
\begin{equation} \label{eq:fromtraceid}
	\phi_1(z) = z -  \frac{1}{2} \sum_{i=1}^N \left( \frac{1}{z-\eta_i} + \frac{1}{z+\eta_i} \right) - \frac{1}{z}(n-N), 
\end{equation} 
where $\eta_1, \ldots, \eta_N$ are the singular values of $E$.  
Thus, $\phi_1(z)$ is a strictly increasing function since $z \geq M \geq |\eta_i|$ for all $i$. A similar argument shows that $\phi_2(z)$ is also positive and strictly increasing on $[M, \infty)$. Thus, $\varphi(z) = \phi_1(z)\phi_2(z)$ is the product of two positive strictly increasing functions and so is  strictly increasing.  In addition, it follows from \eqref{eq:fromtraceid} that $0< \phi_1(z)< z$ (and similarly $0< \phi_2(z)< z$) for $z \ge M$. Thus,  property \eqref{item:crude} follows immediately.  
\end{proof}
 
Fix an index $j\in \llbracket 1, r_0\rrbracket$. Since $\varphi(M) < M^2$ and $\lim_{z \to \infty} \varphi(z) = \infty$, it follows from the previous lemma that there exists a unique positive real number $z_j > M$ such that $\varphi(z_j) =\sigma_j^2$.  Similarly, if $\sigma_{l} > M$ for  $\sigma_l\neq \sigma_j$, then there exists a unique positive real number $z_{l}$ with $\varphi(z_{l}) = \sigma_{l}^2$ so that $z_j > z_{l}$ if $l>j$ and $z_j< z_l$ if $l<j$.   

For the next result, we define the half space 
\[ H_j := \{z \in \mathbb{C} : \Re(z) \geq z_j - 20 \eta r\}. \]
\begin{proposition} \label{prop:H}
Under the assumptions of Theorem \ref{thm:subspace}, for every $z \in H_j$, 
\[ |z| \geq \sigma_j \ge M. \]
In particular, 
\begin{align}\label{eq:zjbd}
\sigma_j < z_j < \frac{8}{7} \sigma_j. 
\end{align}
\end{proposition} 
\begin{proof}
In view of Lemma \ref{lemma:varphi}, it follows that $\sigma_j^2 = \varphi(z_j) < z_j^2$.
Thus, as $\sigma_j \geq M + 140 \eta r$ by assumption,
\[ |z| \geq \Re(z) \geq z_j - 20 \eta r \geq  \sigma_j - 20 \eta r \geq M+ 100\eta r\ge M \]
for any $z \in H_j$.  

Since $\sigma_j^2 = \varphi(z_j)= (z_j-\tr \mathcal I^{\mathrm d} G(z_j))(z_j-\tr \mathcal I^{\mathrm u} G(z_j))$, 
\begin{equation}\label{eq:diffoftwo}
 z_j^2 - \sigma_j^2 =z_j \tr G(z_j) - \tr \mathcal I^{\mathrm u} G(z_j) \tr \mathcal I^{\mathrm d} G(z_j) \le 2(N+n),
 \end{equation} 
where we invoked Lemma \ref{lemma:resolventnorm} and the fact that $\tr \mathcal I^{\mathrm u} G(z_j) \tr \mathcal I^{\mathrm d} G(z_j) \geq 0$ (see \eqref{eq:traceid} and \eqref{eq:traceid2}) in the last inequality. 
Hence, using that $z_j \geq M$ and $\sigma_j \geq M$, we conclude that
\[ z_j - \sigma_j \le \frac{N+n}{M} \le \frac{1}{4}(\sqrt N + \sqrt n)\le \frac{1}{16}\sigma_j, \]
where the last inequality follows from the assumption that $n + N \geq 32$.  This proves \eqref{eq:zjbd} (where we use a slightly worse but simpler constant).  
\end{proof}

If $\sigma_j$ is sufficiently large, we get a finer estimate for $z_j$ than what is given in \eqref{eq:zjbd}.
\begin{proposition}\label{prop:bigbound}
If $\sigma_j> \frac{1}{2}n^2$, then $|z_j - \sigma_j | \le \frac{8}{n}$.
\end{proposition}
\begin{proof} Following the computation in \eqref{eq:diffoftwo}, together with Lemma \ref{lemma:resolventnorm}, we get
\begin{align*}
|z_j - \sigma_j| (z_j +\sigma_j) =|z_j^2 - \sigma_j^2|  &\le z_j |\tr G(z_j)| + \tr \mathcal I^{\mathrm u} G(z_j) \tr \mathcal I^{\mathrm d} G(z_j)\\
&\le z_j (N+n)\|G(z_j)\| + Nn \|G(z_j)\|^2\\
&\le 2(N+n) +  4\frac{Nn}{z_j^2} \\
&\le 2(N+n) +  4\frac{Nn}{\sigma_j^2},
\end{align*}
 where we used \eqref{eq:zjbd} in the last inequality.  Since $z_j +\sigma_j \ge 2 \sigma_j$, we further get $$|z_j - \sigma_j| \le \frac{N+n}{\sigma_j} + 2\frac{Nn}{\sigma_j^3}, $$
 and the conclusion follows from  $\sigma_j > \frac{1}{2}n^2$ and the supposition $N \leq n$.
\end{proof}

We now complete the proof of Theorem \ref{thm:singularlocation}.  Let $j$ be a fixed index in $\llbracket i_0, r_0 \rrbracket$. We will work in the set $H_j \cap S_{\sigma_j}$, where $S_{\sigma_j}$ is specified in \eqref{eq:resolventset}.  By Corollary 2.14 in \cite{Ipsen}, it follows that 
\begin{equation} \label{eq:rouche}
	\frac{ |f(z) - g(z)|}{|g(z)|} \leq \left( 1 + \varepsilon(z) \right)^{2r} - 1, 
\end{equation} 
where
\[ \eps(z) := \left \| \left( \mathcal D^{-1} - \mathcal U^\T \Phi(z) \mathcal U \right)^{-1} \right \| \left\| \mathcal U^\T (G(z)-\Phi(z)) \mathcal U \right\|. \]

The next result facilitates the comparison of the numbers of zeros of $f$ and $g$ inside a region and will be used repeatedly in the later arguments. 
\begin{lemma}\label{lem:comparezero}
For any circle $\mathcal C\subset \mathbb C$, if $\eps(z) \le \frac{1}{4r}$ for all $z\in \mathcal C$, then the number of zeros of $f$ inside $\mathcal{C}$ is the same as the number of zeros of $g$ inside $\mathcal{C}$.
\end{lemma}
\begin{proof}
Continuing from \eqref{eq:rouche}, we find that
\begin{equation} \label{eq:rouchebnd}
	\frac{ |f(z) - g(z)|}{|g(z)|} \leq \left( 1 + \frac{1}{4r} \right)^{2r} - 1 \leq e^{1/2} - 1 < 1 
\end{equation} 
for each $z \in \mathcal{C}$.  Therefore, by Rouch\'{e}'s theorem, we conclude that the numbers of zeros of $f$ and $g$ inside $\mathcal{C}$ are the same.
\end{proof}

We first bound $\eps(z)$ for $z\in S_{\sigma_j}$. By Proposition \ref{prop:evblock}, 
\begin{align*}
\left \| \left( \mathcal D^{-1} - \mathcal U^\T \Phi(z) \mathcal U \right)^{-1} \right \|= \max_{1\le l \le r} \frac{\sigma_l}{|\sigma_l^2 - \phi_1 \phi_2|} \mathcal Q^{1/2},
\end{align*}
where $$\mathcal Q:=|\phi_1 \phi_2|^2 + \frac{1}{2}\sigma_l^2 (|\phi_1|^2 + |\phi_2|^2) + \frac{1}{2}\sigma_l \left[ 4 |\phi_1 \phi_2|^2 |\phi_1 + \bar{\phi}_2|^2 + \sigma_l^2(|\phi_1|^2 - |\phi_2|^2)^2\right]^{1/2}.$$
Using \eqref{eq:phibd} from Lemma \ref{lemma:resolventnorm}, for $z\in S_{\sigma_j}$, we get 
\begin{align*}
\mathcal Q &\le \left(\frac{9}{8} \right)^4 |z|^4 + \left(\frac{9}{8} \right)^2 \sigma_l^2 |z|^2 + \left(\frac{9}{8} \right)^2\sigma_l  |z|^2 \sqrt{\sigma_l^2 +4\left(\frac{9}{8} \right)^2|z|^2 }\\
&\le \frac{9}{4}|z|^4 + \frac{27}{4}|z|^3 \sigma_l + \frac{9}{2} |z|^2 \sigma_l^2 \\
&\le \frac{9}{4}|z|^2\left(|z|+\frac{3}{2}\sigma_l\right)^2, 
\end{align*}
and thus
\begin{align*}
\left \| \left( \mathcal D^{-1} - \mathcal U^\T \Phi(z) \mathcal U \right)^{-1} \right \| 
&\le\frac{3}{2}|z| \max_{1\le l \le r}  \frac{\sigma_l (|z|+\frac{3}{2}\sigma_l)}{|\sigma_l^2 - \phi_1(z) \phi_2(z)|}.
\end{align*}
Combining this bound with  
the bound in Lemma \ref{lemma:assumpGOE}, we obtain with probability at least $1 - 10(N+n)^{-(K+1)}$
\begin{align}\label{eq:epsbd1}
\eps(z) \le\max_{1\le l \le r}  \frac{3}{2}\frac{\eta}{|z|}\frac{\sigma_l (|z|+\frac{3}{2}\sigma_l)}{|\sigma_l^2 - \phi_1(z) \phi_2(z)|}
\end{align}
for all $z \in S_{\sigma_j}$. 

We now restrict ourselves to values of $z$ contained on a circle $\mathcal{C}_j$ in $H_j \cap S_{\sigma_j}$.  Here we take $\mathcal{C}_j$ to be the circle of radius $20\eta r$ centered at $z_j$ (which by definition is contained in $H_j \cap S_{\sigma_j}$).  

The goal is to show $\eps(z)$ is small for all $z\in \mathcal C_j$. Continuing from \eqref{eq:epsbd1}, it suffices to show $$\frac{3}{2}\frac{\eta}{|z|}\frac{\sigma_l (|z|+\frac{3}{2}\sigma_l)}{|\sigma_l^2 - \phi_1(z) \phi_2(z)|}$$ is small for all $1\le l \le r.$ We split the discussion into two cases: $l=j$ and $l\neq j$.

\emph{Case 1. When $l=j$, }in view of \eqref{eq:lipschitz}, for $z \in \mathcal{C}_j$, we have
\begin{align*}
|\sigma_j^2 - \phi_1(z) \phi_2(z)| &= |\phi_1(z_j) \phi_2(z_j) - \phi_1(z) \phi_2(z)|\\
& \ge \frac{9}{16} |z_j^2 - z^2| = \frac{45}{4}\eta r |z_j + z|\ge \frac{585}{28} \eta r\sigma_j, 
\end{align*}
where, in the last inequality, we used $|z+z_j| =|2z_j +z-z_j|\ge 2z_j - 20 \eta r \ge \frac{13}{7}\sigma_j$ by \eqref{eq:zjbd} and the assumption  $\sigma_j \ge 140 \eta r$.
Similarly, using \eqref{eq:lipschitz} and Proposition \ref{prop:H}, 
\begin{align}\label{ieq:jbd}
|\sigma_j^2 - \phi_1(z) \phi_2(z)| \le \frac{23}{16} |z_j^2 - z^2| \le \frac{1955}{28} \eta r\sigma_j .
\end{align}
Thus, to bound $\varepsilon(z)$, we apply the bound 
\begin{align}\label{ieq:zbd}
\frac{6}{7}\sigma_j \le z_j - 20 \eta r \le |z| \le z_j + 20 \eta r\le \frac{9}{7}\sigma_j
\end{align}
 that follows from \eqref{eq:zjbd} and $\sigma_j \ge 140 \eta r$ to obtain 
\begin{align*}
\frac{3}{2}\frac{\eta}{|z|} \frac{\sigma_j (|z|+\frac{3}{2}\sigma_j)}{|\sigma_j^2 - \phi_1(z) \phi_2(z)|}\le \frac{0.13}{r}<\frac{1}{4r}.
\end{align*}

\emph{Case 2. When $l\neq j$, } note that 
\begin{align*}
|\sigma_l^2 - \phi_1(z) \phi_2(z)| &\ge |\sigma_l^2- \sigma_j^2| -| \sigma_j^2 - \phi_1(z) \phi_2(z)|\\
&\ge |\sigma_j-\sigma_l| (\sigma_j+\sigma_l) - \frac{1955}{28}  \eta r \sigma_j \\
&\ge \frac{169}{560}|\sigma_j-\sigma_l| (\sigma_j+\sigma_l)
\end{align*}
by \eqref{ieq:jbd} and the assumption that $|\sigma_j-\sigma_l| \ge 100  \eta r$. Hence, using \eqref{ieq:zbd}, we get
\begin{align*}
&\frac{3}{2}\frac{\eta}{|z|} \frac{\sigma_l (|z|+\frac{3}{2}\sigma_l)}{|\sigma_l^2 - \phi_1(z) \phi_2(z)|} \le \frac{3}{2} \frac{\eta}{ \frac{6}{7}\sigma_j } \frac{560\sigma_l (\frac{9}{7}\sigma_j+\frac{3}{2}\sigma_l)}{169 |\sigma_j-\sigma_l| (\sigma_j+\sigma_l)} \le 9 \frac{\eta \sigma_l}{\sigma_j |\sigma_j-\sigma_l|}.
\end{align*}
If $\sigma_l \le 1.8 \sigma_j$, then 
\begin{align*}
&\frac{3}{2}\frac{\eta}{|z|} \frac{\sigma_l (|z|+\frac{3}{2}\sigma_l)}{|\sigma_l^2 - \phi_1(z) \phi_2(z)|} \le 9 \frac{1.8\eta}{100 \eta r} < \frac{1}{4r}.
\end{align*}
If $\sigma_l \ge 1.8 \sigma_j$, then $\sigma_l-\sigma_j \ge \frac{4}{9} \sigma_l $ and 
\begin{align*}
&\frac{3}{2}\frac{\eta}{|z|} \frac{\sigma_l (|z|+\frac{3}{2}\sigma_l)}{|\sigma_l^2 - \phi_1(z) \phi_2(z)|} \le 9 \frac{\eta}{140 \eta r \frac{4}{9}} < \frac{1}{4r}.
\end{align*}

Thus, we conclude that 
\begin{equation}\label{eq:epsbd}
 \eps(z) \leq \frac{1}{4 r} 
\end{equation} 
for all $z \in \mathcal{C}_j$. By Lemma \ref{lem:comparezero}, the number of zeros of $f$ inside $\mathcal{C}_j$ is the same as the number of zeros of $g$ inside $\mathcal{C}_j$. Since $g$ has $\alpha_j$ zeros inside $\mathcal{C}_j$, it follows that there exists exactly $\alpha_j$ values of $l_{1}(j)\le \ldots \le l_{\alpha_j}(j)$  such that 
\begin{equation} \label{eq:eigconclusion}
	|\widetilde{\sigma}_{l_s(j)}- z_j| < 20  \eta r \quad \text{for} \quad 1\le s \le \alpha_j. 
\end{equation} 
In particular, this means that for $1\le s \le \alpha_j$
\begin{align*}
| \varphi(\widetilde{\sigma}_{l_s(j)}) - \varphi(z_j)|  \leq \frac{23}{16}|\widetilde{\sigma}_{l_s(j)}^2- z_j^2| &\le \frac{115}{4}  \eta r (\widetilde{\sigma}_{l_s(j)}+z_j ) \le \frac{115}{4}   \eta r \left(\widetilde{\sigma}_{l_s(j)}+\frac{8}{7}\sigma_j \right)
\end{align*}
by Lemma \ref{lemma:varphi} and \eqref{eq:zjbd}.  Since $\varphi(z_j)= \sigma_j^2$, we conclude that
\begin{equation} \label{eq:eigconclusion2}
	|\varphi(\widetilde{\sigma}_{l_s(j)}) - \sigma_j^2| \leq \frac{115}{4}   \eta r \left(\widetilde{\sigma}_{l_s}+\frac{8}{7}\sigma_j \right). 
\end{equation} 
It remains to show that $l_s(j) = j+s-1$ for $1\le s \le \alpha_j$ and $i_0\le j \le r_0$.  We will do so by proving the following claims hold with probability at least $1-30(N+n)^{-(K+1)}$ (see Figure \ref{fig:cycles} for an illustration): 
\begin{figure}[!ht]
 \begin{center}
   \includegraphics[width=8cm]{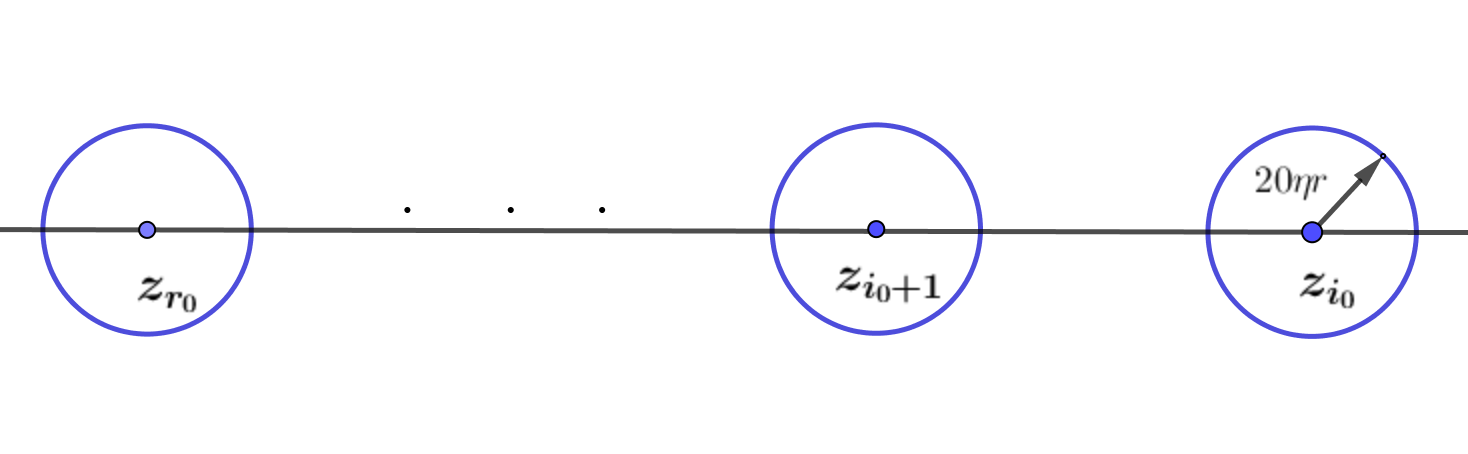}
   \caption{Distinct circles $\mathcal C_j$ with centers $z_j$ on the real line and radius $20\eta r$ for $i_0 \le j \le r_0$.}
    \label{fig:cycles}
 \end{center}
\end{figure}

\noindent\emph{Claim 1.} Distinct circles do not intersect.\\
\noindent\emph{Claim 2.}  $\mathcal A+\mathcal E$ has exactly $i_0-1$ eigenvalues larger than $z_{i_0}+20\eta r$.\\
\noindent\emph{Claim 3.} No eigenvalues of $\mathcal A+\mathcal E$ lie between distinct circles. 

For the moment, let us assume these claims are true. Note that $\widetilde \sigma_{i_0}$ has to lie inside one of the $\mathcal C_j$'s ($i_0 \le j \le r_0$) because it is the largest eigenvalue of $\mathcal A+\mathcal E$ that is no larger than $z_{i_0}+20\eta r$ (due to \emph{Claim 2}) and thus it satisfies $\widetilde \sigma_{i_0} > z_{r_0} - 20\eta r$ by \eqref{eq:eigconclusion}. Since the number of zeros of $g(z)$ located inside $\mathcal C_j$'s for all $i_0 \le j \le r_0$, which is $r_0-i_0+1$, is the same as that of $f(z)$ inside $\mathcal C_j$'s ($i_0 \le j \le r_0$), we have $\widetilde \sigma_{i_0},\ldots,\widetilde \sigma_{r_0}$ lie inside $\mathcal C_j$'s ($i_0 \le j \le r_0$). The conclusion follows by the fact that the number of zeros of $g(z)$ in each $\mathcal C_j$ is the same as that of $f(z)$. 

We start with the proof of the \emph{Claim 1}. For $\sigma_l \neq \sigma_j$, by Lemma \ref{lemma:varphi},
$$|z_l^2 - z_j^2| \ge \frac{16}{23}|\varphi(z_l) - \varphi(z_j)| =\frac{16}{23} |\sigma_l^2 - \sigma_j^2| \ge \frac{16}{23} 100  \eta r(\sigma_l + \sigma_j).$$
Since $|z_l^2 - z_j^2| = (z_l+z_j) |z_l-z_j| \le \frac{8}{7}(\sigma_l + \sigma_j) |z_l-z_j|$ by Proposition \ref{prop:H}, we have
\begin{equation}\label{eq:zljdiff}
|z_l - z_j |\ge \frac{1400}{23}   \eta r, 
\end{equation}
and thus
$$\dist(\mathcal C_j, \mathcal C_l) \ge |z_l - z_j| - 40  \eta r > 20  \eta r. $$  

\medskip

Next, we prove \emph{Claim 2}. We split the proof into two cases: $i_0=1$ and $i_0>1$.

\noindent\emph{Case 1: $i_0=1$.} We prove that no eigenvalues of $\mathcal A+\mathcal E$ are larger than $z_1+20\eta r$. We now take $\mathcal{C}_0$ to be any circle with radius $20  \eta r$ centered at a point $z_0>z_1+20\eta r$ on the real line inside the region $H_1 \cap \hat{S}_{\sigma_1}$ such that $\dist(z_1,\mathcal{C}_0) \geq 20  \eta r$.  Here 
\begin{equation}\label{eq:modifiedS}
\hat{S}_{\sigma_1} : = \{w \in \mathbb{C} : | \Im(w) | \leq 20 \eta r, 4(\sqrt N + \sqrt n)+120 \eta r \leq \Re(w) \leq \frac{3}{2}\sigma_1 +  20 \eta r \}
\end{equation}
 is a slight modification of the set ${S}_{\sigma}$ in \eqref{eq:resolventset}. Note that $\widetilde{\sigma}_1 \in \hat{S}_{\sigma_1}$: the upper bound $\widetilde{\sigma}_1\le \frac{3}{2}\sigma_1$ follows from the Weyl's inequality and the supposition $\|\mathcal E\|\le \frac{1}{2}\sigma_1$; the lower bound is because it is the largest eigenvalue and $\widetilde{\sigma}_1 \ge z_j -20\eta r \ge \sigma_j -20\eta r$ from \eqref{eq:eigconclusion}. An inspection of the proof of Lemma \ref{lemma:assumpGOE} reveals that the conclusion of Lemma \ref{lemma:assumpGOE} also holds on the set $\hat{S}_{\sigma_1}$. Hence, the bound \eqref{eq:epsbd1} also holds for $z\in \hat{S}_{\sigma_1}$. 
 We show
\[ \eps(z) < \frac{1}{4r} \]
for all $z \in \mathcal{C}_0$.  The proof is similar to the proof of \eqref{eq:epsbd} and we sketch it here. For any $z \in \mathcal{C}_0$, from $|z-z_0|=20\eta r$, $|z-z_1| \ge 20 \eta r$ and $z_0-z_1 > 40 \eta r$, we obtain 
$|z| \le z_0 + 20 \eta r$ and $$|z| \ge z_0 - 20 \eta r \ge z_1 + 20\eta r > \sigma_1 + 20\eta r > \sigma_1.$$ Again, by Lemma \ref{lemma:varphi}, we see for any $1\le l \le r$, 
\begin{align}\label{eq:onelastineq}
|\sigma_l^2 - \phi_1(z)\phi_2(z)| &= |\varphi(z_l) -\varphi(z)|\nonumber \\
&\ge \frac{9}{16} |z_l^2 - z^2| \nonumber\\
&\ge  \frac{9}{16} (z_l + \Re(z))(\Re(z) -z_l)\nonumber\\
&\ge \frac{9}{16}(\sigma_l + z_0 -20\eta r)(z_0-z_l -20\eta r) \nonumber\\
&\ge \frac{9}{16}(\sigma_l + z_0 -20\eta r) 20 \eta r.
\end{align}
Plugging these estimates back into \eqref{eq:epsbd1}, we see
\begin{align*}
\eps(z) \le  \max_{1\le l \le r}\frac{3}{2} \frac{\eta\sigma_l}{\sigma_1} \frac{z_0 + 20 \eta r + \frac{3}{2} \sigma_l}{\frac{9}{16}(\sigma_l + z_0 -20\eta r)20 \eta r} < \frac{1}{4r},
\end{align*}
 where we used the bound $z_0 + 20 \eta r + \frac{3}{2} \sigma_l \le \frac{3}{2}(\sigma_l + z_0 -20\eta r)$ in the last inequality. 

By Lemma \ref{lem:comparezero}, $f$ has the same number of zeros inside $\mathcal{C}_0$ as $g$.  As $g$ has no zeros inside $\mathcal{C}_0$\footnote{This follows from Lemma \ref{lemma:varphi} and the fact that $\Im(\varphi(z)) \neq 0$ whenever $\Im z \neq 0$ for all $|z| > M$.}, $\mathcal A+\mathcal E$ has no eigenvalues inside $\mathcal{C}_0$.  Since the circle $\mathcal{C}_0$ was arbitrarily chosen inside this region, we conclude that $\mathcal A+\mathcal E$ has no eigenvalues larger than $z_1+20\eta r$.

\medskip

\noindent\emph{Case 2: $i_0>1$.} We work on the event 
\begin{equation}\label{ineq:big}
\max_{l\in \llbracket 1,r_0 \rrbracket; \sigma_l>n^2/2}|\widetilde \sigma_l - \sigma_l |\le \eta r.
\end{equation}
By Lemma \ref{lem:svloc-large}, this event holds with probability at least $1-(N+n)^{-2r^4(K+8)}$. 

Note that $\sigma_{i_0-1}>n^2$.  Combining \eqref{ineq:big}, Proposition \ref{prop:bigbound} and $z_{i_0-1} - z_{i_0}\ge \frac{1400}{23}   \eta r > 60\eta r$ from \eqref{eq:zljdiff}, we get 
\begin{align*}
\widetilde \sigma_{i_0-1} \ge \sigma_{i_0-1} - \eta r \ge z_{i_0-1} - \frac{8}{n} - \eta r \ge z_{i_0} + 59\eta r- \frac{8}{n} > z_{i_0} + 20\eta r.
\end{align*}
Hence, $\mathcal A+\mathcal E$ has at least $i_0-1$ eigenvalues larger than $z_{i_0}+20\eta r$. 

We first consider $\sigma_{i_0} >\frac{1}{2}n^2$.  It follows from \eqref{ineq:big} and Proposition \ref{prop:bigbound} that$$\widetilde \sigma_{i_0} \le \sigma_{i_0}+\eta r \le z_{i_0} + \frac{8}{n} +\eta r \le z_{i_0} + 20\eta r.$$ This shows that $\mathcal A+\mathcal E$ has exactly $i_0-1$ eigenvalues larger than $z_{i_0}+20\eta r$. 

Now consider $\sigma_{i_0} \le \frac{1}{2}n^2$. By Weyl's inequality, $\widetilde \sigma_{i_0} \le \sigma_{i_0} + \|E\| < \frac{3}{2}\sigma_{i_0}$. If $\frac{3}{2}\sigma_{i_0} \le z_{i_0} + 20\eta r$,   the proof is already done. Now we assume $\frac{3}{2}\sigma_{i_0} > z_{i_0} + 20\eta r$. If $\frac{3}{2}\sigma_{i_0} - (z_{i_0} + 20\eta r)< 40\eta r$, following \eqref{eq:zjbd},  we have $\frac{8}{7}\sigma_{i_0} > z_{i_0} > \frac{3}{2}\sigma_{i_0} - 60\eta r$ and thus $\sigma_{i_0} < 168 \eta r$. From the assumption $\sigma_{i_0} \ge 4(\sqrt N + \sqrt n) + 140 \eta r$, we further have $4(\sqrt N + \sqrt n) \le 28 \eta r$ and hence $\|E\|\le 2(\sqrt N + \sqrt n) < 14 \eta r$. It follows from Weyl's inequality and \eqref{eq:zjbd} that $\widetilde \sigma_{i_0}\le \sigma_{i_0} + \|E\| \le z_{i_0} + 14\eta r < z_{i_0} + 20\eta r$. 

It remains to consider the case when $\frac{3}{2}\sigma_{i_0} - (z_{i_0} + 20\eta r)\ge 40\eta r$. To prove $\widetilde \sigma_{i_0} \le z_{i_0} + 20\eta r$, we show that $f$ has no zeros on the interval $(z_{i_0}+20\eta r, \frac{3}{2}\sigma_{i_0})$. The proof is similar to the proof of \emph{Case 1} when $i_0=1$. We only mention the differences. Define $\hat{S}_{\sigma_{i_0}}$ as in \eqref{eq:modifiedS} and the bound \eqref{eq:epsbd1} also holds for $z\in \hat{S}_{\sigma_{i_0}}$. The goal is to show $\varepsilon(z)<1/4r$ for all $z\in\mathcal C_0$, where $\mathcal C_0$ is any circle with radius $20  \eta r$ centered at a point $z_0 \in (z_{i_0}+20\eta r, \frac{3}{2}\sigma_{i_0})$  inside the region $H_{i_0} \cap \hat{S}_{\sigma_{i_0}}$ such that $\dist(z_{0},z_{i_0}+20\eta r) \geq 20  \eta r$ and $\dist(z_{0},\frac{3}{2}\sigma_{i_0}) \geq 20  \eta r$. If so, by Lemma \ref{lem:comparezero}, $f$ has the same number of zeros inside $\mathcal C_0$ as $g$.  Note that $g$ has no zeros inside $\mathcal C_0$ since $\Im(\varphi(z)) \neq 0$ whenever $\Im z \neq 0$ for all $|z| > M$ and $z_{i_0-1} \ge \sigma_{i_0-1} - \frac{8}{n} > n^2 - \frac{8}{n} > \frac{3}{2}\sigma_{i_0}$ by Proposition \ref{prop:bigbound}. Since $\mathcal{C}_0$ was arbitrarily chosen, $\mathcal A + \mathcal E$ has no eigenvalues on $(z_{i_0}+20\eta r, \frac{3}{2}\sigma_{i_0})$. 

It remains to bound $\eps(z)$ from \eqref{eq:epsbd1}. The same arguments as those in \emph{Case 1} yield that 
$$\max_{i_0\le l \le r}  \frac{3}{2}\frac{\eta}{|z|}\frac{\sigma_l (|z|+\frac{3}{2}\sigma_l)}{|\sigma_l^2 - \phi_1(z) \phi_2(z)|} < \frac{1}{4r}$$ for any $z\in \mathcal C_0$. 
We only need to control
\begin{align*}
\max_{1\le l \le i_0-1}  \frac{3}{2}\frac{\eta}{|z|}\frac{\sigma_l (|z|+\frac{3}{2}\sigma_l)}{|\sigma_l^2 - \phi_1(z) \phi_2(z)|}.
\end{align*}For any $z\in \mathcal C_0,$ $|z| \ge z_0 - 20\eta r \ge z_{i_0} \ge \sigma_{i_0}$ and $|z| \le z_0 + 20 \eta r \le \frac{3}{2} \sigma_{i_0} + 40\eta r$. For any $1\le l \le i_0-1$,  using similar computation from \eqref{eq:onelastineq}, we get
\begin{align*}
|\sigma_l^2 - \phi_1(z)\phi_2(z)| &\ge \frac{9}{16}(\sigma_l + z_0 -20\eta r)(z_l-z_0 -20\eta r).
\end{align*}
Note that $z_0 \ge z_{i_0} + 40 \eta r \ge \sigma_{i_0} + 40 \eta r$. Hence, $\sigma_l + z_0 -20\eta r \ge \sigma_l + \sigma_{i_0} +20\eta r$. From $\sigma_l  \geq n^2$, we see $\sigma_{i_0}\leq \frac{1}{2}n^2 \leq \frac{1}{2}\sigma_l$. This, together with \eqref{eq:zjbd} and $z_0\le \frac{3}{2} \sigma_{i_0} - 20\eta r$, implies that $$z_l-z_0 -20\eta r \ge \sigma_l -\frac{3}{2} \sigma_{i_0} \ge \sigma_l - \frac{3}{4} \sigma_l = \frac{1}{4} \sigma_l.$$
Hence, 
\begin{align*}
\max_{1\le l \le i_0-1}  \frac{3}{2}\frac{\eta}{|z|}\frac{\sigma_l (|z|+\frac{3}{2}\sigma_l)}{|\sigma_l^2 - \phi_1(z) \phi_2(z)|}&\le \frac{3}{2}\frac{\eta}{\sigma_{i_0}} \max_{1\le l \le i_0-1} \frac{\sigma_l (\frac{3}{2} \sigma_{i_0} + 40 \eta r + \frac{3}{2}\sigma_l)}{\frac{9}{16} (\sigma_l + \sigma_{i_0} +20\eta r)\frac{1}{4} \sigma_l }\\
& \le  \frac{16\eta}{\sigma_{i_0}} \\
&< \frac{1}{4r}
\end{align*}using the assumption $\sigma_{i_0}> 140\eta r$.
Therefore, $\eps(z) < {1}/{4r}$ for all $z\in \mathcal C_0.$

\medskip

The proof of \emph{Claim 3} is similar to the previous argument. Let $\mathcal C_{j_1}, \mathcal C_{j_2}$ be two distinct adjacent circles. Note that from the proof of \emph{Claim 1}, $\dist(\mathcal C_{j_1}, \mathcal C_{j_2})  > 20  \eta r.$ We show that $\mathcal A + \mathcal E$ has no eigenvalues lying on the real line between $\mathcal C_{j_1}$ and $\mathcal C_{j_2}$. Take any point $x$ on the real line between the two circles so that $\mathcal C_x$, the circle centered at $x$ with radius $10\eta r$, is inside the region $H_{j_1}\cap S_{\sigma_{j_1}}$ or $H_{j_2}\cap S_{\sigma_{j_2}}$, where $\dist(x, \mathcal C_{j_1}) > 10\eta r$ and $\dist(x, \mathcal C_{j_2}) > 10\eta r$. Then using similar calculations as in the proof of \emph{Claim 2}, it suffices to show that $\eps(z)<\frac{1}{4r}$. The remaining arguments are similar to those in the proof of \emph{Claim 2}; we omit the details.

\section{Proof of Lemma \ref{lemma:assumpGOE}} \label{sec:lem:GOE}


In this section, we present the proof of Lemma \ref{lemma:assumpGOE}. We first show that $G(z)$ is close to $\Phi(z)$ for any fixed $z\in \mathbb C$ satisfying $|z| \ge 4(\sqrt{N}+ \sqrt{n})$. Then we extend this result to any $z\in S_{\sigma_j}$ by a net argument. Throughout the proof, we will sometimes write $G$ instead of $G(z)$ for convenience.  The proof presented here takes advantage of the fact that the entries of $E$ are jointly independent standard Gaussian random variables; this assumption greatly simplifies the forthcoming calculations, although the method can be extended to other distributions.   In particular, when the entries are non-Gaussian, one also needs to estimate the off-diagonal entries of $G$.  This can be accomplished by modifying some of the techniques from \cite{BEKYY14}. 

We begin with the following notation. 
\begin{definition}[Minors]
For $I \subset \llbracket 1,N+n \rrbracket$, we define $\mathcal E^{(I)}$ by
\[ \mathcal E^{(I)}_{st} := \left\{
        \begin{array}{ll}
            \mathcal E_{st} & \text{ if } s,t \not\in I, \\
            0 & \text{ otherwise}.  
        \end{array}
    \right. \]
We define the resolvent of $\mathcal E^{(I)}$ by
\[ G_{st}^{(I)}(z) := \left\{
        \begin{array}{ll}
            (z - \mathcal E^{(I)})^{-1}_{st}  & \text{ if } s,t \not\in I, \\
            0 & \text{ otherwise,}
        \end{array}
    \right. \]
whenever the inverse is defined.  We use the summation notation 
\[ \sum_{s}^{(I)} := \sum_{s \in \llbracket 1,N+n \rrbracket : s \not\in I}. \]
When $I = \{a\}$, we abbreviate $(\{a\})$ by $(a)$ in the above definitions.  
\end{definition} 

\begin{lemma}[Resolvent identities] \label{lemma:resolventident}
For any $k \in \llbracket 1,N+n \rrbracket$ and for $|z| > \|\mathcal E\|$
\[ G_{kk}(z) = \frac{1}{z - \mathcal E_{kk} - \sum_{s,t}^{(k)} \mathcal E_{sk} G_{st}^{(k)}(z) \mathcal E_{tk} }. \]
Moreover, for $i \neq j$ and any $|z| > \|\mathcal E\|$, 
\[ G_{ij}(z) = - G_{ii}(z) \sum_k^{(i)} \mathcal E_{ik} G^{(i)}_{kj}(z). \]
\end{lemma}
\begin{proof}
The formula for the diagonal entries follows from the Schur complement (see \cite[Theorem A.4]{BS}).  The off-diagonal entries can be computed in a similar way (see \cite[Lemma 4.2]{EYY} or \cite[Lemma 6.10]{EKYY12}).  
\end{proof}

Next, we show that the resolvent matrix $G(z)=(z-\mathcal E)^{-1}$ is well approximated by the diagonal matrix $\Phi(z)$ for any fixed $z\in \mathbb{C}$ with sufficiently large modulus.   
\begin{lemma}\label{lemma:assumpfixed}Let $K>0$ be any constant and assume $(\sqrt N + \sqrt n)^2 \ge 64 (K+2)\log(N+n)$. For any $z \in \mathbb{C}$ with $|z| \ge 4(\sqrt{N}+ \sqrt{n})$, 
\[ \left\| \mathcal U^T \left(G(z)-\Phi(z) \right) \mathcal U  \right\| \leq  48 r \frac{\sqrt{(K+1)\log (N+n)}}{|z|^2} \] 
with probability at least $1 - (N+n)^{-K}$.  
\end{lemma}
\begin{proof}
By the rotational invariance of $E$, it suffices to assume that $U$ is the matrix with columns $e_1, \ldots, e_r$, where $e_1, \ldots, e_N$ is the canonical basis in $\mathbb{R}^N$, and the columns of $V$ are given by $f_1, \ldots, f_r$, where $f_1, \ldots, f_n$ is the canonical basis in $\mathbb{R}^n$.   We use the shorthand notations $\bar i = i-r$ and $\bar j = j-r$ if $r+1\le i, j \le 2r$. Thus,$$2(\mathcal U^T G \mathcal U)_{ij} =\begin{cases} 
 G_{ij} + G_{i,N+j} + G_{N+i, j} + G_{N+i, N+j} & \text{if } i,j \in \llbracket 1, r\rrbracket;\\
 G_{i\bar j} - G_{i,N+\bar j} + G_{N+i, \bar j} - G_{N+i, N+\bar j} & \text{if } i \in \llbracket 1, r\rrbracket, j \in \llbracket r+1, 2r\rrbracket;\\
 G_{\bar i j} + G_{\bar i,N+j} - G_{N+\bar i, j} - G_{N+\bar i, N+j} & \text{if } i \in \llbracket r+1, 2r\rrbracket, j \in \llbracket 1, r\rrbracket;\\
 G_{\bar i \bar j} - G_{\bar i,N+\bar j} - G_{N+\bar i, \bar j} + G_{N+\bar i, N+\bar j} & \text{if }  i,j \in \llbracket r+1, 2r\rrbracket.
  \end{cases} 
 $$
Fix $z \in \mathbb{C}$ with $|z| \ge 4 (\sqrt N +\sqrt{n})$. Denote the set $$\mathcal S_r := \{ (i,j) : i, j \in \llbracket 1, r\rrbracket \cup \llbracket N+1, N+r\rrbracket\}. $$ Since $\|\mathcal M \| \le 2r \|\mathcal M\|_{\max} := 2r \max_{1\le i,j\le 2r} |\mathcal M_{ij}|$ for any $2r\times 2r$ matrix $\mathcal M$,  we first get
\begin{equation}\label{eq:opbd}
\left\| \mathcal U^T \left(G(z)-\Phi(z) \right) \mathcal U  \right\| \leq 2r \max_{1\le i, j \le 2r} \left| (\mathcal U^T G(z) \mathcal U)_{ij} - (\mathcal U^T \Phi(z) \mathcal U)_{ij} \right|.
\end{equation}
In order to prove Lemma \ref{lemma:assumpfixed},  we claim that it suffices to show that 
\begin{equation}\label{eq:maxbd}
\max_{(i,j) \in \mathcal S_r} \left| G_{ij}(z) - \Phi_{ij}(z) \right| \leq  12\frac{ \sqrt{(K+1)\log (N+n)}}{|z|^2}. 
\end{equation}
with probability at least $1-10(N+n)^{-K}$.

Let us assume for the moment that \eqref{eq:maxbd} holds. For any $(i,j) \in \llbracket 1, r\rrbracket$, in view of \eqref{eq:productPhi} and the definition of $\Phi(z)$ in \eqref{eq:Phi}, we find that
\begin{align*}
&2r \left| (\mathcal U^T G(z) \mathcal U)_{ij} - (\mathcal U^T \Phi(z) \mathcal U)_{ij} \right| \\
&=r  \left|G_{ij} + G_{i,N+j} + G_{N+i, j} + G_{N+i, N+j} - \frac{1}{\phi_1(z)}\delta_{ij} - \frac{1}{\phi_2(z)}\delta_{ij} \right|\\
&\le r |G_{ij}(z)-\Phi_{ij}(z)| + r |G_{i,N+j}(z)-\Phi_{i,N+j}(z)|\\
&\quad+r |G_{N+i,j}(z)-\Phi_{N+i,j}(z)|+r |G_{N+i,N+j}(z)-\Phi_{N+i,N+j}(z)|\\
&\le  48r \frac{\sqrt{(K+1)\log (N+n)}}{|z|^2}.
\end{align*}
Similar discussion applies to every $(i,j) \in \llbracket 1, 2r\rrbracket$ and the details are omitted. Hence, 
$$2r \max_{1\le i, j \le 2r} \left| (\mathcal U^T G(z) \mathcal U)_{ij} - (\mathcal U^T \Phi(z) \mathcal U)_{ij} \right| \le 48 r \frac{\sqrt{(K+1)\log (N+n)}}{|z|^2}$$ and the conclusion of Lemma \ref{lemma:assumpfixed} follows from \eqref{eq:opbd}.

Now we turn to the proof of \eqref{eq:maxbd}. For the remainder of the proof, we work on the event where $\|\mathcal E\|=\|E \| \leq 2 (\sqrt N +\sqrt{n})$; recall that Lemma \ref{lemma:norm} shows this event holds with probability at least $1 - 2 e^{-(\sqrt{N}+\sqrt{n})^2/2}\ge 1 - 2(N+n)^{-32(K+2)}$ since $\frac{(\sqrt N+\sqrt n)^2}{\log(N+n)}>64(K+2)$.  We start by controlling the diagonal entries of $G(z)$.  By Lemma \ref{lemma:resolventident}, for $k \in \llbracket 1, N+n \rrbracket$, 
\[ G_{kk}(z)  = \frac{1}{z - \mathcal E_{kk} - \sum_{ s,t}^{(k)} \mathcal E_{sk} G_{st}^{(k)}(z) \mathcal E_{tk} }, \]
and thus, for $1\le k \le r$, by the block definition of $\mathcal E$ and the expression of $\phi_1(z)$ in \eqref{eq:phi12} and \eqref{eq:defphi}, we have
\begin{align} \label{eq:diagonal}
	\left| G_{kk}(z) - \frac{1}{\phi_1(z)}  \right| &= \left| \frac{1}{z - \sum_{1\le i,j\le n}^{(k)} E_{ki} G_{N+i,N+j}^{(k)}(z) E_{kj}}- \frac{1}{\phi_1(z)} \right| \nonumber\\
	&\le \frac{ |\sum_{1\le i,j\le n}^{(k)} E_{ki} G_{N+i,N+j}^{(k)}(z) E_{kj}  - \sum_{t \in \llbracket N+1,N+n \rrbracket} G_{tt}(z)|}{ |z - \sum_{1\le i,j\le n}^{(k)} E_{ki} G_{N+i,N+j}^{(k)}(z) E_{kj}| |z - \tr \mathcal I^\mathrm{d} G(z)| }. 
\end{align} 
We now turn to bounding the right-hand side of \eqref{eq:diagonal}. We start with obtaining an upper bound for the numerator of this term. By the triangle inequality,
\begin{align}\label{eq:wholebd}
&\Prob \left( \left|\sum_{1\le i,j\le n}^{(k)} E_{ki} G_{N+i,N+j}^{(k)}(z) E_{kj}  - \sum_{t \in \llbracket N+1,N+n \rrbracket} G_{tt}^{(k)}(z)\right|  \ge t \right)\\
&\le \Prob \left( \left|\sum_{1\le i,j\le n}^{(k)} E_{ki} \Re G_{N+i,N+j}^{(k)}(z) E_{kj}  - \sum_{t \in \llbracket N+1,N+n \rrbracket} \Re G_{tt}^{(k)}(z)\right|  \ge t/2 \right) \label{eq:realpart}\\
& +\Prob \left( \left|\sum_{1\le i,j\le n}^{(k)} E_{ki} \Im G_{N+i,N+j}^{(k)}(z) E_{kj}  - \sum_{t \in \llbracket N+1,N+n \rrbracket} \Im G_{tt}^{(k)}(z)\right|  \ge t/2 \right)\nonumber.
\end{align}
Since the $k$-th row of $E$ is independent of $G^{(k)}$, we condition on $G^{(k)}$ in the following estimates.  We start with the term in \eqref{eq:realpart}. For notational convenience, denote $\mathsf X:=\left(\Re G_{N+i,N+j}^{(k)}(z)\right)_{i,j\neq k}$ and $\mathsf{g}^\T$ the $k$th row of $E$ with the $k$th entry removed. By assumption, $\mathsf{g}$ is a standard Gaussian vector in $\mathbb R^{n-1}$. Rewrite $$\sum_{1\le i,j\le n}^{(k)} E_{ki} \Re G_{N+i,N+j}^{(k)}(z) E_{kj} =\mathsf{g}^\T \mathsf X \mathsf{g}$$ and $\sum_{t \in \llbracket N+1,N+n \rrbracket} \Re G_{tt}^{(k)}(z)=\E \mathsf{g}^\T \mathsf X \mathsf{g}$. Assume the singular value decomposition of $\mathsf X$ is given by $\mathsf X=O_1 \Sigma O_2$ where $\Sigma=\diag(s_1,\cdots,s_{n-1})$ and $O_1, O_2$ are orthogonal matrices. Due to the rotation invariance property of Gaussian vectors, $\mathsf{g}^\T \mathsf X \mathsf{g} \sim \mathsf{g}^\T \Sigma \mathsf{g}$ and $\E\mathsf{g}^\T \mathsf X \mathsf{g} =\E \mathsf{g}^\T \Sigma \mathsf{g}=\sum_{i=1}^{n-1} s_i$. To bound  \eqref{eq:realpart}, it is equivalent to bound $\Prob(|\mathsf{g}^\T \Sigma \mathsf{g}- \E \mathsf{g}^\T \Sigma \mathsf{g}|\ge t/2)$. In order to apply Lemma \ref{lem:Bernstein}, we can verify that $\mathsf{g}^\T \Sigma \mathsf{g} = \sum_{i=1}^{n-1} s_i g_i^2$ is sub-exponential with parameters $(4\| \mathsf X\|_F^2, 4 \|\mathsf X\|)$. Indeed, by independence and \eqref{eq:chi1sub}, 
$$\E e^{\lambda(\mathsf{g}^\T \Sigma \mathsf{g} - \E \mathsf{g}^\T \Sigma \mathsf{g})} = \E e^{\sum_{i=1}^{n-1} \lambda(s_i-1) g_i^2} =\prod_{i=1}^{n-1} \E e^{\lambda s_i (g_i^2-1)} \le \prod_{i=1}^{n-1} e^{\frac{4s_i^2 \lambda^2}{2} }=e^{\frac{4\|\mathsf X\|_F^2}{2}}$$ for all $|\lambda|<\frac{1}{4\max_i s_i} =\frac{1}{4\|\mathsf X\|}.$
 It follows from Lemma \ref{lem:Bernstein} that
\begin{align}\label{eq:bdRe}
\left|\sum_{1\le i,j\le n}^{(k)} E_{ki} \Re G_{N+i,N+j}^{(k)}(z) E_{kj}  - \sum_{t \in \llbracket N+1,N+n \rrbracket} \Re G_{tt}^{(k)}(z)\right|  \le t 
\end{align}
with probability at least $1-2 \exp\left( -\frac{1}{32} \min\left\{t^2 |z|^2/n, 2t|z| \right\}\right)$
by combining the following estimates 
\[ \| \mathsf X\| \le \left\|\left(G_{N+i,N+j}^{(k)}(z)\right)_{1\le i,j\le n} \right\| \leq \| G^{(k)}(z) \|\leq \frac{2}{|z|}  \]
and
\[ \|\mathsf X\|_F^2 \le \left\|\left(G_{N+i,N+j}^{(k)}(z)\right)_{1\le i,j\le n} \right\|_F^2 \leq n \left\|\left(G_{N+i,N+j}^{(k)}(z)\right)_{1\le i,j\le n} \right\|^2\le \frac{4n}{|z|^2} \]
due to Lemma \ref{lemma:resolventnorm}.  Since $|z|\ge 4(\sqrt N + \sqrt n)$, by selecting $t=\sqrt{2(K+1)\log(N+n)}$ in \eqref{eq:bdRe}, we find that 
\begin{equation*} 
	\left|\sum_{1\le i,j\le n}^{(k)} E_{ki} \Re G_{N+i,N+j}^{(k)}(z) E_{kj}  - \sum_{t \in \llbracket N+1,N+n \rrbracket} \Re G_{tt}^{(k)}(z)\right|  \leq \sqrt{2(K+1)\log(N+n)}
\end{equation*} 
with probability at least $1 - 2(N+n)^{-(K+1)}$. To be more specific, to get this probability bound, we need the following discussion. When $t|z| \le 2n$ and thus $\min\left\{t^2 |z|^2/n, 2t|z| \right\}={t^2|z|^2}/{n}$, the probability bound in \eqref{eq:bdRe} is at least $1-2(N+n)^{-(K+1)}$ since $t|z| \ge 4\sqrt{2(K+1)\log (N+n)}(\sqrt N + \sqrt n)$. When $t|z| > 4n$ and $\min\left\{t^2 |z|^2/n, 2t|z| \right\}=2t|z| >2n$, we obviously still have the probability bound in \eqref{eq:bdRe} is at least $1-2\exp(-n/8) > 1- 2(N+n)^{-2(K+2)}$ by the suppositions $(\sqrt N + \sqrt n)^2 > 64 (K+2) \log (N+n)$ and $n\ge N$.

Likewise, one also has$$\left|\sum_{1\le i,j\le n}^{(k)} E_{ki} \Im G_{N+i,N+j}^{(k)}(z) E_{kj}  - \sum_{t \in \llbracket N+1,N+n \rrbracket} \Im G_{tt}^{(k)}(z)\right|  \leq \sqrt{2(K+1)\log(N+n)}$$ with probability at least $1 - 2(N+n)^{-(K+1)}$. Inserting the above estimates back into \eqref{eq:wholebd}, we find that
\begin{equation} \label{eq:HW1}
\left|\sum_{1\le i,j\le n}^{(k)} E_{ki} G_{N+i,N+j}^{(k)}(z) E_{kj}  - \sum_{t \in \llbracket N+1,N+n \rrbracket} G_{tt}^{(k)}(z)\right| \le 2\sqrt{2(K+1)\log(N+n)}
\end{equation} 
with probability at least $1 - 4(N+n)^{-(K+1)}$.

Next, we show that the difference between $\sum_{t \in \llbracket N+1,N+n \rrbracket} G_{tt}^{(k)}(z)$ and $\sum_{t \in \llbracket N+1,N+n \rrbracket} G_{tt}(z)$ is quite small. Thus, combining \eqref{eq:HW1}, we get an upper bound for the term in the numerator of the right-hand side of \eqref{eq:diagonal}. Rewrite
\[ \left| \sum_{t \in \llbracket N+1,N+n \rrbracket} G_{tt}^{(k)}(z) -\sum_{t \in \llbracket N+1,N+n \rrbracket} G_{tt}(z)\right| = \left|\tr \mathcal I^\mathrm{d}(G^{(k)}(z)-G(z)) \right|. \]
By the resolvent identity \eqref{eq:resolvent}, the above term is written as
\[ \left|\tr \mathcal I^\mathrm{d}(G^{(k)}(z)-G(z)) \right| = \left| \tr \mathcal I^\mathrm{d} G(z) (\mathcal E - {\mathcal E}^{(k)}) {G}^{(k)}(z)  \right|. \]
Since $\mathcal E - {\mathcal E}^{(k)}$ has at most rank $2$, we conclude that
\begin{align*}
 \left| \tr \mathcal I^\mathrm{d} G(z) (\mathcal E - {\mathcal E}^{(k)}) {G}^{(k)}(z)  \right| &\leq 2 \| G(z) (\mathcal E - {\mathcal E}^{(k)}) {G}^{(k)}(z) \| \\
 &\leq 2 \| G(z)\| (\|\mathcal E\| + \|{\mathcal E}^{(k)}\|) \|{G}^{(k)}(z) \| \\
 &\leq \frac{16\|E\|}{|z|^2} \\
 & \leq  \frac{8}{|z|} 
\end{align*}
since $|z| \ge 4(\sqrt N+ \sqrt{n})$ and $\|E\| \leq 2(\sqrt N+ \sqrt{n})$.  Thus, we have shown that
\[ \left| \sum_{t \in \llbracket N+1,N+n \rrbracket} G_{tt}^{(k)}(z) -\sum_{t \in \llbracket N+1,N+n \rrbracket} G_{tt}(z)\right| \leq \frac{8}{|z|} \le \frac{2}{\sqrt N + \sqrt n}. \]
  Returning to \eqref{eq:HW1}, we see that
\begin{equation} \label{eq:HW2}
	\left|\sum_{1\le i,j\le n}^{(k)} E_{ki} G_{N+i,N+j}^{(k)}(z) E_{kj}  - \sum_{t \in \llbracket N+1,N+n \rrbracket} G_{tt}(z)\right| \leq 4 \sqrt{(K+1)\log (N+n) }
\end{equation} 
with probability at least $1 - 4(N+n)^{-(K+1)}$.  

To finish the estimate of \eqref{eq:diagonal}, we provide a lower bound for the denominator of the right-hand side of \eqref{eq:diagonal}. Note from \eqref{eq:bdid} that
\begin{align*}
 \left|\sum_{t \in \llbracket N+1,N+n \rrbracket} G_{tt}(z)\right|= |\tr \mathcal I^d G(z)  | \le \frac{|z|}{8}.
\end{align*}
Combined with \eqref{eq:HW2} and the fact that$$4 \sqrt{(K+2)\log (N+n)} \le \frac{1}{2}(\sqrt N + \sqrt n) \le \frac{|z|}{8}$$ since $(\sqrt N + \sqrt n)^2 \ge 64 (K+2)\log(N+n)$ by supposition and $|z|\ge 4(\sqrt N + \sqrt n)$, we arrive at 
$$\left|\sum_{1\le i,j\le n}^{(k)} E_{ki} G_{N+i,N+j}^{(k)}(z) E_{kj} \right| \le \frac{|z|}{4}.$$Hence, by triangle inequality, 
\begin{align*}
 |z - \sum_{1\le i,j\le n}^{(k)} E_{ki} G_{N+i,N+j}^{(k)}(z) E_{kj}| |z - \tr \mathcal I^\mathrm{d} G(z)| \ge \frac{21}{32}|z|^2.
\end{align*}
Together with \eqref{eq:HW2}, inserting the above estimates into \eqref{eq:diagonal} yields
\[ \left| G_{kk}(z) - \frac{1}{\phi_1(z)}  \right|  \leq \frac{128}{21}\frac{ \sqrt{(K+1)\log (N+n)}}{|z|^2} \]
with probability at least $1 - 4(N+n)^{-(K+1)}$ for $1\le k \le r$. Analogously, one also has
\[ \left| G_{kk}(z) - \frac{1}{\phi_2(z)}  \right|  \leq \frac{128}{21}\frac{ \sqrt{(K+1)\log (N+n)}}{|z|^2}  \]
with probability at least $1 - 4(N+n)^{-(K+1)}$ for $N+1\le k \le N+r$.
Union bounding over all $k \in \llbracket 1, r\rrbracket \cup \llbracket N+1, N+r\rrbracket$ completes the proof for the diagonal entries.  

For the off-diagonal entries, by Lemma \ref{lemma:resolventident}, we have
\[ |G_{ij}(z)| \leq \frac{2}{|z|} \left| \sum_k^{(i)} \mathcal E_{ik} G^{(i)}_{kj}(z) \right| \]
on the event where $\|E \| \leq 2 (\sqrt N +\sqrt{n})$. Since the $i$-th row of $\mathcal E$ is independent of $G^{(i)}$, conditioning on $G^{(i)}$, $\sum_k^{(i)} \mathcal E_{ik} G^{(i)}_{kj}(z)$ still has a Gaussian distribution.  In particular, the real part $\sum_k^{(i)} \mathcal E_{ik} \Re G^{(i)}_{kj}(z)$ has a Gaussian distribution with mean $0$ and variance $\sum_{k}^{(i)}\left(\Re G^{(i)}_{kj}\right)^2 \le \sum_{k}^{(i)} \overline{G^{(i)}_{kj}}G^{(i)}_{kj}   = ((G^{(i)})^\ast G^{(i)} )_{jj}$. Likewise, the imaginary part $\sum_k^{(i)} \mathcal E_{ik} \Im G^{(i)}_{kj}(z)$ also has a Gaussian distribution with mean $0$ and variance at most $((G^{(i)})^\ast G^{(i)} )_{jj}$. Using the tail bounds for the Gaussian distribution \cite[Proposition 2.1.2]{Vbook}, we get
\begin{align*}
\left| \sum_k^{(i)} \mathcal E_{ik} G^{(i)}_{kj}(z) \right| &\leq \left| \sum_k^{(i)} \mathcal E_{ik} \Re G^{(i)}_{kj}(z) \right| + \left| \sum_k^{(i)} \mathcal E_{ik} \Im G^{(i)}_{kj}(z) \right|\\
&\le 2\sqrt{2(K+1)\log (N+n)} \sqrt{ ((G^{(i)})^\ast G^{(i)} )_{jj} } 
\end{align*}
with probability at least $1 - 0.5 (N+n)^{-(K+1)}$. 
By bounding $((G^{(i)})^\ast G^{(i)} )_{jj}  \leq \|G^{(i)} \|^2 \le (2/|z|)^2$, we conclude that 
\[  |G_{ij}(z)| \leq \frac{ 8 \sqrt{2(K+1)\log (N+n)}}{ |z|^2} \]
with probability at least $1 - 0.5 (N+n)^{-(K+1)}$.  Union bounding over $i,j \in \llbracket 1, r\rrbracket \cup \llbracket N+1, N+r\rrbracket$ completes the proof.  
\end{proof}

We conclude this section with the proof of Lemma \ref{lemma:assumpGOE}.
\begin{proof}[Proof of Lemma \ref{lemma:assumpGOE}]
Fix an index $j\in \llbracket i_0, r_0\rrbracket$. By Lemma \ref{lemma:assumpfixed},  for any $z\in \mathbb C$ with $|z| \ge 4(\sqrt N + \sqrt n)$,
\[ \left\| \mathcal U^T \left(G(z)-\Phi(z) \right) \mathcal U  \right\| \leq 48r\frac{\sqrt{(K+8)\log (N+n)}}{|z|^2} \] 
with probability at least $1 - (N+n)^{-(K+7)}$. Note that every $z \in S_{\sigma_j}$ satisfies $|z| \ge 4 (\sqrt N+\sqrt{n})$ by assumption.  Let $\mathcal{N}$ be a $1$-net of $S_{\sigma_j}$.  Since $\sigma_j \le n^2$, a simple volume argument (see for instance \cite[Lemma 3.3]{OW}) shows $\mathcal{N}$ can be chosen so that $|\mathcal{N}| \leq  20[(20\eta r)^2 + (8n^2/7  + 20\eta r)^2]$. 
By the union bound, the supposition $(\sqrt N + \sqrt n)^2 \ge 64 (K+9)\log(N+n)$ and Lemma \ref{lemma:assumpfixed},
\begin{equation} \label{eq:net}
	\max_{ z \in \mathcal{N}} |z|^2 \left\| \mathcal U^\T \left(G(z)-\Phi(z) \right) \mathcal U \right\| \leq 48r \sqrt{(K+8)\log (N+n)} 
\end{equation} 
with probability at least $1 - 10(N+n)^{-(K+1)}$.  We now wish to extend this bound to all $z \in S_{\sigma_j}$.  

Define the functions
\[ f(z) := z^2 \mathcal U^\T G(z) \mathcal U, \qquad g(z) := z^2\mathcal U^\T \Phi(z) \mathcal U. \]
In order to complete the proof, it suffices to show that $f$ and $g$ are $8$-Lipschitz in $S_{\sigma_j}$.  In other words, we want to show that $\| f(z) - f(w)\| \leq 8 |z-w|$ and $\| g(z) - g(w)\| \leq 8 |z-w|$ for all $z,w \in S_{\sigma_j}$.  Indeed, in view of \eqref{eq:net}, if $z \in S_{\sigma_j}$, then there exists $w \in \mathcal{N}$ so that $|z - w| \leq 1$, and hence \begin{align*}
	|z|^2 \left\| \mathcal U^\T G(z) \mathcal U - \mathcal U^T \Phi(z) \mathcal U \right\| &
	\le \|f(z) - f(w)\| +\|f(w) - g(w)\| + \|g(w) - g(z)\| \\
	&\leq 16 + |w|^2 \left\| \mathcal U^\T G(w) \mathcal U - \mathcal U^T \Phi(w) \mathcal U \right\| \\
	&\leq 16 + 48r \sqrt{(K+8)\log (N+n)}  \\
	&\leq  54 r \sqrt{(K+8)\log (N+n)},  
\end{align*}
 where we used the Lipschitz continuity of $f$ and $g$ in the second inequality. 
 
It remains to show that $f$ and $g$ are $8$-Lipschitz in $S_{\sigma_j}$.  To do so we will only work on the event where $\|E \| \leq 2(\sqrt N+ \sqrt{n})$; the probability of this event is at least $1-2e^{(\sqrt N + \sqrt n)^2/2} \ge 1- 2(N+n)^{-32(K+9)}$ by Lemma \ref{lemma:norm} and the supposition $(\sqrt N + \sqrt n)^2 \ge 64 \log(N+n)(K+9)$.  Let $z,w \in S_{\sigma_j}$, and assume without loss of generality that $|z| \geq |w| \ge 4(\sqrt N+ \sqrt{n})$.  Then
\begin{align*}
	\| f(z) - f(w) \| &\leq \|z^2 \mathcal U^\T G(z) \mathcal U - zw \mathcal U^\T G (z) \mathcal U \| + \| zw \mathcal U^\T G(z) \mathcal U - w^2 \mathcal U^\T G(z) \mathcal U \| \\
	&\qquad + \| w^2 \mathcal U^\T G(z) \mathcal U - w^2 \mathcal U^\T G(w) \mathcal U \| \\
	&\leq |z| \|G(z)\| |z - w| + |w| |z-w| \|G(z) \| + |w|^2 |z-w| \|G(z) \| \| G(w) \| \\
	&\leq 8 |z-w|,
\end{align*}
where we used the resolvent identity \eqref{eq:resolvent}, Lemma \ref{lemma:resolventnorm}, and the fact that $\frac{|w|}{|z|} \leq 1$.  This shows that $f$ is $8$-Lipschitz in $S_{\sigma_j}$.  

The proof for $g$ is similar. First, for $|z| \geq |w| \ge 4(\sqrt N+ \sqrt{n})$, by the triangle inequality, we have 
\begin{align}\label{eq:bdforg}
	\| g(z)& - g(w) \| \nonumber\\
	&\leq \|z^2 \mathcal U^\T \Phi(z) \mathcal U - zw \mathcal U^\T \Phi (z) U \| + \| zw \mathcal U^\T \Phi(z) \mathcal U - w^2 \mathcal U^\T \Phi(z) \mathcal U \| \nonumber\\
	&\qquad + \| w^2 \mathcal U^\T \Phi(z) \mathcal U - w^2 \mathcal U^\T \Phi(w) \mathcal U \| \nonumber \\
	&\leq |z||z - w| \|\mathcal U^\T \Phi(z) \mathcal U\|  + |w| |z-w| \|\mathcal U^\T \Phi(z) \mathcal U\| + |w|^2 \|\mathcal U^\T (\Phi(z)-\Phi(w)) \mathcal U \|.
\end{align}
Using the explicit expression in \eqref{eq:productPhi}, we find that
\begin{align*}
\|\mathcal U^\T (\Phi(z)-\Phi(w)) \mathcal U \| = \max\left\{ \frac{|\phi_1(z)-\phi_1(w)|}{|\phi_1(z)\phi_1(w)|}, \frac{|\phi_2(z)-\phi_2(w)|}{|\phi_2(z)\phi_2(w)|} \right\}.
\end{align*}
By \eqref{eq:defphi} and the resolvent identity \eqref{eq:resolvent}, 
\begin{align*}
|\phi_1(z)-\phi_1(w)| &= |z-w - \tr \mathcal I^\mathrm{d} (G(z)-G(w))|\\
&=|z-w - (z-w)\tr \mathcal I^\mathrm{d} G(z)G(w)| \\
&\le |z-w| \left( 1+ (N+n) \|G(z)\| \|G(w)\| \right)\\
&\le |z-w| \left( 1+ \frac{4(N+n) }{|z| |w|} \right) \\
&\le \frac{5}{4}|z-w|,
\end{align*}
where we used Lemma \ref{lemma:resolventnorm} and the facts that $\frac{N+n}{|w|}\le \frac{|w|}{16}$ and $\frac{|w|}{|z|} \leq 1$. The same upper bound also holds for $|\phi_2(z)-\phi_2(w)|$. Combining these estimates with \eqref{eq:phibd}, we have $$\|\mathcal U^\T (\Phi(z)-\Phi(w)) \mathcal U \| \le \frac{80}{49} \frac{|z-w|}{|z| |w|}.$$
Notice that 
$\|\mathcal U^\T \Phi(z) \mathcal U\| \le \frac{8}{7|z|}$ for any $|z| \ge 4(\sqrt N + \sqrt n)$, which can be verified using \eqref{eq:Phinorm} and the bounds in \eqref{eq:phibd}. Inserting these bounds into \eqref{eq:bdforg} yields that $g$ is 8-Lipschitz in $S_{\sigma_j}$; we omit the details. 

The proof is complete by taking a union bound over $j \in \llbracket i_0, r_0\rrbracket$.
\end{proof}

\appendix
\section{Proof of Theorem \ref{thm:lowerbd}}\label{app} 
This section is devoted to the proof of Theorem \ref{thm:lowerbd}.  
Unless otherwise specified, we let $C$ and $c$ be positive constants, which may change from one occurrence to the next, depending only on the sub-gaussian moment of the entries of $E$.  Without loss of generality, we assume $N\le n$, for if not, one can simply replace $A$ and $E$ by their transposes.

Let $P_1\in \mathbb R^{N\times N}$ be the orthogonal projection matrix onto the subspace $\Span\{u_1,\ldots,u_r \}^\perp$ and $P_2\in \mathbb R^{n\times n}$ be that onto $\Span\{v_1,\ldots,v_r \}^\perp$. 

We first prove that
\begin{align}
\widetilde \sigma_1 \sin\angle (u_1, \widetilde{u}_1) + \sqrt{2}\|E \| \sin\angle (v_1, \widetilde{v}_1) &\ge \|P_1 E v_1\|,\nonumber\\
\widetilde \sigma_1 \sin\angle (v_1, \widetilde{v}_1) + \sqrt{2}\|E \| \sin\angle (u_1, \widetilde{u}_1) &\ge \|P_2 E^T u_1\|.\label{eq:ineq}
\end{align}
To start, observe that 
\begin{align}\label{eq:decom0}
\sin^2\angle( u_1, \widetilde{ u}_1) &= 1- \langle  u_1, \widetilde{ u}_1 \rangle^2 = \|\widetilde{ u}_1\|^2 - \langle  u_1, \widetilde{ u}_1 \rangle^2 \nonumber\\
&=\sum_{i=2}^{r} \langle  u_i, \widetilde{ u}_1 \rangle^2 + \| P_1 \widetilde{ u}_1\|^2 \ge \| P_1 \widetilde{ u}_1\|^2.
\end{align}
By multiplying by $P_1$ on all sides of the equation $\widetilde{ A} \widetilde{ v}_1 = { A} \widetilde{ v}_1 + { E} \widetilde{ v}_1=\widetilde \sigma_1  \widetilde{ u}_1$, we get $ P_1  E \widetilde{ v}_1 = \widetilde \sigma_1  P_1 \widetilde{ u}_1$. Continuing from \eqref{eq:decom0}, we have
\begin{align}\label{eq:sinbd1}
\widetilde \sigma_1 \sin\angle( u_1, \widetilde{ u}_1) \ge \| P_1 E  \widetilde{v}_1\|.
\end{align}
Likewise, we also have 
\begin{align}\label{eq:sinbd2}
\widetilde \sigma_1 \sin\angle( v_1, \widetilde{ v}_1) \ge  \| P_2 E^T  \widetilde{u}_1\|.
\end{align}
Let $\alpha$ denote the angle between $u_1$ and $\widetilde u_1$ (taken in $[0, \pi]$) and $\beta$ denote the angle between $v_1$ and $\widetilde v_1$ (taken in $[0, \pi]$).  By possibly multiplying $\widetilde{u}_1, \widetilde{v}_1$ by $-1$, it suffices to consider one of the following cases: either (i) $\alpha,\beta \in [0,\frac{\pi}{2}]$; or (ii) $\alpha \in [\frac{\pi}{2},\pi]$ and $\beta \in [0,\frac{\pi}{2}]$.

If $\alpha,\beta \in [0,\frac{\pi}{2}]$, by simple trigonometric identities, 
\begin{align}\label{eq:geo}
\|{ u}_1- \widetilde{ u}_1\| = 2\sin (\alpha/2) = \frac{\sin\alpha}{\cos (\alpha/2)} \le \sqrt{2}\sin\alpha= \sqrt{2} \sin\angle ( u_1, \widetilde{ u}_1).
\end{align}
Similarly, we have 
$$\|{ v}_1- \widetilde{ v}_1\| \le \sqrt{2} \sin\angle ( v_1, \widetilde{ v}_1).$$
From \eqref{eq:sinbd1} and the triangle inequality, we obtain 
\begin{align*}
\widetilde \sigma_1 \sin\angle( u_1, \widetilde{ u}_1) &\ge \| P_1 E  \widetilde{v}_1\| \\
&\ge \| P_1 E v_1\|-\|P_1 E (\widetilde{v}_1-v_1)\|\\
&\ge \| P_1 E v_1\| - \|E\| \|\widetilde{v}_1-v_1\| \\
&\ge \| P_1 E v_1\| - \sqrt{2}\|E\|\sin\angle ( v_1, \widetilde{ v}_1).
\end{align*}
Repeating the same argument, we obtain
\begin{align*}
\widetilde \sigma_1 \sin\angle( v_1, \widetilde{ v}_1) &\ge \| P_2 E^T u_1\| - \sqrt{2}\|E\|\sin\angle ( u_1, \widetilde{ u}_1).
\end{align*}

In the second case, if $\alpha \in [\frac{\pi}{2},\pi]$ and $\beta \in [0,\frac{\pi}{2}]$, we still have $\|{ v}_1- \widetilde{ v}_1\| \le \sqrt{2} \sin\angle ( v_1, \widetilde{ v}_1)$, and hence
\begin{align*}
\widetilde \sigma_1 \sin\angle( u_1, \widetilde{ u}_1) &\ge \| P_1 E v_1\| - \sqrt{2}\|E\|\sin\angle ( v_1, \widetilde{ v}_1).
\end{align*}
Since $\theta:=\pi-\alpha \in [0,\frac{\pi}{2}]$, using the same estimate as in \eqref{eq:geo}, we see that
$$\|u_1 + \widetilde{ u}_1\| = 2\sin (\theta/2) = \frac{\sin\theta}{\cos (\theta/2)} \le \sqrt{2}\sin\theta=\sqrt{2}\sin(\pi-\alpha)= \sqrt{2} \sin\angle ( u_1, \widetilde{ u}_1).$$
Thus, from \eqref{eq:sinbd2}, we conclude that 
\begin{align*}
\widetilde \sigma_1 \sin\angle( v_1, \widetilde{ v}_1) &\ge  \| P_2 E^T  \widetilde{u}_1\| \\
&\ge \| P_2 E^T u_1\|-\|P_2 E^T (\widetilde{u}_1+u_1)\|\\
&\ge \| P_2 E^T u_1\| - \|E\| \|\widetilde{u}_1+u_1\| \\
&\ge \| P_2 E^T u_1\| - \sqrt{2}\|E\|\sin\angle ( u_1, \widetilde{ u}_1).
\end{align*}
Rearranging the terms yields \eqref{eq:ineq}. 

We now turn to the proof of Theorem \ref{thm:lowerbd}. It follows immediately from \eqref{eq:ineq} that 
\begin{align}\label{eq:2bd}
\max\{\sin\angle (u_1, \widetilde{u}_1), \sin\angle (v_1, \widetilde{v}_1)\}&\ge \frac{\max\{\|P_1 E v_1\|, \|P_2 E^T u_1\| \}}{\widetilde \sigma_1 + \sqrt{2} \|E\|} \ge \frac{\max\{\|P_1 E v_1\|, \|P_2 E^T u_1\| \}}{ \sigma_1 + (1 + \sqrt{2}) \|E\|},
\end{align}
where in the last inequality we used $\widetilde \sigma_1\le \sigma_1 + \|E\|$ from the classical Weyl's inequality.
Furthermore, using \cite[Lemma 11.8]{OVW}, one has
\begin{align*}
&\Prob(|\|P_2 E^T u_1\|^2 - (n-r)| > t) \le C \exp\left(-\min \left\{ \frac{t^2}{n-r}, t \right\} \right)
\end{align*}
for any $t \geq 0$.  Thus, since $N \leq n$, $$\max\{\|P_1 E v_1\|, \|P_2 E^T u_1\| \} \ge \|P_2 E^T u_1\| \ge \frac{1}{2}\sqrt{n-r}$$ with probability at least $1-C \exp(-c(n-r)).$ Plugging into \eqref{eq:2bd}, we get
\begin{align*}
\max\{\sin\angle (u_1, \widetilde{u}_1), \sin\angle (v_1, \widetilde{v}_1)\}& \ge \frac{1}{2}\frac{\sqrt{n-r}}{ \sigma_1 + (1+\sqrt{2})\|E\|}=\frac{\|E\|}{2\sigma_1}\frac{ \frac{\sqrt{n - r}}{\|E\|} }{1+(1 + \sqrt{2})\frac{\|E\|}{\sigma_1}}.
\end{align*}
Applying \cite[Proposition 2.4]{MR2827856}, we have $\|E\| \le 2(\sqrt N + \sqrt n)$ with probability at least $1-C\exp(-c(\sqrt N + \sqrt n)^2)$. It follows that $\frac{\sqrt{n - r}}{\|E\|} \ge \frac{1}{4\sqrt{2}}$, where we used that $r\le n/2$. Finally, we conclude that
\begin{align*}
\max\{\sin\angle (u_1, \widetilde{u}_1), \sin\angle (v_1, \widetilde{v}_1)\}& \ge \frac{1}{8\sqrt{2}}\frac{ \frac{\|E\|}{\sigma_1} }{1+(1 + \sqrt{2})\frac{\|E\|}{\sigma_1}}
\end{align*}
with probability at least $1-C \exp(-c(n-r))$. The claim now follows from the fact that $r \le n/2$.  

\section{Proofs of Propositions \ref{prop:spacebd}, \ref{prop:blockphi}, \ref{prop:evblock}, \eqref{eq:difference} and Lemma \ref{lem:svloc-large}}\label{proofdiff}\label{app:B}

In this section, we collect the proofs of Propositions \ref{prop:spacebd}, \ref{prop:blockphi}, \ref{prop:evblock}, \eqref{eq:difference} and Lemma \ref{lem:svloc-large} from Section \ref{sec:prelim}. We continue to use the notation introduced in the previous sections.  

\subsection{Proof of Proposition \ref{prop:spacebd}}\label{appB:prop7}
Let $0\le \alpha_1\le \cdots \le \alpha_p \le \pi/2$ be the principal angles between $U$ and $\widetilde U$. Then $$\cos\alpha_p = \cos\angle (U, \widetilde U).$$ Let $\{u_1,\ldots, u_p\}$ (resp. $\{\widetilde u_1,\ldots, \widetilde u_p\}$) be an orthonormal basis for $U$ (resp. for $\widetilde U$). Denote the matrices $\mathbf U:=(u_1,\ldots,u_p)$ and $\mathbf V:=(v_1,\ldots,v_p)$.  \cite[Theorem 1] {MR348991} provides the connection between the principle angles and the singular values of $\mathbf U^\T \widetilde{\mathbf U}$. That is, consider the SVD of $\mathbf U^\T \widetilde{\mathbf U}$ given by $$\mathbf U^\T \widetilde{\mathbf U}= Y_1 C_1 Z_1^T,$$ where $C_1$ is a diagonal matrix composed of the singular values and $Y_1, Z_1$ are $p \times p$ orthogonal matrices. Then the diagonal entries of $C_1$ are exactly $\cos\alpha_1,\ldots,\cos\alpha_p.$

Analogously, let $0\le \beta_1\le \cdots \le \beta_p \le \pi/2$  be the principal angles between $V$ and $\widetilde V$. Then $$\cos\beta_p = \cos\angle (V, \widetilde V).$$ For the orthonormal basis $\{v_1,\ldots, v_p\}$ for $V$ and $\{\widetilde v_1,\ldots, \widetilde v_p\}$ for $\widetilde V$, denote $\mathbf V:=(v_1,\ldots,v_p)$ and $\widetilde{\mathbf V}:=({\widetilde v}_1,\ldots,{\widetilde v}_p)$. Consider the SVD  $$\mathbf V^\T \widetilde{\mathbf V} = Y_2 C_2 Z_2^T,$$ where $C_2$ is a diagonal matrix composed of the singular values and $Y_2, Z_2$ are orthogonal matrices. Then the diagonal entries of $C_2$ are $\cos\beta_1,\cdots,\cos\beta_p.$

Next, define $\mathbf w_i:=(u_i^\T, v_i^\T)^\T$ and $\mathbf w_{i+p}:=(u_{i}^\T, -v_{i}^\T)^\T$ for $1\le i \le p$. Define $\widetilde{\mathbf w}_i$'s ($1\le i \le 2p$) analogously. It is easy to verify that $\frac{1}{\sqrt{2}}\{\mathbf w_1, \cdots, \mathbf w_{2p}\}$ forms an orthonormal basis of $W$ and $\frac{1}{\sqrt{2}}\{\widetilde{\mathbf w}_1, \cdots, \widetilde{\mathbf w}_{2p}\}$ forms an orthonormal basis of $\widetilde W$. Denote the matrices $\mathbf W:=\frac{1}{\sqrt{2}}(\mathbf w_1, \cdots, \mathbf w_{2p})$ and $\widetilde{\mathbf W}:=\frac{1}{\sqrt{2}}(\widetilde{\mathbf w}_1, \cdots, \widetilde{\mathbf w}_{2p})$. Let $0\le \gamma_1\le \cdots \le \gamma_{2p} \le \pi/2$ be the principal angles between $W$ and $\widetilde W$. Then $\cos \gamma_1, \ldots, \cos \gamma_{2p}$ are the singular values of $\mathbf W^\T \widetilde{\mathbf W}$ by \cite[Theorem 1]{MR348991}. In particular, the smallest singular value of $\mathbf W^\T \widetilde{\mathbf W}$ is $\cos \gamma_{2p} = \cos \angle(W, \widetilde W)$. 

Observe that
\begin{align*}
\mathbf W^\T \widetilde{\mathbf W} = \frac{1}{2}\begin{pmatrix}
U&U\\
V&-V
\end{pmatrix}^\T 
\begin{pmatrix}
\widetilde U& \widetilde U\\
\widetilde V&-\widetilde V
\end{pmatrix}=
\frac{1}{2}\begin{pmatrix}
U^\T \widetilde U + V^\T \widetilde V & U^\T \widetilde U - V^\T \widetilde V\\
U^\T \widetilde U - V^\T \widetilde V & U^\T \widetilde U + V^\T \widetilde V
\end{pmatrix}.
\end{align*}
Using the SVDs of $U^\T \widetilde U$ and $V^\T \widetilde V$ and simple computations, we further get
\begin{align*}
\mathbf W^\T \widetilde{\mathbf W} = \frac{1}{\sqrt 2}\begin{pmatrix}
Y_1 & -Y_2\\
Y_1 & Y_2
\end{pmatrix}
\begin{pmatrix}
 C_1 & 0 \\
0  & C_2
\end{pmatrix}
\frac{1}{\sqrt 2}\begin{pmatrix}
Z_1^\T & Z_1^\T \\
-Z_2^\T & Z_2^\T
\end{pmatrix} =:Y C Z^\T.
\end{align*}
Since $Y_1, Y_2, Z_1, Z_2$ are orthogonal matrices, it is easy to verify that both $Y$ and $Z$ are $2p \times 2p$ orthogonal matrices. Hence, $YC Z^\T$ is the SVD of  $\mathbf W^\T \widetilde{\mathbf W}$. By \cite[Theorem 1]{MR348991},  the singular values $\cos \gamma_1, \cdots, \cos \gamma_{2p}$ of $\mathbf W^\T \widetilde{\mathbf W}$ are exactly $$\cos\alpha_1,\cdots,\cos\alpha_p, \cos\beta_1,\cdots,\cos\beta_p.$$ Hence, $$ \cos \angle(W, \widetilde W)=\cos \gamma_{2p} =\min\{\cos\alpha_p,  \cos\beta_p\}=\min\{\cos\angle(U, {\widetilde U}),  \cos\angle(V, {\widetilde V})\},$$which is equivalent to
 $$\sin \angle(W, \widetilde W) = \max \{ \sin\angle(U, {\widetilde U}) ,\sin \angle(V, {\widetilde V})  \}.$$ This completes the proof.

\subsection{Proof of Propositions \ref{prop:blockphi}}
Note that $\Phi(x)$ is well-defined when $|x|>\|\mathcal E\|$. From the expression of $\mathcal U^\T \Phi(x) {\mathcal U}$ in \eqref{eq:productPhi}, using the block structure of matrices, we rewrite
\begin{equation*}
I_{2r-2r_0} -\mathcal U_J^\T \Phi(x) {\mathcal U}_J \mathcal D_J= 
\left(\begin{array}{cc}
I_{r-r_0} - \alpha \mathcal D_{\llbracket r_0+1, r\rrbracket} & -\beta \mathcal D_{\llbracket r+r_0+1, 2r\rrbracket} \\
- \beta \mathcal D_{\llbracket r_0+1, r\rrbracket} & I_{r-r_0}-\alpha \mathcal D_{\llbracket r+r_0+1, 2r\rrbracket}
\end{array}\right).
\end{equation*}
Note that $\mathcal D_{\llbracket r+r_0+1, 2r\rrbracket}=-\mathcal D_{\llbracket r_0+1, r\rrbracket}$ by the definition of $\mathcal D$. We compute the eigenvalues of $(I_{2r-2r_0} - \mathcal U_J^\T \Phi(\widetilde \sigma_j) {\mathcal U}_J {\mathcal D}_J)(I_{2r-2r_0} - \mathcal U_J^\T \Phi(\widetilde \sigma_j) {\mathcal U}_J {\mathcal D}_J)^T$, which, after simplification, has the following format
\begin{equation}\label{eq:blkmatrix}
\begin{pmatrix}
(I -\alpha \mathcal D_{\llbracket r_0+1, r\rrbracket})^2 + \beta^2 \mathcal D_{\llbracket r_0+1, r\rrbracket}^2 &  2 \alpha \beta \mathcal D_{\llbracket r_0+1, r\rrbracket}^2 \\
2 \alpha \beta \mathcal D_{\llbracket r_0+1, r\rrbracket}^2 &(I +\alpha \mathcal D_{\llbracket r_0+1, r\rrbracket})^2 + \beta^2 \mathcal D_{\llbracket r_0+1, r\rrbracket}^2
\end{pmatrix}.
\end{equation}
Note that each block of \eqref{eq:blkmatrix} is a diagonal matrix. Using basic linear algebra and a simple computation, we further obtain the eigenvalues of the above matrix are given by 
$$1+\sigma_t^2(\alpha^2 + \beta^2) \pm 2|\alpha| \sigma_t \sqrt{1+\beta^2 \sigma_t^2}=\left(\sqrt{1+\beta^2 \sigma_t^2} \pm |\alpha| \sigma_t \right)^2$$ 
for $r_0+1\le t \le r$. Taking the square roots yields the singular values of $I_{2r-2r_0} -\mathcal U_J^\T \Phi(x) {\mathcal U}_J \mathcal D_J$ and completes the proof.

\subsection{Proof of Propositions \ref{prop:evblock}}
Note that $\Phi(z)$ is well-defined when $|z|>\|\mathcal E\|$.  We first compute the eigenvalues of 
\begin{align}\label{eq:positivematrix}
\left( \mathcal D^{-1} - \mathcal U^\T \Phi(z) \mathcal U \right)\left( \mathcal D^{-1} - \mathcal U^\T \Phi(z) \mathcal U \right)^*,
\end{align}
where we remind the readers that $\mathcal D$ and $\mathcal U$ are defined in Section \ref{sec:linearalgebra}. In particular, $\mathcal D=\diag(\diag(D),-\diag(D))$. Recall the definition of  $\mathcal U^\T \Phi(z) \mathcal U$ in \eqref{eq:productPhi}. Note that 
$$\mathcal D^{-1} - \mathcal U^\T \Phi(z) \mathcal U =
\begin{pmatrix}
 D^{-1}-\alpha I_r & -\beta I_r\\
-\beta I_r & -D^{-1}-\alpha I_r
\end{pmatrix}
$$
where each block is a diagonal matrix. Elementary (yet tedious) calculations yield that the eigenvalues of \eqref{eq:positivematrix} are
\begin{align*}
|\alpha|^2 +|\beta|^2 +\frac{1}{\sigma_l^2} \pm \left( (2|\alpha|^2 + 2|\beta|^2 +\bar\alpha^2 - \bar\beta^2+\alpha^2-\beta^2)\frac{1}{\sigma_l^2} + 2|\alpha|^2|\beta|^2  +\bar\alpha^2\beta^2 +\alpha^2\bar\beta^2 \right)^{1/2}.
\end{align*}
With further simplification after plugging in the expressions of $\alpha,\beta$, we  denote these eigenvalues by 
\begin{align*}
&\gamma_{l_\pm}:=\frac{1}{\sigma_l^2} + \frac{1}{2} \left( \frac{1}{|\phi_1|^2} + \frac{1}{|\phi_2|^2} \right) \pm \frac{1}{2}\left[ \frac{4}{\sigma_l^2} \left| \frac{1}{\phi_1} +\frac{1}{\bar{\phi}_2} \right|^2  + \left( \frac{1}{|\phi_1|^2} - \frac{1}{|\phi_2|^2} \right)^2 \right]^{1/2}
\end{align*}
for $1\le l \le r$. Since the entries of $E$ have continuous distribution, $ \mathcal D^{-1} - \mathcal U^\T \Phi(z) \mathcal U $ is invertible with probability 1. Consequently, 
\begin{align*}
\left \| \left( \mathcal D^{-1} - \mathcal U^\T \Phi(z) \mathcal U \right)^{-1} \right \|^2=\max_{1\le l \le r} \frac{1}{\gamma_{l-}}= \max_{1\le l \le r} \frac{\sigma_l^2}{|\sigma_l^2 - \phi_1 \phi_2|^2} \mathcal Q,
\end{align*}
where $$\mathcal Q:=|\phi_1 \phi_2|^2 + \frac{1}{2}\sigma_l^2 (|\phi_1|^2 + |\phi_2|^2) + \frac{1}{2}\sigma_l \left[ 4 |\phi_1 \phi_2|^2 |\phi_1 + \bar{\phi}_2|^2 + \sigma_l^2(|\phi_1|^2 - |\phi_2|^2)^2\right]^{1/2}.$$
The conclusion of Propositions \ref{prop:evblock} follows by taking a square root.

\subsection{Proof of \eqref{eq:difference}}\label{app:diff}
The singular value decomposition of $E$ is given by $E=X \diag(\eta_1,\ldots,\eta_N) Y^\T$, where the columns of $X$ (resp. $Y$) are $\mathbf x_1,\ldots, \mathbf x_N \in \mathbb{R}^{N}$ (resp. $\mathbf y_1,\ldots, \mathbf y_N \in \mathbb{R}^{n}$).  Clearly, some of the $\eta_i$'s may be zero.  However, if $n > N$, then $E$ trivially maps an $(n-N)$-dimensional space to zero; we assume this subspace has orthonormal basis $\mathbf h_{1}, \ldots, \mathbf h_{n-N} \in \mathbb{R}^{n}$. 

The spectral decomposition of $\mathcal E$ is then given by 
\[ \mathcal E = \mathcal W \diag(\eta_1,\ldots,\eta_N, -\eta_1,\ldots,-\eta_N)\mathcal W^\T, \]
where the columns of $\mathcal W$ are the orthonormal vectors $\mathbf w_i = \frac{1}{\sqrt 2} (\mathbf x_i^\T, \mathbf y_i^\T)^\T \in \mathbb{R}^{N+n}$ and $\mathbf w_{N+i} = \frac{1}{\sqrt 2} (\mathbf x_i^\T, -\mathbf y_i^\T)^\T \in \mathbb{R}^{N+n}$ for $1\le i \le N$. Likewise, $\mathcal E$ trivially maps an $(n-N)$-dimensional space to zero; this subspace is spanned by the orthonormal vectors $\mathbf w_{2N+j} = (0, \mathbf h_j^\T)^\T \in \mathbb{R}^{N+n}$ for $j=1,\ldots,n-N$. Thus, the spectral decomposition of the resolvent $G(z)$ can be expressed as
\[ G(z)=(z-\mathcal E)^{-1} = \sum_{i=1}^N \frac{\mathbf w_i \mathbf w_i^\T}{z- \eta_i} + \sum_{i=1}^N \frac{\mathbf w_{N+i} \mathbf w_{N+i}^\T}{z+ \eta_i} + \frac{1}{z}\sum_{j=1}^{n-N} \mathbf w_{2N+j} \mathbf w_{2N+j}^\T. \]
It follows that 
\begin{align}\label{eq:traceid}
&\tr \mathcal I^\mathrm{d} G(z) = \frac{1}{2} \sum_{i=1}^N \left( \frac{1}{z-\eta_i} + \frac{1}{z+\eta_i} \right) + \frac{1}{z}(n-N), \\
& \tr \mathcal I^\mathrm{u} G(z)= \frac{1}{2} \sum_{i=1}^N \left( \frac{1}{z-\eta_i} + \frac{1}{z+\eta_i} \right), \label{eq:traceid2}
\end{align}
and thus, from \eqref{eq:defphi}, we see that \eqref{eq:difference} holds.

\subsection{Proof of Lemma \ref{lem:svloc-large}}\label{appb:lem} We work on the event that $ \|E\| \le 2(\sqrt N + \sqrt n) $, which holds with probability at least $1 - 2 e^{-(\sqrt{N}+\sqrt{n})^2/2}$ by Lemma \ref{lemma:norm}. 

If $r^2 > \frac{1}{27}(\sqrt N + \sqrt n)$, then the conclusion follows directly from the Weyl's inequality and the supposition on $n,N$:
$$\max_{1\le l \le r}|\widetilde \sigma_l - \sigma_l | \le \|E\| \le 2(\sqrt N + \sqrt n) \le \eta r = 54r^2 \sqrt{(K+8)\log (N+n)}.$$

If  $r^2 \le \frac{1}{27}(\sqrt N + \sqrt n)$, then $\eta r < \frac{1}{4}n^2 <\frac{1}{2}\sigma_l$ for any $\sigma_l >\frac{1}{2}n^2$. We apply \cite[Theorem 23]{OVW2} by noting that $E$ is $(2, \frac{1}{2}, 2)$-concentrated in the Definition 11 from \cite{OVW2} to get
$$\widetilde{\sigma}_l \ge \sigma_l -\eta r $$ uniformly for any $1\le l \le r$ with probability at least $1-9^{r+1} \exp(-(\eta r)^2/32)$. Conditioning on this event, with the assumption that $\sigma_l >\frac{1}{2}n^2$ which implies $\widetilde{\sigma}_l > \frac{1}{2}\sigma_l$, \cite[Theorem 23]{OVW2} yields the lower bound
\begin{align*}
\widetilde{\sigma}_l \le \sigma_l + t \sqrt{r} + 128\frac{\sqrt r}{n} + 2^9 \frac{r}{n^{5/2}}
\end{align*}
with probability at least $1- 4\cdot 9^{2r} \exp(-r t^2/32)$ for any $t > 0$. Note that $128\frac{\sqrt r}{n} + 2^9 \frac{r}{n^{5/2}} \le (3/100) \eta r$ by the supposition on $n$. The conclusion follows by taking $t\sqrt r = \frac{1}{2}\eta r$, a union bound over $l \le r_0$ and simplifying the probability bounds using $(\eta r)^2 > 54^2 r^4$. 

\section{Proofs of Lemma \ref{lemma:resolventnorm}, \ref{lem:simpleapprox}, and \ref{lem:Unoise}}\label{app:C}

In this section, we collect the proofs of Lemma \ref{lemma:resolventnorm}, \ref{lem:simpleapprox}, and \ref{lem:Unoise}.

\subsection{Proof of Lemma \ref{lemma:resolventnorm}}\label{appc:resolventnorm}
By writing $G(z) := (z - \mathcal E)^{-1}$ as a Neumann series, we find
\[ \|G(z) \| \leq \frac{1}{|z|} \sum_{k=0}^\infty \left( \frac{ \|\mathcal E\|}{|z|} \right)^k. \]
Since $\|E\|= \|\mathcal E\|$ and $\frac{\|\mathcal E\|}{|z|} \leq 1/2$ by assumption, the claim for $\| G(z) \|$ follows.  Since $\|\mathcal E^{(k)}\| \leq \|\mathcal E\|$, the same proof also applies to $G^{(k)}(z)$.   The bounds in \eqref{eq:phibd} follow from \eqref{eq:defphi}, the bound
\begin{equation}\label{eq:bdid}
\max\{|\tr \mathcal I^u G(z)  |, |\tr \mathcal I^d G(z)  |\} \le (N+n) \|G(z)\| \le \frac{2(N+n)}{|z|} \le \frac{|z|}{8}, 
\end{equation}
and the triangle inequality.  

\subsection{Proof of Lemma \ref{lem:simpleapprox}}\label{appc:simpleapprox}
By writing $G(z) := (z - \mathcal E)^{-1}$ as a Neumann series, we see that
\begin{align*}
\left\| G(z) - \frac{1}{z} I_{N+n} - \frac{\mathcal E}{z^2}\right\| \le  \frac{1}{|z|} \sum_{k=2}^\infty \left( \frac{ \|\mathcal E\|}{|z|} \right)^k =\frac{\|\mathcal E \|^2}{|z|^2(|z|-\|\mathcal E \|)}  \le \frac{2\|\mathcal E\|^2}{|z|^3}
\end{align*}
since $\frac{\|\mathcal E\|}{|z|} \leq 1/2$ and $|z|-\|\mathcal E\|\ge \frac{1}{2}|z|$ by assumption. 

\subsection{Proof of Lemma \ref{lem:Unoise}}
By the rotational invariance of $E$, it suffices to assume that $U$ is the matrix with columns $e_1, \ldots, e_r$, where $e_1, \ldots, e_N$ is the canonical basis in $\mathbb{R}^N$, and the columns of $V$ are given by $f_1, \ldots, f_r$, where $f_1, \ldots, f_n$ is the canonical basis in $\mathbb{R}^n$. By the definition of $\mathcal U$ in \eqref{eq:A}, it is easy to verify that
\begin{align*}
2(\mathcal U^T \mathcal E \mathcal U)_{ij}=
\begin{cases}
E_{ij} + E_{ji} &\text{for } 1\le i,j\le r;\\
-E_{i,j-r} + E_{j-r,i} &\text{for } 1\le i \le r, r+1\le j \le 2r;\\
E_{i-r,j}-E_{j,i-r}  &\text{for } r+1\le i \le 2r, 1\le j \le r;\\
-E_{i-r,j-r} - E_{j-r,i-r} &\text{for } r+1\le i, j \le 2r.
\end{cases}
\end{align*}
Denote $E_r$ the $r\times r$ upper-left corner of $E$. We rewrite 
$$\mathcal U^T \mathcal E \mathcal U=\frac{1}{2} \begin{pmatrix}
E_r + E_r^\T & -E_r + E_r^\T\\
E_r - E_r^\T& -E_r - E_r^\T\\
\end{pmatrix} = \frac{1}{2}\begin{pmatrix}
I_r& I_r\\
I_r& -I_r
\end{pmatrix} \begin{pmatrix}
E_r& 0\\
0& E_r^\T
\end{pmatrix}
\begin{pmatrix}
I_r& -I_r\\
I_r& I_r
\end{pmatrix}.
$$By elementary computation and the fact that $E_r E_r^\T$ and $E_r^\T E_r$ share the same non-trivial eigenvalues, we get
$$\|\mathcal U^T \mathcal E \mathcal U\| \le \frac{1}{2}\left\|\begin{pmatrix}
I_r& I_r\\
I_r& -I_r
\end{pmatrix}\right\|\left\| \begin{pmatrix}
E_r& 0\\
0&E_r^\T
\end{pmatrix} \right\|
\left\|\begin{pmatrix}
I_r& -I_r\\
I_r& I_r
\end{pmatrix}\right\| = \|E_r\|.$$
Hence, by invoking Lemma \ref{lemma:norm}, we have
$$\| \mathcal U^T \mathcal E \mathcal U \|  \le \|E_r\| \le 2\sqrt{r} + \sqrt{2K\log(N+n)}$$
with probability at least $1-2(N+n)^{-K}$.

\bibliography{eigenvectorsine} 
\bibliographystyle{abbrv}

\end{document}